\documentclass[runningheads]{llncs}

\usepackage{eccv}

\usepackage{eccvabbrv}

\usepackage{graphicx}
\usepackage{booktabs}
\usepackage{amsmath}
\usepackage{amssymb}
\usepackage{mathtools}

\usepackage{xcolor}
\usepackage{pifont}

\usepackage[accsupp]{axessibility}  

\usepackage{tabularx}
\usepackage{multirow}
\usepackage{array}
\newcolumntype{Y}{>{\raggedright\arraybackslash}X}

\definecolor{stage2border}{HTML}{10B981}
\definecolor{dropcolor}{HTML}{EF4444}
\newcommand{\cmark}{\textcolor{stage2border}{\ding{51}}}
\newcommand{\xmark}{\textcolor{dropcolor}{\ding{55}}}
\newcolumntype{C}{>{\centering\arraybackslash}X}
\newcommand{\hdrone}[1]{\shortstack[c]{\textbf{#1}\\\phantom{(Alignment)}}}
\newcommand{\hdrtwo}[2]{\shortstack[c]{\textbf{#1}\\(#2)}}

\usepackage{hyperref}

\usepackage{orcidlink}

\usepackage[capitalize,noabbrev]{cleveref}
\crefname{appendix}{Appendix}{Appendices}
\Crefname{appendix}{Appendix}{Appendices}
\crefname{subappendix}{Appendix}{Appendices}
\Crefname{subappendix}{Appendix}{Appendices}
\crefname{subsubappendix}{Appendix}{Appendices}
\Crefname{subsubappendix}{Appendix}{Appendices}

\begin{document}

\title{MOOZY: A Patient-First Foundation Model for Computational Pathology}

\titlerunning{MOOZY: A Patient-First Foundation Model for Computational Pathology}

\author{%
  Yousef Kotp\inst{1,2}\orcidlink{0009-0003-0674-5230} \and
  Vincent Quoc-Huy Trinh\inst{3,4} \and
  Christopher Pal\inst{2,5,6} \and
  Mahdi S. Hosseini\inst{1,2,7}\orcidlink{0000-0002-9147-0731}%
}

\authorrunning{Kotp et al.}

\institute{%
  CSSE, Concordia University, Montr\'{e}al, Canada \and
  Mila -- Qu\'{e}bec AI Institute, Montr\'{e}al, Canada \and
  CHUM, Universit\'{e} de Montr\'{e}al, Montr\'{e}al, Canada \and
  IRIC, Universit\'{e} de Montr\'{e}al, Montr\'{e}al, Canada \and
  Universit\'{e} de Montr\'{e}al, Montr\'{e}al, Canada \and
  Canada CIFAR Chair, Polytechnique Montr\'{e}al, Montr\'{e}al, Canada \and
  Dept.\ of Pathology, McGill University, Montr\'{e}al, Canada%
}

\maketitle

\begingroup
\centering
\small
\textbf{Code:} \href{https://github.com/AtlasAnalyticsLab/MOOZY}{\nolinkurl{github.com/AtlasAnalyticsLab/MOOZY}}\par
\endgroup
\vspace{0.5em}

\begin{abstract}
Computational pathology needs whole-slide image (WSI) foundation models that transfer across diverse clinical tasks, yet current approaches remain largely slide-centric, often depend on private data and expensive paired-report supervision, and do not explicitly model relationships among multiple slides from the same patient. We present MOOZY, a patient-first pathology foundation model in which the patient case, not the individual slide, is the core unit of representation. MOOZY explicitly models dependencies across all slides from the same patient via a case transformer during pretraining, combining multi-stage self-supervision with scaled low-cost task supervision. In Stage 1, we pretrain a vision-only slide encoder on 77,134 public slide feature grids using masked self-distillation. In Stage 2, we align these representations with clinical semantics using a case transformer and multi-task supervision over 333 tasks from 56 public datasets, including 205 classification and 128 survival tasks across four endpoints. Across sixteen held-out tasks, MOOZY improves macro weighted F1, balanced accuracy, and macro weighted ROC-AUC relative to PRISM by +4.19\%, +7.93\%, and +6.95\%, respectively. MOOZY is also parameter efficient with 85.77M parameters, 14$\times$ smaller than GigaPath. These results suggest that patient-level pretraining yields transferable embeddings, providing a path toward scalable patient-first histopathology foundation models.
\end{abstract}

\section{Introduction}
\label{sec:intro}
\begin{figure*}[tb]
  \centering
  \begin{minipage}[t]{0.31\linewidth}
    \centering
    \includegraphics[width=\linewidth]{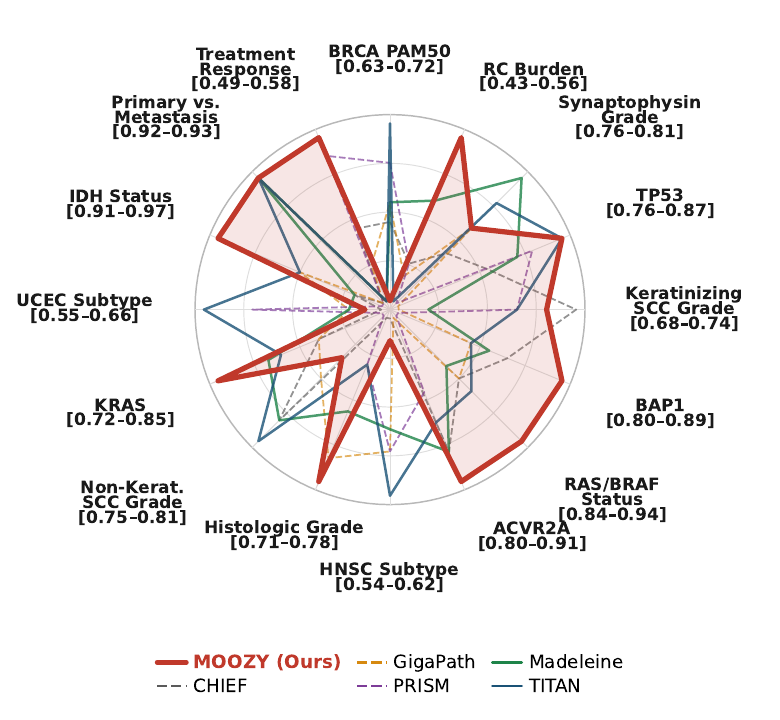}\\[1pt]
    \textbf{(a)}
  \end{minipage}%
  \hspace{0.015\linewidth}%
  \begin{minipage}[t]{0.31\linewidth}
    \centering
    \includegraphics[width=\linewidth]{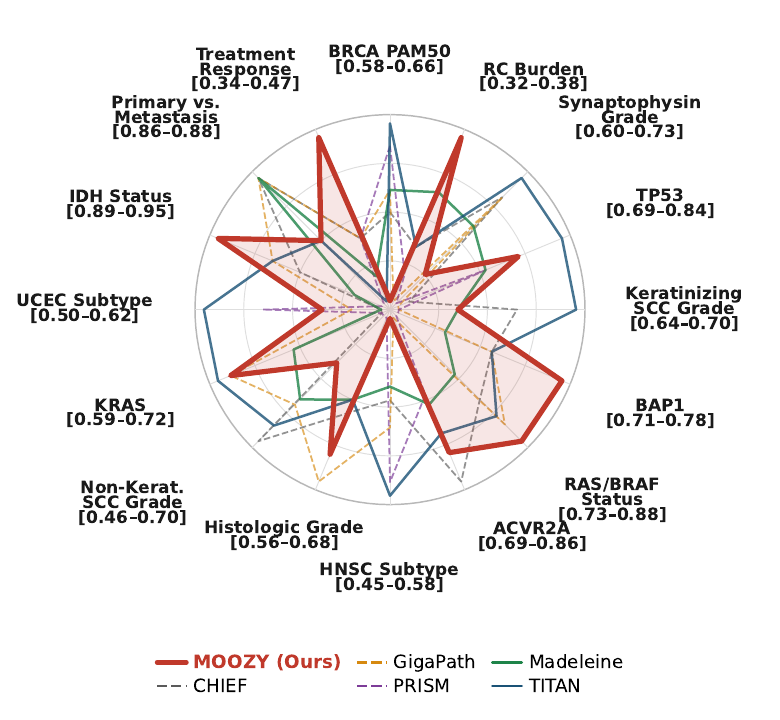}\\[1pt]
    \textbf{(b)}
  \end{minipage}%
  \hspace{0.015\linewidth}%
  \begin{minipage}[t]{0.31\linewidth}
    \centering
    \includegraphics[width=\linewidth]{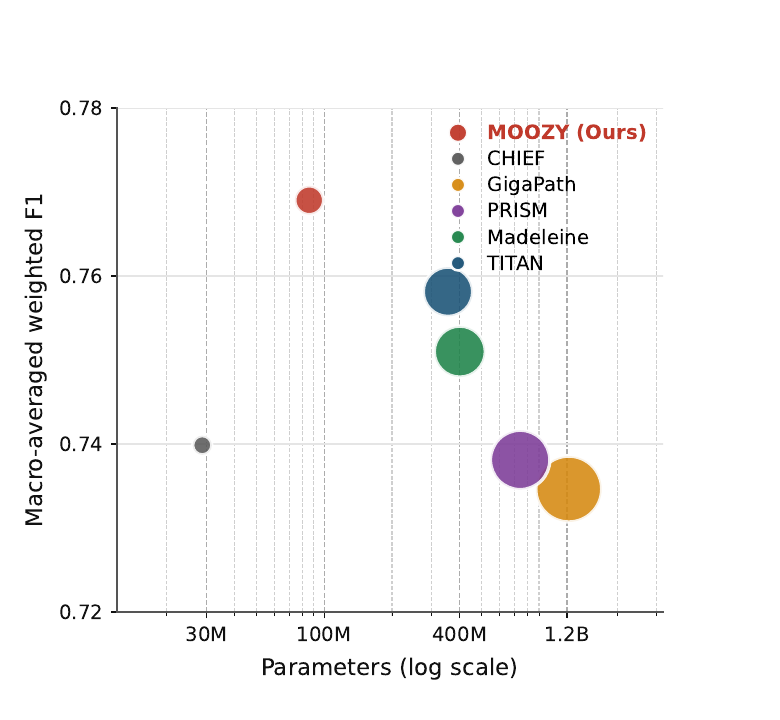}\\[1pt]
    \textbf{(c)}
  \end{minipage}
  \captionsetup{skip=2pt}
  \caption{%
    \textbf{(a)} MLP-probe weighted F1 across sixteen held-out tasks.
    \textbf{(b)} Linear-probe weighted F1 across the same sixteen held-out tasks.
    In both (a) and (b), brackets report \mbox{[min\,--\,max]} weighted F1 per task, with the center corresponding to the minimum and the outer ring to the maximum observed value.
    \textbf{(c)} Macro-averaged weighted F1 (MLP probe) versus total parameter count (log scale).}
  \label{fig:encoder_overview}
\end{figure*}

The fundamental challenge in computational pathology is learning whole-slide image (WSI) representations that transfer across cancer types, clinical endpoints, and patient populations without task-specific retraining. Much of the field has historically advanced through task- and cohort-specific supervised pipelines~\cite{bejnordi2017diagnostic, bulten2022artificial, courtiol2019deep, kather2020pan, skrede2020deep} that must be rebuilt whenever the organ, scanner domain, or clinical objective changes, limiting scalability and reuse. A generalizable alternative requires models capable of encoding diagnostically relevant structure from unlabeled data at two levels of scale that are critical in clinical workflows: the intra-slide level, where diagnostic meaning emerges from long-range interactions across tissue regions rather than isolated patch morphology, and the patient level, where multiple slides from the same patient must be jointly interpreted to form predictions. Self-supervised learning in vision~\cite{chen2020simple, he2020momentum, grill2020bootstrap, chen2021exploring, caron2021emerging, oquab2023dinov2, he2022masked, zhou2021ibot} has demonstrated that scaling data and compute can yield general-purpose encoders with strong task-agnostic transferability~\cite{brown2020language, kaplan2020scaling, hoffmann2022training, wei2022emergent}, suggesting a similar paradigm shift is possible in pathology.

However, applying these methods to WSIs is non-trivial: WSIs are gigapixel-scale, and clinically relevant semantics arise from interactions between cellular detail and global tissue architecture. Consequently, early pathology foundation models emerged as \emph{tile-level} encoders, with the field converging on the DINOv2~\cite{oquab2023dinov2} pretraining recipe and scaling from ViT-Large~\cite{chen2024towards, filiot2024phikon} to ViT-Giant backbones~\cite{xu2024whole, nechaev2024hibou} trained on up to millions of slides~\cite{vorontsov2023virchow, zimmermann2024virchow2}. In typical pipelines, tile features are extracted and a separate MIL aggregator~\cite{ilse2018attention, lu2021data, shao2021transmil} is trained for each downstream task, necessitating retraining whenever the clinical endpoint changes. More recently, the community has moved toward \emph{slide-level} encoders that pretrain whole-slide representations, reducing reliance on task-specific MIL training. These can be broadly grouped into vision-only self-supervised methods~\cite{chen2022scaling, lazard2023gigassl, hou2024self, xu2024gigapath, belagali2025ticon, lenz2025cobra}, multimodal approaches that align slides with text~\cite{shaikovski2024prism, ding2024multimodal, xiang2025musk}, genomic profiles~\cite{jaume2024tangle, xu2025mstar, vaidya2501molecular}, or cross-stain views~\cite{jaume2024multistain, hua2024pathoduet}, and supervised methods that learn from task labels~\cite{nicke2025tissue, wang2024pathology, shao2025miltransfer}.

Despite this progress, structural limitations persist. Many top-performing models rely on proprietary data, and some do not release checkpoints~\cite{hou2024self, vaidya2501molecular} or training recipes~\cite{ding2024multimodal,vaidya2501molecular,shaikovski2024prism}. Current architectures concentrate capacity in heavyweight tile encoders~\cite{shaikovski2024prism,ding2024multimodal,jaume2024multistain,vaidya2501molecular} while using lightweight slide aggregators, even though the key challenge in WSIs is long-range context rather than per-tile morphology. Finally, while some methods accommodate multiple slides per patient, they typically rely on simple fusion heuristics: early fusion by concatenating or unioning patch features into one enlarged bag, or late fusion by averaging slide-level embeddings or predictions across slides~\cite{shaikovski2024prism, lenz2025cobra, vaidya2501molecular, xu2025mstar, vaidya2501molecular}. These strategies treat a case as an unordered pool rather than explicitly modeling slide-to-slide relationships, discarding cross-slide interactions that carry diagnostic signal.

To this end, we introduce MOOZY, a \emph{patient-first} model in the sense that the patient case, not the individual slide, serves as the fundamental unit of representation. Rather than encoding slides independently and merging their embeddings post-hoc, MOOZY explicitly models dependencies across all slides belonging to the same patient via a dedicated case transformer during pretraining. We adopt a lightweight patch encoder~\cite{kang2023benchmarking} that compresses gigapixel slides into feature grids, making slide-level SSL tractable. MOOZY decouples representation quality from task-specific adaptation where Stage~1 pretrains a slide encoder on unlabeled public whole-slide images via self-supervised learning, establishing general-purpose spatial representations without any label signal. Stage~2 then steers these representations toward clinical semantics through large-scale multi-task supervision, benefiting from the generalizable prior built in Stage~1.Stage~2 moves beyond per-slide encoding where a case-level aggregator explicitly models dependencies across all slides of the same patient, rather than collapsing multi-slide cases into a single bag or averaging independent predictions. To the best of our knowledge, this is the first method that jointly addresses slide-level representation learning and explicit patient-level inter-slide dependency modeling. Our contributions can be summarized as follows:
\begin{itemize}
    \item We propose a two-stage framework that decouples vision-only slide SSL pretraining from patient-aware semantic alignment, where a case-level aggregator explicitly models dependencies across all slides of the same patient, making MOOZY the first attempt to move pathology foundational models beyond naive early/late multi-slide fusion.
    \item We construct a large-scale multi-task supervision regime spanning 333 tasks from 56 public datasets, covering classification and four survival endpoints across 23 anatomical sites, entirely from public data without private slides, paired reports, or expert annotations.
    \item We provide comprehensive quantitative and qualitative evaluation by benchmarking MOOZY on sixteen held-out tasks against both slide encoders and MIL baselines, complemented by attention map analysis and embedding visualization, showing competitive, transferable, and parameter-efficient representations.
\end{itemize}

\section{Related Work}
\label{sec:related_work}

{\textbf{Pathology Patch Encoders.}}\label{related_work:pathology_patch_encoders} Pathology-specific SSL on public tiles outperforms ImageNet initialization~\cite{kang2023benchmarking, filiot2023scaling}, leveraging vision SSL advances~\cite{chen2020simple, he2020momentum, grill2020bootstrap, zbontar2021barlow, chen2021exploring, caron2020unsupervised, he2022masked, zhou2021ibot, caron2021emerging, oquab2023dinov2, simeoni2025dinov3}. The field has converged on the DINOv2 recipe~\cite{oquab2023dinov2} combining self-distillation, masked image modeling, KoLeo regularization~\cite{kozachenko1987sample, sablayrolles2019spreading}, and KDE-based objectives~\cite{wang2020understanding}, with vision--language alignment~\cite{lu2024visual, ding2024multimodal} and knowledge distillation~\cite{filiot2025distilling} as complementary directions. Architectures have scaled from ViT-Large~\cite{chen2024towards, filiot2024phikon, yan2025pathorchestra} to ViT-Huge~\cite{vorontsov2023virchow, zimmermann2024virchow2, chen2024towards} and ViT-Giant~\cite{xu2024whole, hoptimus0, hoptimus1, nechaev2024hibou} on proprietary~\cite{chen2024towards, vorontsov2023virchow, nechaev2024hibou} and public data~\cite{TCGA_Program_NCI, edwards2015cptac, gtex2013genotype, filiot2023scaling, filiot2024phikon}. Yet scaling laws for tile encoders remain unclear: models match much larger systems~\cite{karasikov2025trainingstateoftheartpathologyfoundation, openmidnight2025}, suggesting benchmark discriminability~\cite{wei2021mhist} and training-recipe effects dominate data volume. We hypothesize this saturation is fundamental as H\&E tissue occupies a far more constrained visual space than natural images, with a narrow color palette and bounded set of morphological primitives (\eg, cell types, glandular architectures, stromal patterns), so tile-level representations approach a performance ceiling well before general-vision thresholds. The true bottleneck therefore lies in \emph{slide- and context-level modeling}.

{\textbf{Multi-Instance Learning.}}\label{related_work:multi_instance_learning} MIL treats a WSI as a bag of patch features with a single slide-level label, with approaches spanning permutation-invariant pooling~\cite{campanella2019clinical}, attention scoring~\cite{ilse2018attention, lu2021data}, transformer-based inter-patch modeling~\cite{shao2021transmil, vaswani2017attention}, dual-stream objectives~\cite{li2021dual}, pseudo-bag augmentation~\cite{zhang2022dtfd}, and efficient variants via low-rank approximations~\cite{xiang2023exploring}, knowledge graphs~\cite{li2024dynamic}, and regional re-embedding~\cite{tang2024feature}. All aggregators are trained from scratch per task, motivating universal pretrained slide representations.

{\textbf{Slide Encoders.}}\label{related_work:slide_encoders} Slide-level pretraining operates on unordered sets of thousands of heterogeneous tile embeddings. Vision-only methods apply self-distillation~\cite{chen2022scaling, caron2021emerging}, contrastive tile sampling~\cite{lazard2023gigassl, chen2020simple}, view transformations~\cite{hou2024self}, dilated attention in masked autoencoders~\cite{xu2024gigapath, ding2023longnet}, lightweight contextualizers~\cite{belagali2025ticon}, and state-space contrastive learning~\cite{lenz2025cobra, gu2023mamba}. Multimodal methods align slides with clinical text~\cite{shaikovski2024prism, ding2024multimodal, xiang2025musk, lu2023foundational}, genomic or transcriptomic profiles~\cite{jaume2024tangle, xu2025mstar, vaidya2501molecular}, or cross-stain sections~\cite{jaume2024multistain, hua2024pathoduet}. Supervised approaches train on slide-level labels~\cite{nicke2025tissue, nicke2025tissuev1, wang2024pathology, shao2025miltransfer}. Three gaps persist: many methods rely on expensive proprietary data~\cite{vaidya2501molecular, ding2024multimodal, shaikovski2024prism}, capacity concentrates in tile encoders over slide aggregators~\cite{shaikovski2024prism, ding2024multimodal, jaume2024multistain, vaidya2501molecular}, and multi-slide fusion remains naive~\cite{shaikovski2024prism, lenz2025cobra, vaidya2501molecular, xu2025mstar}, treating cases as unordered pools. MOOZY directly addresses all three: a two-stage design decouples vision-only SSL pretraining from case-aware multi-task alignment, replacing naive multi-slide fusion with explicit inter-slide dependency modeling at the case level, while training with cheap available supervision.

\section{Methodology}
\label{sec:methodology}

{\textbf{Stage 1: Self-Supervised Slide Encoder Pretraining.}}\label{sec:ssl_pretraining}\begin{figure}[tb]
  \centering
  \includegraphics[width=0.97\linewidth]{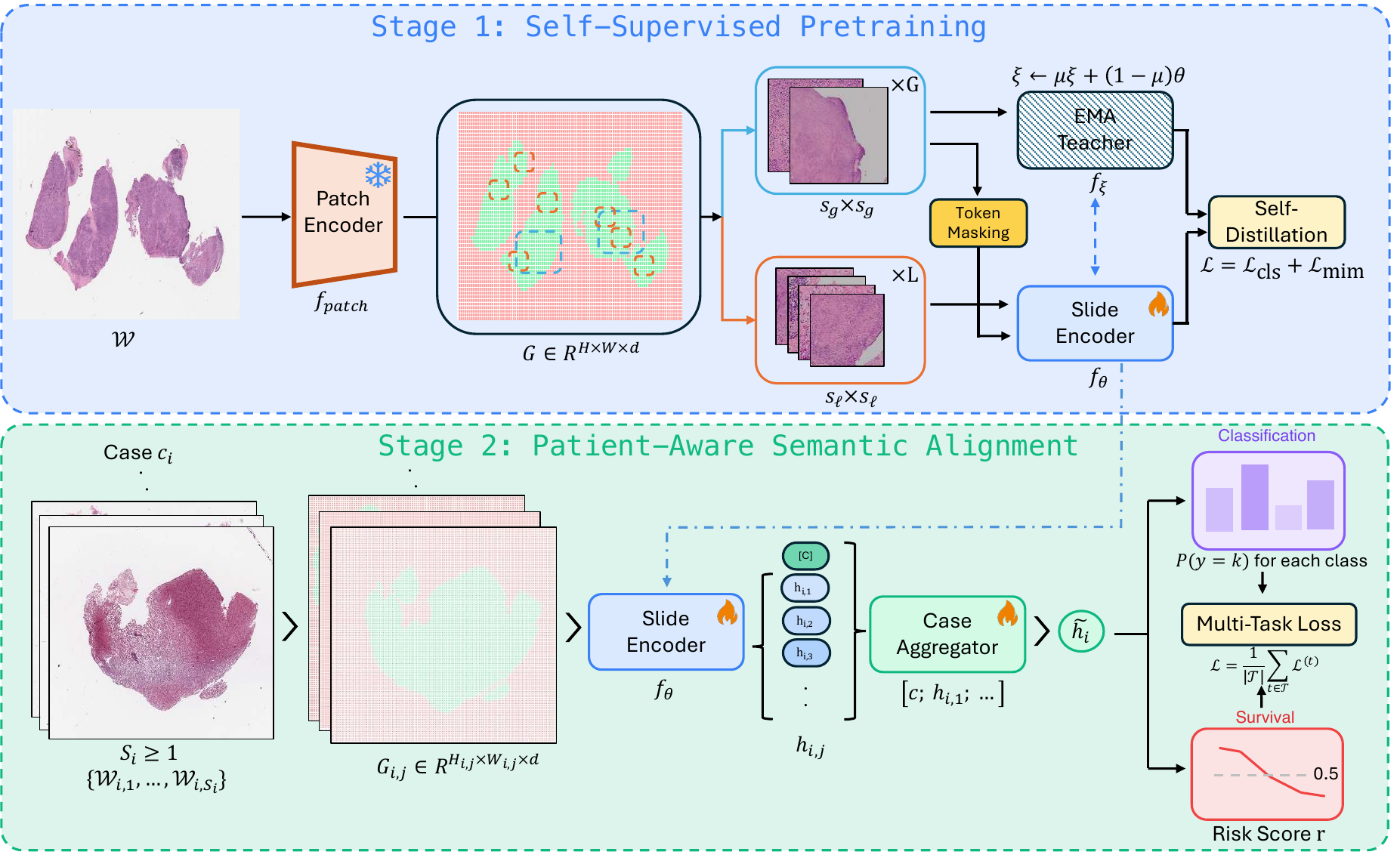}
  \caption{Overview of the proposed two-stage framework. \textit{Stage~1 (up):} A frozen patch encoder extracts per-patch features arranged into a spatial grid. Multi-scale crops are sampled with spatial augmentations and block-based masking. A student slide encoder and EMA teacher are jointly trained via CLS-level self-distillation ($\mathcal{L}_\text{cls}$) and masked patch prediction ($\mathcal{L}_\text{mim}$). \textit{Stage~2 (down):} The pretrained slide encoder produces per-slide embeddings; a case transformer aggregates them into a unified case embedding $\tilde{\mathbf{h}}_i$, routed to task-specific classification and survival heads.}
  \label{fig:pipeline}
\end{figure}
We cast WSI representation learning as self-supervised pretraining on precomputed patch features (\cref{fig:pipeline}, up). Given a WSI $\mathcal{W}$, we partition tissue into non-overlapping 224-pixel patches and extract features with a frozen patch encoder $f_\text{patch}$:
\begin{equation}
    \mathbf{z}_i = f_\text{patch}(p_i), \quad \mathbf{z}_i \in \mathbb{R}^{d_\text{patch}}.
\end{equation}
We arrange patch features and coordinates into a 2-D grid $\mathbf{G} \in \mathbb{R}^{H \times W \times d_\text{patch}}$ with a binary validity mask for tissue positions (grid construction in \Cref{appendix_subsec:spatial_grid}). If a slide is available at multiple magnifications, each level-specific grid is treated as an independent training sample.
To capture global context and local detail, we sample $G$ global crops of size $S_g \times S_g$ and $L$ local crops of size $S_l \times S_l$ ($S_g > S_l$) uniformly from valid grid locations, with each crop required to satisfy a minimum valid-token ratio $\rho_{\min}$. Unlike~\cite{ding2024multimodal}, which draws global and local views from the same fixed ROI, we sample crops independently over the full slide grid. This increases spatial diversity and lowers view mutual information, which benefits self-supervised WSI representation learning~\cite{hou2024self}. We apply DINOv3-style block masking~\cite{simeoni2025dinov3} to global crops only. Because histopathology tissue is spatially continuous, contiguous masking encourages reasoning over broader morphology instead of reconstructing isolated tokens. In each batch, we select a fraction of global crops for masking, assign mask ratios uniformly over $[\gamma_{\min}, \gamma_{\max}]$, and shuffle them for uniform coverage of the masking range. We then iteratively place rectangular blocks with log-uniform aspect ratios until the target fraction of valid tokens is masked, restricting masking to tissue tokens. The full algorithm is in \Cref{appendix_subsec:block_masking}.

\begin{figure}[tb]
  \centering
  \includegraphics[width=0.53\linewidth,trim=8 6 0 9,clip]{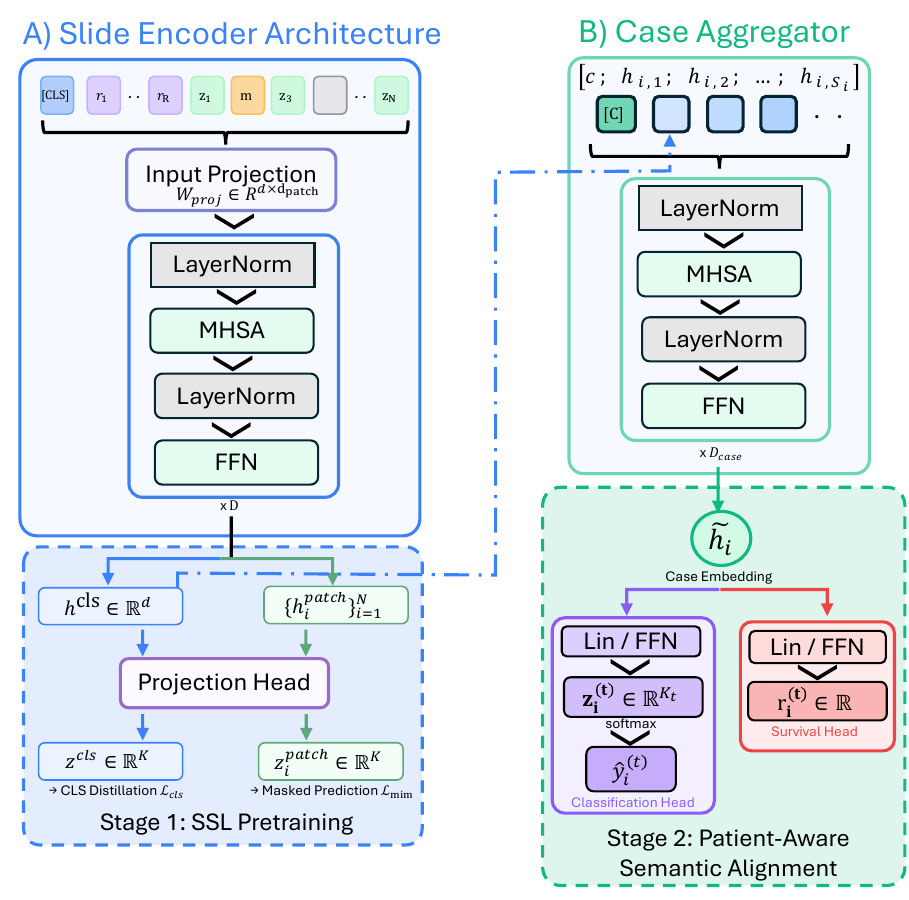}
  \caption{Architecture of the slide encoder and case aggregator. \textbf{(A)}~The slide encoder takes patch embeddings, a learnable \texttt{[CLS]} token, $R$ register tokens, and mask tokens, processed through $D$ transformer blocks. \textbf{(B)}~The case aggregator prepends a learnable \texttt{[CASE]} token to per-slide embeddings and produces a case embedding $\tilde{\mathbf{h}}_i$, routed to heads for classification and survival prediction.}
  \label{fig:slide_enc_case_agg}
\end{figure}

Our slide encoder (\cref{fig:slide_enc_case_agg}A) is a Vision Transformer~\cite{dosovitskiy2020image} adapted to precomputed feature grids. Patch features are projected to dimension $d$ with a linear layer and GELU~\cite{hendrycks2016gaussian}. We prepend a learnable \texttt{[CLS]} token and $R$ register tokens~\cite{darcet2024vision}, masked student positions are replaced by a learnable mask embedding. Each block uses pre-norm multi-head self-attention and an FFN with stochastic depth~\cite{Huang_2016}. LayerScale~\cite{Touvron_2021} is disabled in our training runs. To encode spatial structure without learned positional embeddings, we use 2-D ALiBi~\cite{press2022train} as adapted for WSIs in TITAN~\cite{ding2024multimodal}. For each attention head $h$, we add:
\begin{equation}
    b^{(h)}_{i,j} = -s_h \cdot \frac{\|\mathbf{p}_i - \mathbf{p}_j\|_2}{\Delta},
\end{equation}
where $\mathbf{p}_i$ are level-0 token coordinates, $\Delta$ is patch spacing, and $s_h > 0$ is a head-specific geometric slope. \texttt{[CLS]} and register tokens receive zero bias to remain spatially neutral. We also apply an additive attention mask that sets background-involving pairs to $-\infty$. The projection head maps encoder tokens to prototype logits with an MLP, an L2-normalized bottleneck, and a weight-normalized prototype layer~\cite{caron2021emerging,zhou2021ibot}, shared for \texttt{[CLS]} and patch tokens (full formulation in \Cref{appendix_sec:projection_head}).

We use an EMA teacher for self-distillation, updating teacher parameters as $\xi \leftarrow \mu \xi + (1 - \mu) \theta$ with cosine momentum schedule from $\mu_0$ to $\mu_T$. To avoid mode collapse, teacher outputs are centered with momentum-updated running averages~\cite{caron2021emerging}. The objective combines global CLS distillation and masked patch prediction. The teacher provides soft targets from global views, while the student predicts from all views:
\begin{equation}
    \mathcal{L}_\text{cls} = -\frac{1}{G(G+L-1)} \sum_{j=1}^{G} \sum_{\substack{i=1 \\ i \neq j}}^{G+L} \sum_{k=1}^{K} P^{(j)}_k \log Q^{(i)}_k,
\end{equation}
where $P^{(j)}$ and $Q^{(i)}$ are teacher and student softmax distributions with temperatures $\tau_t$ and $\tau_s$. For masked positions in global crops, the student additionally predicts teacher patch-level distributions:
\begin{equation}
    \mathcal{L}_\text{mim} = -\frac{1}{|\mathcal{M}|} \sum_{(j, r, c) \in \mathcal{M}} \sum_{k=1}^{K} P^{(j)}_{r,c,k} \log Q^{(j)}_{r,c,k},
\end{equation}
where $\mathcal{M}$ is the set of masked valid positions, and patch distributions use temperature $\tau_t^\text{patch}$. The total loss is $\mathcal{L} = \mathcal{L}_\text{cls} + \mathcal{L}_\text{mim}$.

{\textbf{Stage 2: Case-Aware Semantic Alignment.}}\label{sec:semantic_alignment}
A central design principle of MOOZY is to decouple representation learning from semantic alignment: Stage~1 builds a general-purpose slide encoder on unlabeled data, and Stage~2 steers it toward clinical utility through multi-task supervision without re-learning spatial representations from task labels alone. This contrasts with task-specific MIL pipelines that learn both aggregation and task adaptation simultaneously from scratch, and with multimodal slide encoders that couple representation quality to the availability of paired text or genomic data. Concretely, we fine-tune the Stage~1 encoder with multi-task supervision across diverse clinical endpoints (\cref{fig:pipeline}, down). Let $\mathcal{T} = \{\mathcal{T}_1, \ldots, \mathcal{T}_T\}$ be $T$ supervised tasks. Each case $c_i$ contains one or more WSIs $\{\mathcal{W}_{i,1}, \ldots, \mathcal{W}_{i,S_i}\}$, and each task provides either a class label (classification) or a time-to-event label with event indicator (survival).
Stage~2 uses full-slide grids without crop sampling. To handle gigapixel inputs under GPU memory limits, we apply a hardware-adaptive token cap $K_\text{max}(\cdot)$: if valid tokens exceed $K_\text{max}$, we perform stratified random sampling to preserve whole-slide spatial coverage (algorithm in \Cref{appendix_subsec:token_capping}). Retained tokens are compacted and passed to the slide encoder:
\begin{equation}
    \mathbf{h}_{i,j} = f_\theta(\mathbf{X}_{i,j}^\star, \mathbf{P}_{i,j}^\star, \Delta_{i,j}) \in \mathbb{R}^{d}.
\end{equation}
To form one case representation from slide embeddings $\mathbf{H}_i = \{\mathbf{h}_{i,1}, \ldots, \mathbf{h}_{i,S_i}\}$, we use a lightweight transformer aggregator (\cref{fig:slide_enc_case_agg}B). A learnable \texttt{[CASE]} token is prepended and processed through $D_\text{case}$ pre-norm transformer blocks with LayerScale, while DropPath is disabled, yielding:
\begin{equation}
    \tilde{\mathbf{h}}_i = \mathrm{LN}(\mathbf{z}_{i,D_\text{case}})[0] \in \mathbb{R}^{d}.
\end{equation}
We apply this aggregator to all cases, including single-slide cases ($S_i = 1$), so that the learned embedding space is always case-centric and consistent regardless of slide count at inference.
Each task $\mathcal{T}_t$ has a prediction head $g_t: \mathbb{R}^{d} \to \mathbb{R}^{o_t}$, either linear or MLP (formulations in \Cref{appendix_sec:task_heads}). For classification, we use weighted cross-entropy with label smoothing~\cite{Szegedy_2016} coefficient $\epsilon$:
\begin{equation}
    \mathcal{L}_\text{cls}^{(t)} = \mathrm{CE}_{w^{(t)},\epsilon}\!\left(\{\mathbf{z}_i^{(t)}, y_i^{(t)}\}_{i \in \mathcal{B}_t}\right),
\end{equation}
where class weights use inverse frequency: $w_k^{(t)} = |\mathcal{D}_t| / (K_t \cdot |\{i \in \mathcal{D}_t : y_i^{(t)} = k\}|)$, and $\mathcal{B}_t$ is the valid labeled set. For survival prediction, we use a discrete-hazard objective: survival times are quantized into $B_t$ bins with edges at training event-time quantiles, and $B_t$ adapts to per-task event count. With predicted hazards $h_{i,k}^{(t)} = \sigma(a_{i,k}^{(t)})$, we minimize the negative log-likelihood over events and censored cases (full loss and bin-selection details in \Cref{appendix_subsec:survival_bins}). For ranking-based metrics, hazards are converted to scalar risk:
\begin{equation}
    r_i^{(t)} = -\sum_{k=1}^{B_t} \log(1 - h_{i,k}^{(t)}).
\end{equation}
Let $\mathcal{T}_\text{active} \subseteq \mathcal{T}$ be tasks with usable supervision in the current batch. We average losses over active tasks:
\begin{equation}
    \mathcal{L} = \frac{1}{|\mathcal{T}_\text{active}|} \sum_{t \in \mathcal{T}_\text{active}} \mathcal{L}^{(t)},
\end{equation}
which naturally handles sparse multi-task labels by excluding unlabeled tasks for each case. At inference, the slide encoder and case transformer output $\tilde{\mathbf{h}}_i$ as the final case embedding, and task heads are discarded.

\section{Experiment Setup}
\label{sec:experiment_setup}
{\textbf{Dataset.}}
We collect 56 different open-sourced datasets which include: REG Dataset~\cite{lee2025reg2025}, TCGA (all 32 cohorts)~\cite{TCGA_Program_NCI}, CPTAC (all 10 cohorts)~\cite{edwards2015cptac}, BC-Therapy~\cite{sammut2022bctherapy}, BRACS~\cite{brancati2022bracs}, CAMELYON17~\cite{bandi2018detection}, DHMC Kidney~\cite{zhu2021development}, DHMC LUAD~\cite{wei2019pathologist}, EBRAINS~\cite{roetzer2022ebrainsdataset,roetzer2022digitalbrain}, IMP Colorectum~\cite{Oliveira2021,Neto2022,Neto2024}, IMP Cervix~\cite{Oliveira2023}, MBC~\cite{bergstrom2024hrd,galland2022mbc}, MUT-HET-RCC~\cite{rajaram2022muthetrcc}, NADT Prostate~\cite{wilkinson2020nascent}, NAT-BRCA~\cite{peikari2017automatic}, and PANDA~\cite{bulten2022artificial}. All collected slides are processed using AtlasPatch~\cite{atlaspatch}, which performs tissue segmentation using SAM2~\cite{kirillov2023segment, ravi2024sam} model. The resulting tissue masks define the valid regions from which non-overlapping $224 \times 224$ patches are extracted at both $20\times$ and $40\times$ magnification levels, reaching $\sim$1.6 billion extracted patches. We extract features for each patch using a pretrained lightweight patch encoder from \cite{kang2023benchmarking}, which has 21.67M parameters and the architecture of ViT-S~\cite{dosovitskiy2020image}. Detailed patch counts at each magnification are reported in \Cref{appendix_sec:dataset_statistics}.

\begin{figure*}[!t]
  \centering
  \includegraphics[width=\textwidth,height=0.50\textheight,keepaspectratio]{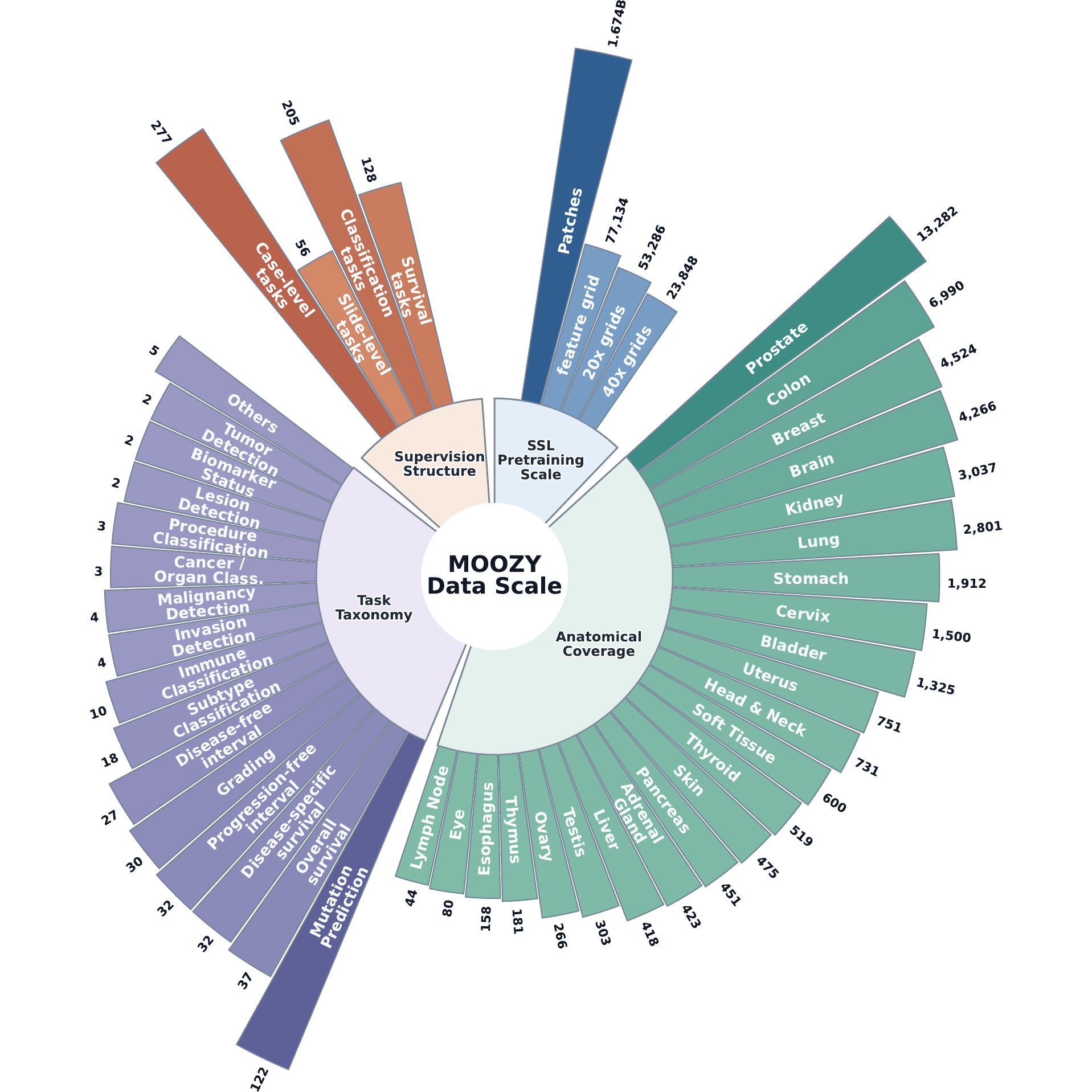}
  \captionsetup{skip=2pt}
  \caption{Radial hierarchy of MOOZY data scale across four dimensions: pretraining scale, anatomical coverage, task taxonomy, and supervision structure.}
  \label{fig:data_scale_overview}
\end{figure*}

For Stage~1 self-supervised pretraining, the preprocessing yields $77{,}134$ slide feature grids: $53{,}286$ at $20\times$ and $23{,}848$ at $40\times$ magnification, sourced from $\sim$31.8\,TB of raw WSI data. The two corresponding feature grids from those two magnification levels are treated as independent training samples and sampled uniformly. These slides span 23 distinct anatomical sites as shown in \Cref{fig:data_scale_overview}.

For Stage~2 supervised fine-tuning, we construct $333$ tasks ($205$ classification and $128$ survival) across all $56$ datasets, averaging $6$ tasks per dataset. Survival supervision covers overall survival (OS), disease-specific survival (DSS), disease-free interval (DFI), and progression-free interval (PFI), depending on endpoint availability. Of these, $56$ are slide-level and $277$ are case-level (\ie, predictions aggregated over all slides of a case). The labeled Stage~2 subset comprises $41{,}089$ supervised cases and $45{,}179$ unique whole-slide images, a subset of the $53{,}286$ Stage~1 slides, since unlabeled slides are excluded. The mean per-task slide-to-case ratio is $1.330$\footnote{Computed over tasks as the mean of each task's unique-slide / unique-case ratio, not as the global union ratio.}. Per-organ statistics, class counts, and task-construction details are provided in \Cref{appendix_sec:organ_task_distribution,tab:class_distribution_appendix,appendix_sec:task_preparation}. A small subset of tasks was drawn from prior work~\cite{zhang2025standardizing,vaidya2501molecular}. For a broader landscape overview of computational pathology tasks, see~\cite{hosseini2024computational}. A consolidated scale overview is shown in \Cref{fig:data_scale_overview}.

{\textbf{SSL Pretraining.}}\label{sec:ssl_pretraining_setup}
We train the slide encoder using the Stage~1 self-supervised framework (\cref{sec:methodology}) on all $77{,}134$ slide feature grids. Training uses $8$ GPUs with an effective batch of $1{,}024$ slides (micro batch $64$, $2$ accumulation steps) for $200$ epochs ($14{,}400$ optimizer steps, ${\approx}436$ GPU-hours). Full hyperparameters are listed in \Cref{appendix_sec:ssl_hyperparameters}.
The encoder is a $6$-layer transformer ($d{=}768$, $12$ heads, $4$ register tokens~\cite{darcet2024vision}). Multi-crop sampling uses $G{=}2$ global crops of $20{\times}20$ tokens and $L{=}4$ local crops of $12{\times}12$ tokens, with block masking applied to global crops at mask ratio $\gamma \sim \mathcal{U}[0.1, 0.5]$. Optimization uses AdamW~\cite{loshchilov2019decoupled} with a cosine learning rate schedule and an EMA teacher whose momentum follows a cosine schedule from $\mu_0{=}0.996$ to $\mu_T{=}1.0$.

{\textbf{Case-Aware Semantic Alignment.}}\label{sec:semantic_alignment_setup}
Following the Stage~2 framework (\cref{sec:methodology}), training uses $8$ GPUs with an effective batch of $1{,}024$ cases (micro batch $1$, $128$ accumulation steps) for $20$ epochs ($1{,}000$ optimizer steps, ${\approx}512$ GPU-hours). Complete hyperparameters are provided in \Cref{appendix_sec:semantic_alignment_hyperparameters}. For case-level aggregation, a case transformer ($D_\text{case}{=}3$ layers, $12$ heads, learnable \texttt{[CASE]} token) pools all slides assigned to a supervised case into a single embedding. We jointly train on $205$ classification and $128$ survival tasks ($333$ total) spanning all $56$ datasets. Classification tasks use cross-entropy with label smoothing ($\epsilon{=}0.03$) and inverse-frequency class weighting. Survival tasks spanning OS, DSS, DFI, and PFI use a discrete-time hazard model with adaptive per-task binning (target 8 bins, minimum 2, maximum 16) and NLL loss. Optimization uses AdamW~\cite{loshchilov2019decoupled} with base learning rate $5 \times 10^{-5}$, cosine schedule, and gradient clipping at $0.3$. A stratified $5\%$ (task-wise) validation holdout is used to monitor overfitting. A summary of stage-specific augmentation strategies is provided in Appendix~\Cref{tab:aug_strategies}, and visual examples are provided in Appendix~\Cref{fig:appendix_augmentation_strategies}.

\section{Results}
\label{sec:results}
We evaluate on sixteen held-out tasks spanning diverse clinical settings. These cover Residual Cancer Burden (BC Therapy), TP53 mutation (CPTAC-BRCA), BAP1 mutation (CPTAC-CCRCC), ACVR2A mutation (CPTAC-COAD), Histologic Grade (CPTAC-LSCC), KRAS mutation (CPTAC-LUAD), IDH Status (EBRAINS), Treatment Response (MBC), molecular subtypes from TCGA-BRCA, TCGA-HNSC, and TCGA-UCEC, keratinizing and non-keratinizing squamous cell carcinoma grade and primary-versus-metastatic classification (HANCOCK)~\cite{zhang2025standardizing,dorrich2025hancock}, and RAS/BRAF status and synaptophysin grade (VALENTINO)~\cite{zhang2025standardizing,trahearn2025crc}. These tasks are excluded from all training stages. Fifteen are case-level and IDH Status is slide-level\footnote{RC Burden is case-level task, but the slide to case ratio is 1.0}. Slide encoders use frozen-feature MLP probes, while MIL baselines train task-specific aggregators on frozen patch features. Both follow five-fold evaluation with case-level fold grouping. We report mean $\pm$ standard deviation for weighted F1, weighted ROC-AUC, and balanced accuracy. Full protocol details are in \Cref{appendix_sec:mlp_probe_setup}. Linear probe results are in \Cref{appendix_sec:linear_probe_setup,tab:linear_probe_results,tab:linear_probe_mil_results}.

{\textbf{Comparison with Slide Encoders.}}
We compare MOOZY against five slide encoders. For case-level tasks, baseline encoders produce case representations by averaging per-slide embeddings before probing, while MOOZY uses its native case-level embedding from the case transformer. Results are in \Cref{tab:encoder_comparison}. Across all sixteen tasks, MOOZY achieves the strongest macro weighted F1 (0.769) and balanced accuracy (0.702), TITAN achieves the strongest weighted ROC-AUC (0.768 versus 0.763).

\begin{table}[p]
\caption{Frozen-feature MLP probe comparison against slide encoder baselines on sixteen held-out tasks. \textbf{Bold}: best; \underline{underline}: second best.}
\label{tab:encoder_comparison}
\centering
\scriptsize
\setlength{\tabcolsep}{2pt}
\renewcommand{\arraystretch}{0.66}
\resizebox{\linewidth}{!}{%
\begin{tabular}{%
  >{\raggedright\arraybackslash}p{1.9cm}
  c
  c
  c
  c
  c
  c
  c}
\toprule
Task & Metric & CHIEF & \shortstack{Giga-\\Path} & PRISM & \shortstack{Made-\\leine} & TITAN & \shortstack{MOOZY\\(Ours)} \\
\midrule
\multirow{3}{*}{\parbox{1.9cm}{\raggedright Residual Cancer Burden}}
  & F1        & $0.46_{\pm0.03}$ & $0.45_{\pm0.05}$ & $0.46_{\pm0.07}$ & $\underline{0.51}_{\pm0.03}$ & $0.43_{\pm0.07}$ & $\mathbf{0.56}_{\pm0.05}$ \\
  & AUC       & $0.60_{\pm0.05}$ & $0.55_{\pm0.08}$ & $0.58_{\pm0.06}$ & $\underline{0.63}_{\pm0.03}$ & $0.58_{\pm0.07}$ & $\mathbf{0.74}_{\pm0.04}$ \\
  & Bal.\ Acc & $0.44_{\pm0.05}$ & $0.40_{\pm0.05}$ & $0.43_{\pm0.04}$ & $\underline{0.48}_{\pm0.03}$ & $0.38_{\pm0.11}$ & $\mathbf{0.51}_{\pm0.06}$ \\
\midrule
\multirow{3}{*}{\parbox{1.9cm}{\raggedright TP53 Mut.}}
  & F1        & $0.82_{\pm0.05}$ & $0.76_{\pm0.03}$ & $\underline{0.85}_{\pm0.03}$ & $0.84_{\pm0.06}$ & $\mathbf{0.87}_{\pm0.04}$ & $\underline{0.87}_{\pm0.04}$ \\
  & AUC       & $0.81_{\pm0.09}$ & $0.76_{\pm0.06}$ & $0.85_{\pm0.05}$ & $0.85_{\pm0.08}$ & $\mathbf{0.91}_{\pm0.04}$ & $\underline{0.86}_{\pm0.06}$ \\
  & Bal.\ Acc & $0.83_{\pm0.04}$ & $0.76_{\pm0.04}$ & $0.84_{\pm0.04}$ & $0.84_{\pm0.05}$ & $\mathbf{0.88}_{\pm0.04}$ & $\underline{0.86}_{\pm0.05}$ \\
\midrule
\multirow{3}{*}{\parbox{1.9cm}{\raggedright BAP1 Mut.}}
  & F1        & $\underline{0.86}_{\pm0.04}$ & $0.84_{\pm0.06}$ & $0.80_{\pm0.07}$ & $0.85_{\pm0.08}$ & $0.84_{\pm0.06}$ & $\mathbf{0.89}_{\pm0.06}$ \\
  & AUC       & $0.75_{\pm0.09}$ & $0.63_{\pm0.17}$ & $0.71_{\pm0.09}$ & $0.78_{\pm0.11}$ & $\mathbf{0.82}_{\pm0.06}$ & $\underline{0.79}_{\pm0.12}$ \\
  & Bal.\ Acc & $0.75_{\pm0.12}$ & $0.66_{\pm0.14}$ & $0.66_{\pm0.10}$ & $0.75_{\pm0.12}$ & $\underline{0.75}_{\pm0.11}$ & $\mathbf{0.78}_{\pm0.11}$ \\
\midrule
\multirow{3}{*}{\parbox{1.9cm}{\raggedright ACVR2A Mut.}}
  & F1        & $\underline{0.89}_{\pm0.07}$ & $0.80_{\pm0.10}$ & $0.85_{\pm0.03}$ & $0.89_{\pm0.09}$ & $0.87_{\pm0.05}$ & $\mathbf{0.91}_{\pm0.05}$ \\
  & AUC       & $0.80_{\pm0.13}$ & $0.74_{\pm0.11}$ & $\underline{0.83}_{\pm0.08}$ & $0.76_{\pm0.19}$ & $0.79_{\pm0.10}$ & $\mathbf{0.91}_{\pm0.09}$ \\
  & Bal.\ Acc & $0.80_{\pm0.12}$ & $0.65_{\pm0.10}$ & $\underline{0.81}_{\pm0.08}$ & $0.81_{\pm0.16}$ & $0.76_{\pm0.15}$ & $\mathbf{0.90}_{\pm0.10}$ \\
\midrule
\multirow{3}{*}{\parbox{1.9cm}{\raggedright Histologic Grade}}
  & F1        & $0.71_{\pm0.07}$ & $\underline{0.77}_{\pm0.03}$ & $0.73_{\pm0.11}$ & $0.75_{\pm0.06}$ & $0.73_{\pm0.06}$ & $\mathbf{0.78}_{\pm0.08}$ \\
  & AUC       & $0.71_{\pm0.09}$ & $\mathbf{0.77}_{\pm0.04}$ & $0.67_{\pm0.11}$ & $0.74_{\pm0.08}$ & $0.71_{\pm0.04}$ & $\underline{0.75}_{\pm0.15}$ \\
  & Bal.\ Acc & $0.73_{\pm0.06}$ & $\underline{0.77}_{\pm0.03}$ & $0.73_{\pm0.12}$ & $0.74_{\pm0.06}$ & $0.73_{\pm0.06}$ & $\mathbf{0.77}_{\pm0.08}$ \\
\midrule
\multirow{3}{*}{\parbox{1.9cm}{\raggedright KRAS Mut.}}
  & F1        & $0.77_{\pm0.08}$ & $0.77_{\pm0.08}$ & $0.72_{\pm0.07}$ & $\underline{0.81}_{\pm0.06}$ & $0.80_{\pm0.05}$ & $\mathbf{0.85}_{\pm0.04}$ \\
  & AUC       & $0.76_{\pm0.14}$ & $0.72_{\pm0.09}$ & $0.61_{\pm0.12}$ & $0.70_{\pm0.05}$ & $\underline{0.80}_{\pm0.05}$ & $\mathbf{0.80}_{\pm0.06}$ \\
  & Bal.\ Acc & $0.74_{\pm0.13}$ & $0.76_{\pm0.08}$ & $0.63_{\pm0.13}$ & $0.77_{\pm0.07}$ & $\mathbf{0.81}_{\pm0.05}$ & $\underline{0.79}_{\pm0.10}$ \\
\midrule
\multirow{3}{*}{\parbox{1.9cm}{\raggedright IDH Status}}
  & F1        & $0.92_{\pm0.01}$ & $\underline{0.94}_{\pm0.02}$ & $0.91_{\pm0.02}$ & $0.92_{\pm0.02}$ & $0.94_{\pm0.02}$ & $\mathbf{0.97}_{\pm0.02}$ \\
  & AUC       & $0.96_{\pm0.01}$ & $0.97_{\pm0.02}$ & $0.95_{\pm0.01}$ & $0.96_{\pm0.01}$ & $\underline{0.97}_{\pm0.01}$ & $\mathbf{0.99}_{\pm0.01}$ \\
  & Bal.\ Acc & $0.92_{\pm0.01}$ & $0.94_{\pm0.02}$ & $0.91_{\pm0.02}$ & $0.91_{\pm0.02}$ & $\underline{0.94}_{\pm0.02}$ & $\mathbf{0.97}_{\pm0.02}$ \\
\midrule
\multirow{3}{*}{\parbox{1.9cm}{\raggedright Treatment Response}}
  & F1        & $0.53_{\pm0.07}$ & $0.51_{\pm0.07}$ & $\underline{0.57}_{\pm0.08}$ & $0.49_{\pm0.05}$ & $0.49_{\pm0.02}$ & $\mathbf{0.58}_{\pm0.14}$ \\
  & AUC       & $\mathbf{0.70}_{\pm0.06}$ & $0.68_{\pm0.10}$ & $\underline{0.69}_{\pm0.07}$ & $0.59_{\pm0.05}$ & $0.60_{\pm0.06}$ & $0.68_{\pm0.07}$ \\
  & Bal.\ Acc & $0.48_{\pm0.10}$ & $0.40_{\pm0.08}$ & $\mathbf{0.51}_{\pm0.11}$ & $0.35_{\pm0.09}$ & $0.37_{\pm0.06}$ & $\underline{0.48}_{\pm0.17}$ \\
\midrule
\multirow{3}{*}{\parbox{1.9cm}{\raggedright BRCA PAM50 Subtype}}
  & F1        & $0.67_{\pm0.02}$ & $0.68_{\pm0.05}$ & $\underline{0.70}_{\pm0.04}$ & $0.68_{\pm0.05}$ & $\mathbf{0.72}_{\pm0.04}$ & $0.63_{\pm0.03}$ \\
  & AUC       & $0.83_{\pm0.03}$ & $0.83_{\pm0.05}$ & $\underline{0.85}_{\pm0.04}$ & $0.84_{\pm0.04}$ & $\mathbf{0.87}_{\pm0.03}$ & $0.80_{\pm0.01}$ \\
  & Bal.\ Acc & $0.51_{\pm0.02}$ & $0.52_{\pm0.06}$ & $\underline{0.57}_{\pm0.07}$ & $0.53_{\pm0.06}$ & $\mathbf{0.58}_{\pm0.04}$ & $0.50_{\pm0.05}$ \\
\midrule
\multirow{3}{*}{\parbox{1.9cm}{\raggedright HNSC mRNA Subtype}}
  & F1        & $0.54_{\pm0.01}$ & $0.60_{\pm0.04}$ & $\underline{0.60}_{\pm0.06}$ & $0.59_{\pm0.03}$ & $\mathbf{0.62}_{\pm0.07}$ & $0.55_{\pm0.03}$ \\
  & AUC       & $0.75_{\pm0.03}$ & $0.77_{\pm0.03}$ & $\underline{0.78}_{\pm0.07}$ & $0.77_{\pm0.03}$ & $\mathbf{0.79}_{\pm0.02}$ & $0.72_{\pm0.01}$ \\
  & Bal.\ Acc & $0.55_{\pm0.04}$ & $\underline{0.61}_{\pm0.07}$ & $0.59_{\pm0.07}$ & $0.58_{\pm0.05}$ & $\mathbf{0.61}_{\pm0.08}$ & $0.54_{\pm0.04}$ \\
\midrule
\multirow{3}{*}{\parbox{1.9cm}{\raggedright UCEC Genomic Subtype}}
  & F1        & $0.55_{\pm0.04}$ & $0.57_{\pm0.04}$ & $\underline{0.63}_{\pm0.03}$ & $0.57_{\pm0.04}$ & $\mathbf{0.66}_{\pm0.04}$ & $0.56_{\pm0.05}$ \\
  & AUC       & $0.75_{\pm0.03}$ & $0.76_{\pm0.04}$ & $\underline{0.81}_{\pm0.04}$ & $0.75_{\pm0.02}$ & $\mathbf{0.82}_{\pm0.03}$ & $0.74_{\pm0.04}$ \\
  & Bal.\ Acc & $0.53_{\pm0.04}$ & $0.54_{\pm0.04}$ & $\underline{0.60}_{\pm0.04}$ & $0.55_{\pm0.03}$ & $\mathbf{0.62}_{\pm0.03}$ & $0.52_{\pm0.05}$ \\
\midrule
\multirow{3}{*}{\parbox{1.9cm}{\raggedright Synaptophysin Grade}}
  & F1        & $0.78_{\pm0.08}$ & $0.79_{\pm0.06}$ & $0.76_{\pm0.05}$ & $\mathbf{0.81}_{\pm0.01}$ & $\underline{0.80}_{\pm0.06}$ & $0.79_{\pm0.08}$ \\
  & AUC       & $0.61_{\pm0.19}$ & $\mathbf{0.74}_{\pm0.08}$ & $0.50_{\pm0.09}$ & $\underline{0.68}_{\pm0.08}$ & $0.64_{\pm0.11}$ & $0.61_{\pm0.15}$ \\
  & Bal.\ Acc & $0.64_{\pm0.09}$ & $\mathbf{0.69}_{\pm0.08}$ & $0.58_{\pm0.03}$ & $0.65_{\pm0.08}$ & $\underline{0.68}_{\pm0.04}$ & $0.65_{\pm0.08}$ \\
\midrule
\multirow{3}{*}{\parbox{1.9cm}{\raggedright RAS/BRAF Status}}
  & F1        & $0.89_{\pm0.07}$ & $0.89_{\pm0.07}$ & $0.84_{\pm0.11}$ & $0.88_{\pm0.08}$ & $\underline{0.90}_{\pm0.07}$ & $\mathbf{0.94}_{\pm0.08}$ \\
  & AUC       & $0.68_{\pm0.14}$ & $0.48_{\pm0.28}$ & $0.56_{\pm0.16}$ & $0.33_{\pm0.20}$ & $\mathbf{0.80}_{\pm0.09}$ & $\underline{0.75}_{\pm0.26}$ \\
  & Bal.\ Acc & $0.68_{\pm0.20}$ & $0.68_{\pm0.18}$ & $0.57_{\pm0.22}$ & $0.66_{\pm0.19}$ & $\underline{0.68}_{\pm0.19}$ & $\mathbf{0.85}_{\pm0.20}$ \\
\midrule
\multirow{3}{*}{\parbox{1.9cm}{\raggedright Keratinizing SCC Grade}}
  & F1        & $\mathbf{0.74}_{\pm0.02}$ & $0.68_{\pm0.06}$ & $0.72_{\pm0.04}$ & $0.69_{\pm0.02}$ & $0.72_{\pm0.03}$ & $\underline{0.73}_{\pm0.03}$ \\
  & AUC       & $\mathbf{0.75}_{\pm0.04}$ & $0.69_{\pm0.07}$ & $\underline{0.74}_{\pm0.02}$ & $0.72_{\pm0.04}$ & $0.72_{\pm0.05}$ & $0.72_{\pm0.06}$ \\
  & Bal.\ Acc & $\mathbf{0.73}_{\pm0.02}$ & $0.68_{\pm0.06}$ & $0.72_{\pm0.03}$ & $0.69_{\pm0.02}$ & $0.71_{\pm0.03}$ & $\underline{0.72}_{\pm0.03}$ \\
\midrule
\multirow{3}{*}{\parbox{1.9cm}{\raggedright Non-Keratinizing SCC Grade}}
  & F1        & $0.80_{\pm0.12}$ & $0.78_{\pm0.12}$ & $0.75_{\pm0.17}$ & $\underline{0.80}_{\pm0.11}$ & $\mathbf{0.81}_{\pm0.12}$ & $0.77_{\pm0.14}$ \\
  & AUC       & $\underline{0.66}_{\pm0.12}$ & $0.61_{\pm0.03}$ & $0.52_{\pm0.20}$ & $0.66_{\pm0.11}$ & $\mathbf{0.71}_{\pm0.11}$ & $0.66_{\pm0.07}$ \\
  & Bal.\ Acc & $\mathbf{0.79}_{\pm0.15}$ & $0.75_{\pm0.13}$ & $0.72_{\pm0.17}$ & $\underline{0.79}_{\pm0.12}$ & $0.78_{\pm0.14}$ & $0.73_{\pm0.15}$ \\
\midrule
\multirow{3}{*}{\parbox{1.9cm}{\raggedright Primary vs.\ Metastasis}}
  & F1        & $0.92_{\pm0.01}$ & $0.92_{\pm0.01}$ & $0.92_{\pm0.01}$ & $\underline{0.93}_{\pm0.02}$ & $0.93_{\pm0.03}$ & $\mathbf{0.93}_{\pm0.02}$ \\
  & AUC       & $0.62_{\pm0.05}$ & $0.60_{\pm0.12}$ & $0.66_{\pm0.10}$ & $\underline{0.75}_{\pm0.07}$ & $\mathbf{0.77}_{\pm0.03}$ & $0.69_{\pm0.13}$ \\
  & Bal.\ Acc & $0.57_{\pm0.02}$ & $0.58_{\pm0.05}$ & $0.63_{\pm0.09}$ & $0.63_{\pm0.05}$ & $\underline{0.64}_{\pm0.08}$ & $\mathbf{0.65}_{\pm0.10}$ \\
\bottomrule
\end{tabular}%
}
\end{table}

{\textbf{Parameter Efficiency.}}
MOOZY totals 85.77M parameters (64.10M slide and case encoder + 21.67M patch encoder), making it 4--14$\times$ smaller than competing encoders (GigaPath 1.22B, PRISM 742M, Madeleine 400M, TITAN 355M). CHIEF is smaller in absolute size but trails MOOZY on all three macro metrics. This efficiency stems from concentrating capacity at the slide level and reusing a lightweight public patch encoder. A detailed breakdown is in \Cref{tab:encoder_parameter_comparison}.

{\textbf{Comparison with MILs.}}
We compare MOOZY against five patch encoders in the MIL setting. Each encoder is paired with five MIL architectures (MeanMIL, ABMIL, CLAM, DSMIL, and TransMIL) following \cite{shao2025miltransfer}, and each entry is the arithmetic mean over the five architectures. Across all sixteen tasks, MOOZY leads the strongest MIL baseline, CONCH v1.5, by 0.029 weighted F1, 0.043 weighted ROC-AUC, and 0.041 balanced accuracy. Macro-average results are in \Cref{tab:mil_comparison_macro}. Task-wise and per-method breakdowns are in \Cref{appendix_sec:mil_specific_taskwise}.

\begin{table}[tb]
\caption{Macro-average MIL comparison across sixteen held-out tasks. Each patch-encoder entry averages over five MIL architectures (MeanMIL, ABMIL, CLAM, DSMIL, TransMIL).}
\label{tab:mil_comparison_macro}
\centering
\small
\setlength{\tabcolsep}{3pt}
\renewcommand{\arraystretch}{0.85}
\resizebox{\linewidth}{!}{%
\begin{tabular}{lcccccc}
\toprule
Metric & Backbone & \shortstack{UNI\\v2} & \shortstack{Phikon\\v2} & \shortstack{CONCH\\v1.5} & MUSK & \shortstack{MOOZY\\(Ours)} \\
\midrule
F1 (weighted)      & $0.723$ & $0.722$ & $0.714$ & $\underline{0.740}$ & $0.720$ & $\mathbf{0.769}$ \\
ROC-AUC (weighted) & $0.707$ & $0.707$ & $0.697$ & $\underline{0.720}$ & $0.695$ & $\mathbf{0.763}$ \\
Balanced Acc       & $0.643$ & $0.637$ & $0.625$ & $\underline{0.661}$ & $0.637$ & $\mathbf{0.702}$ \\
\bottomrule
\end{tabular}%
}
\end{table}

{\textbf{Multi-stage Ablation.}}
\Cref{tab:multistage_ablation_summary} reports macro-average results for five configurations. With the case aggregator held out, adding SSL pretraining improves all three metrics. Stage~2 only without the case aggregator reaches 0.743 weighted F1, 0.721 weighted ROC-AUC, and 0.664 balanced accuracy, which MOOZY without the case aggregator raises to 0.749, 0.737, and 0.682. The case aggregator further raises these values to 0.769, 0.763, and 0.702. When trained from scratch without SSL, adding the case aggregator reduces performance (Stage~2 only at 0.731/0.697/0.659). Task-wise breakdowns are in \Cref{appendix_sec:stage1_moozy_comparison,tab:stage1_moozy_taskwise}, \Cref{appendix_sec:stage2_moozy_comparison,tab:stage2_moozy_taskwise}, and \Cref{appendix_sec:stage2_nocaseagg_moozy_comparison,tab:stage2_nocaseagg_moozy_taskwise}.

{\textbf{Case Aggregator Ablation.}}
We ablate the case aggregator by comparing MOOZY without the case aggregator (Stage~2 slide encoder with mean slide pooling)\footnote{The case aggregator is trained as part of MOOZY and discarded only at inference, where per-slide embeddings are mean-pooled instead.} against full MOOZY (\Cref{tab:multistage_ablation_summary}). The aggregator raises macro weighted F1 from 0.749 to 0.769, AUC from 0.737 to 0.763, and balanced accuracy from 0.682 to 0.702. Task-wise, F1 improves on 14 of 16 tasks, AUC on 12, and balanced accuracy on 13. Full task-wise results are provided in \Cref{appendix_sec:case_aggregator_ablation,tab:meanpool_moozy_taskwise}.

\begin{table}[tb]
\caption{Unified macro-average ablation across sixteen held-out tasks. The table includes Stage~1 only, Stage~2 only with and without the case aggregator, MOOZY without the case aggregator (mean slide pooling), and full MOOZY.}
\label{tab:multistage_ablation_summary}
\centering
\scriptsize
\setlength{\tabcolsep}{2.5pt}
\setlength{\aboverulesep}{0.08ex}
\setlength{\belowrulesep}{0.08ex}
\renewcommand{\arraystretch}{0.85}
\resizebox{\linewidth}{!}{%
\begin{tabular}{lccc ccc}
\toprule
Setting & Stage~1 & Stage~2 & Case Agg. & F1 & AUC & Bal.\ Acc \\
\midrule
Stage~1 only & \cmark & \xmark & \xmark & $0.743$ & $0.715$ & $0.662$ \\
Stage~2 only w/o case agg. & \xmark & \cmark & \xmark & $0.743$ & $0.721$ & $0.664$ \\
Stage~2 only & \xmark & \cmark & \cmark & $0.731$ & $0.697$ & $0.659$ \\
MOOZY w/o case agg. & \cmark & \cmark & \xmark & $0.749$ & $0.737$ & $0.682$ \\
MOOZY & \cmark & \cmark & \cmark & $0.769$ & $0.763$ & $0.702$ \\
\bottomrule
\end{tabular}
}
\end{table}

{\textbf{Attention Map Analysis.}}
To examine where each encoder concentrates on the slide, a board-certified pathologist reviewed attention maps for 20 representative WSIs sampled from eight held-out evaluation cohorts and five encoders. Two scores were assigned per image: the shift score (1–5, where 1 is cancer-focused, 3 is balanced, and 5 is non-cancer-focused) and the semantic-gap score (1–3, where 1 indicates rare and 3 indicates frequent gaps in diagnostically relevant tissue). MOOZY achieved the lowest mean gap score (1.00), followed by TITAN (1.38) and PRISM (1.75). On shift, MOOZY was near balanced (2.63) and TITAN closest to exact balance (3.13), while PRISM (2.38) were more cancer-biased. These two scores suggest that MOOZY attends broadly while maintaining balanced focus between tumor and surrounding tissue. A plausible explanation comes from the two-stage objective. In Stage~1, masked-region prediction and agreement across global and local views encourage the model to use distributed contextual evidence instead of relying on a few high-response hotspots. In Stage~2, supervision across diverse endpoints and cohorts favors features that stay informative in both malignant and non-malignant tissue contexts. A representative example is shown in \Cref{fig:attention_maps_main}. Heatmap generation details and additional examples are provided in \Cref{appendix_sec:attention_maps_generation,appendix_sec:attention_maps}.

\begin{figure*}[t]
  \centering
  \includegraphics[width=0.82\linewidth]{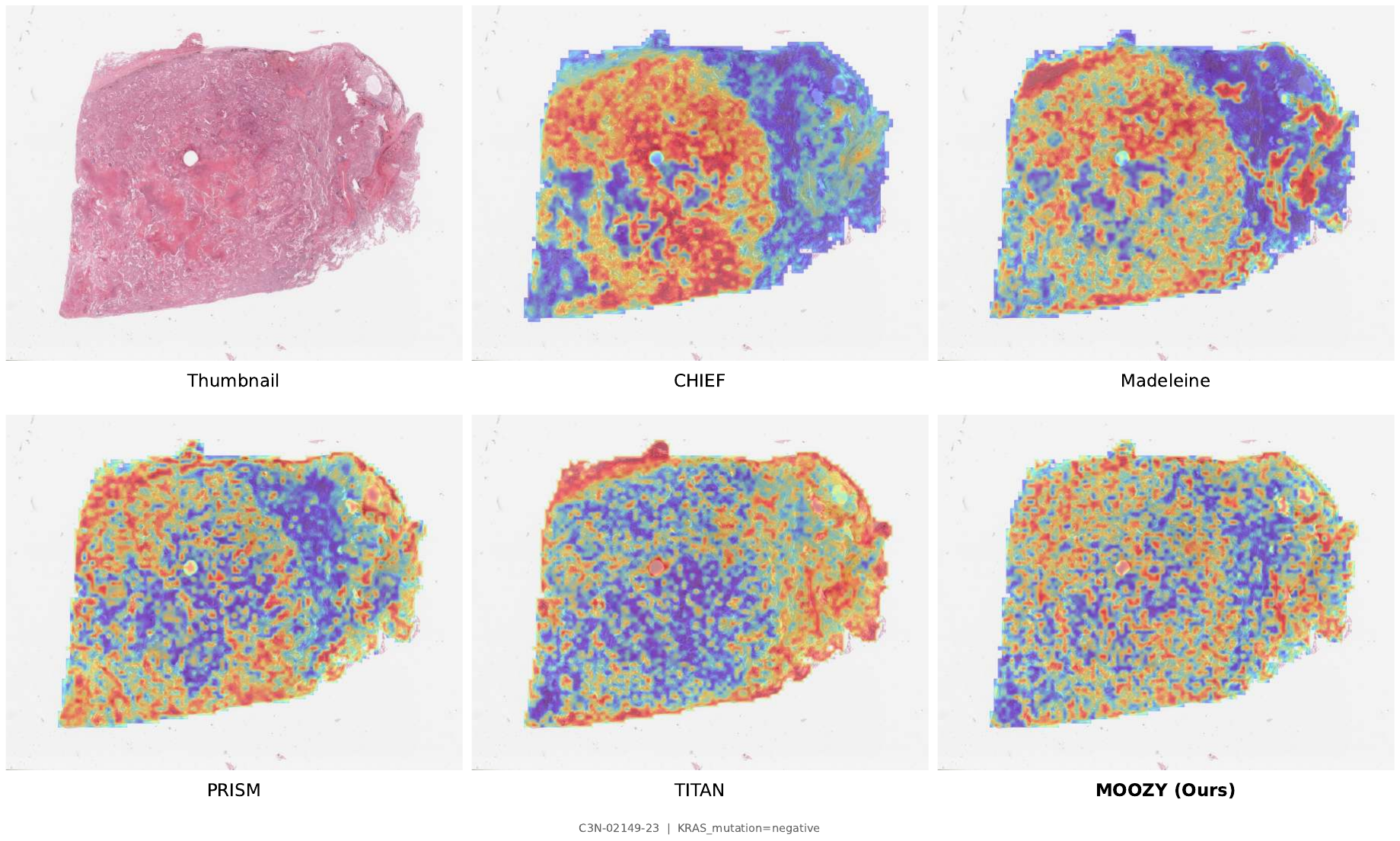}
  \captionsetup{skip=2pt}
  \caption{Attention-map comparison on a lung adenocarcinoma slide. MOOZY and TITAN: balanced, comprehensive coverage (shift~3, gap~1). PRISM: balanced shift with moderate gaps (shift~3, gap~2). CHIEF and Madeleine: cancer-biased with frequent semantic gaps (shift~2, gap~3).}
  \label{fig:attention_maps_main}
\end{figure*}

{\textbf{Qualitative Analysis.}}
To complement quantitative results, we visualize UMAP~\cite{mcinnes2018umap} embeddings for three representative tasks (CPTAC cancer type, pan-dataset anatomical site, and TCGA cancer type) across four slide encoders (\Cref{fig:embedding_qualitative_umap}). MOOZY shows the clearest separation on CPTAC and TCGA cancer type, TITAN is close behind, and Madeleine/PRISM show more overlap. On anatomical site, TITAN is strongest, with MOOZY and Madeleine comparable and PRISM weaker. t-SNE~\cite{vandermaaten2008tsne} and PCA stability diagnostics confirm the same pattern (\Cref{appendix_sec:embedding_visualization_config,fig:embedding_qualitative_tsne,appendix_sec:pca_stability_math}).

\providecommand{\embcell}[1]{%
  \includegraphics[width=\linewidth,trim=6 6 6 6,clip]{#1}%
}
\providecommand{\embcellfirst}[1]{%
  \includegraphics[width=\linewidth,trim=16 6 6 6,clip]{#1}%
}
\providecommand{\embcolhdr}[1]{{\fontsize{6.2}{6.8}\selectfont #1}}
\providecommand{\embrowhdr}[1]{%
  \parbox[c]{\linewidth}{\raggedright\fontsize{6.2}{6.8}\selectfont #1}%
}

\begin{figure*}[!t]
  \centering
  \setlength{\tabcolsep}{0pt}
  \renewcommand{\arraystretch}{0.84}
  \begin{tabular}{>{\raggedright\arraybackslash}m{0.105\textwidth}*{4}{>{\centering\arraybackslash}m{0.170\textwidth}}}
    & \embcolhdr{MOOZY} & \embcolhdr{TITAN} & \embcolhdr{Madeleine} & \embcolhdr{PRISM} \\
    \embrowhdr{CPTAC cancer type} &
      \embcellfirst{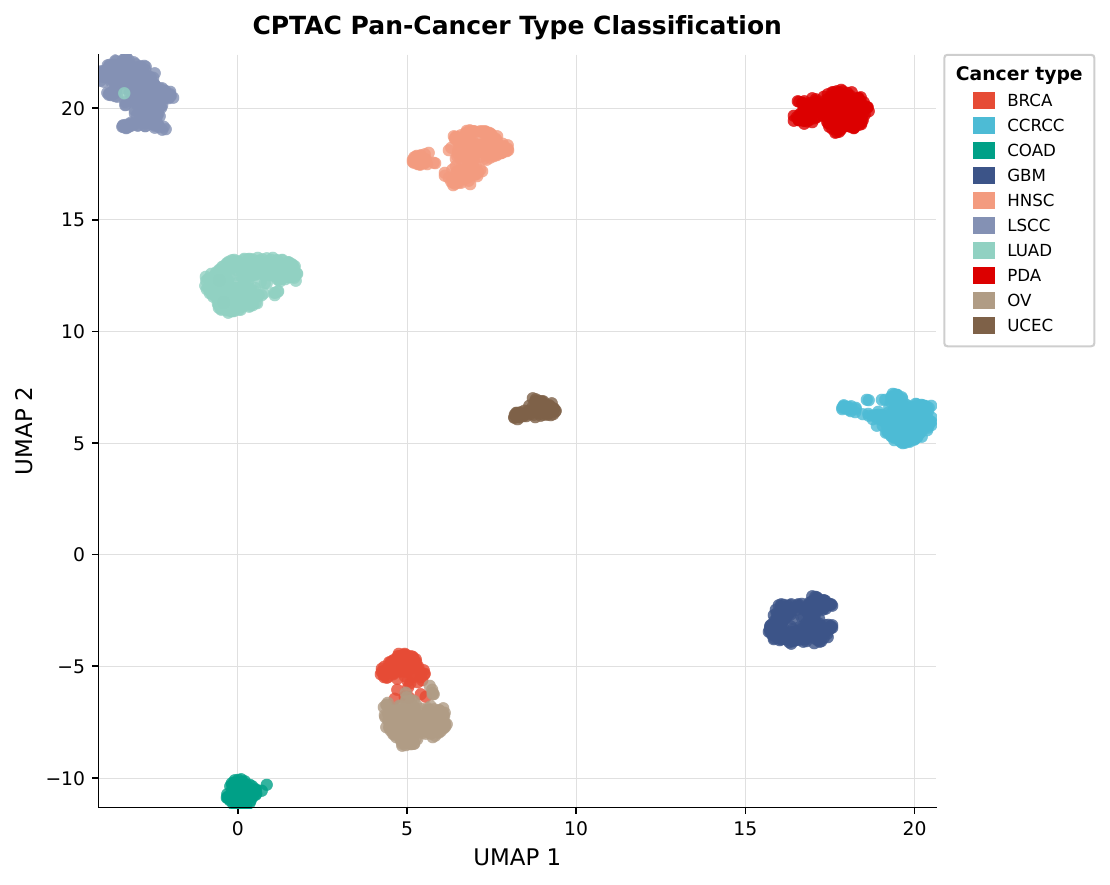} &
      \embcell{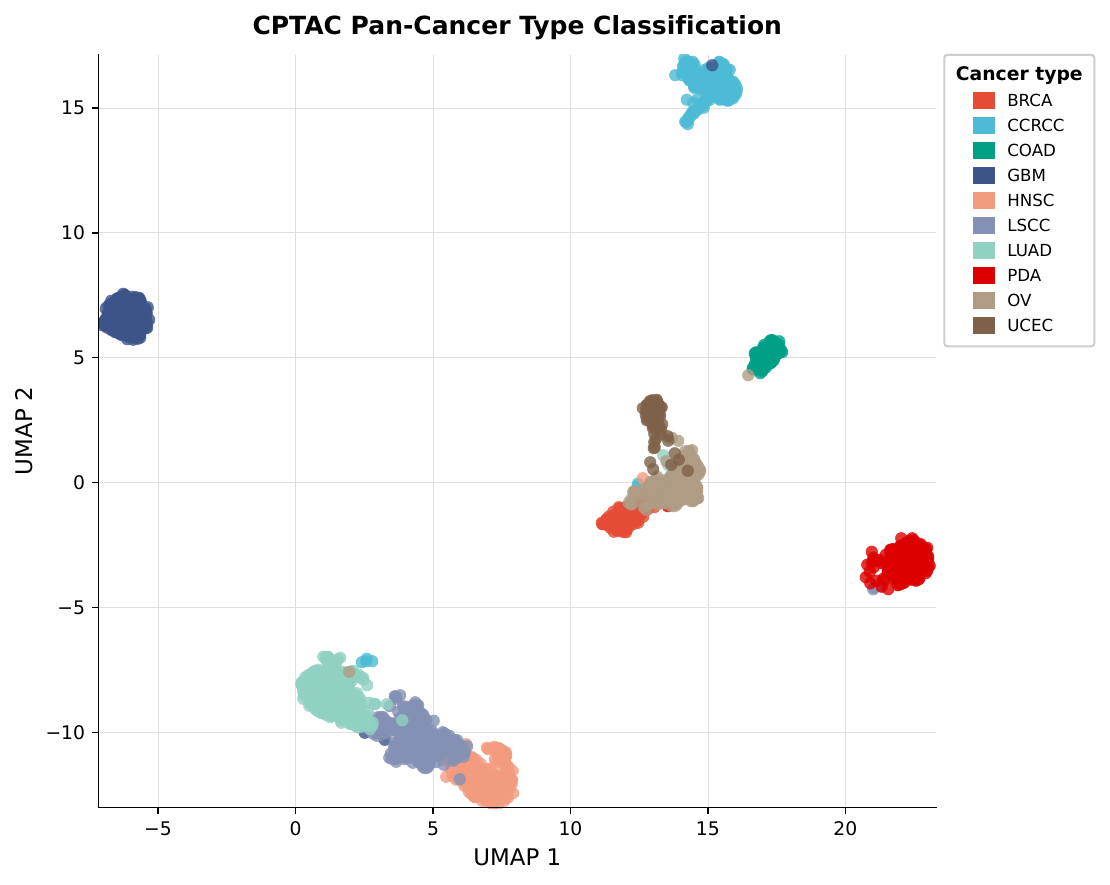} &
      \embcell{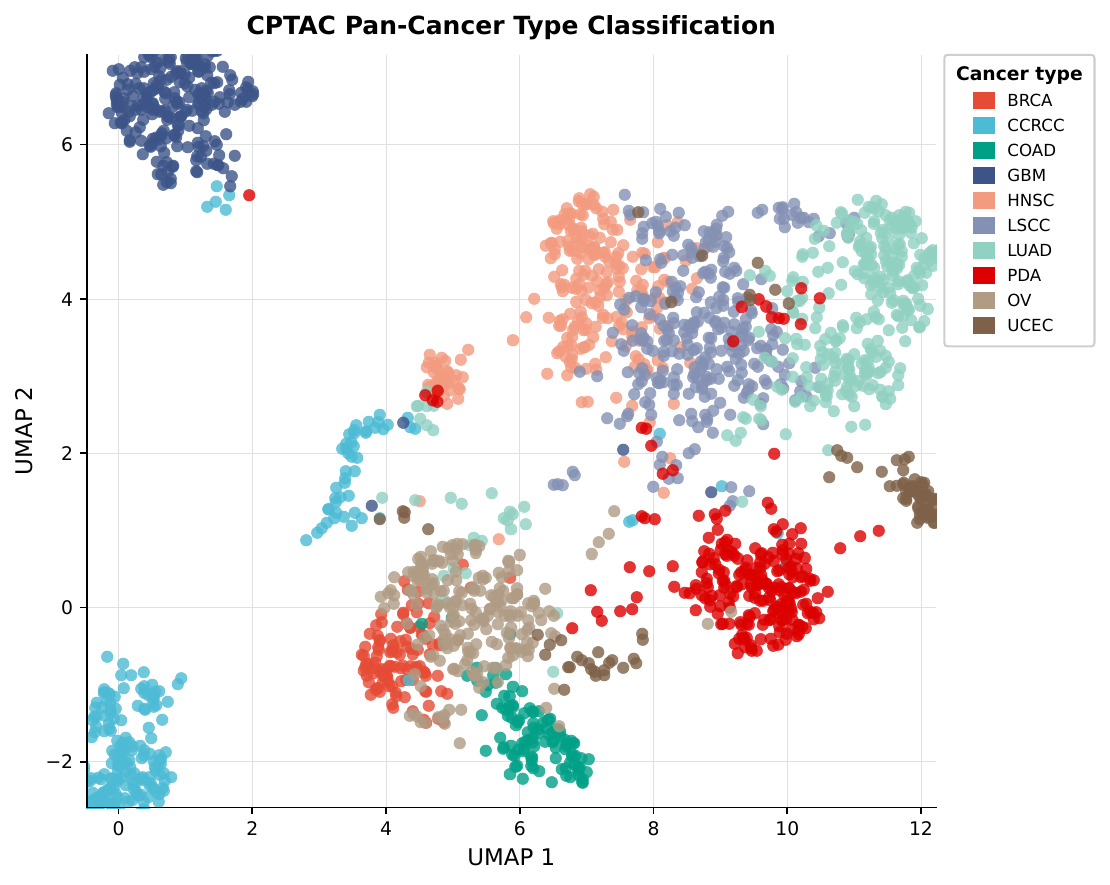} &
      \embcell{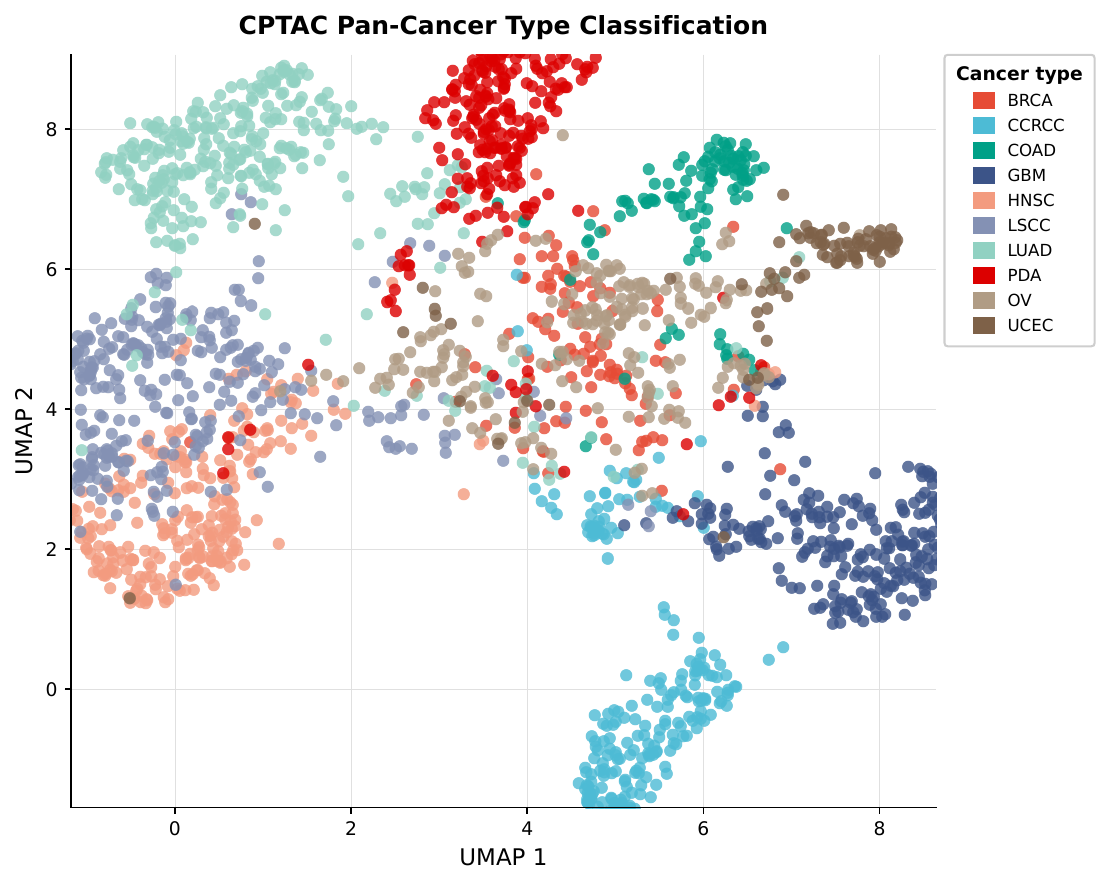} \\
    \embrowhdr{Anatomical site} &
      \embcellfirst{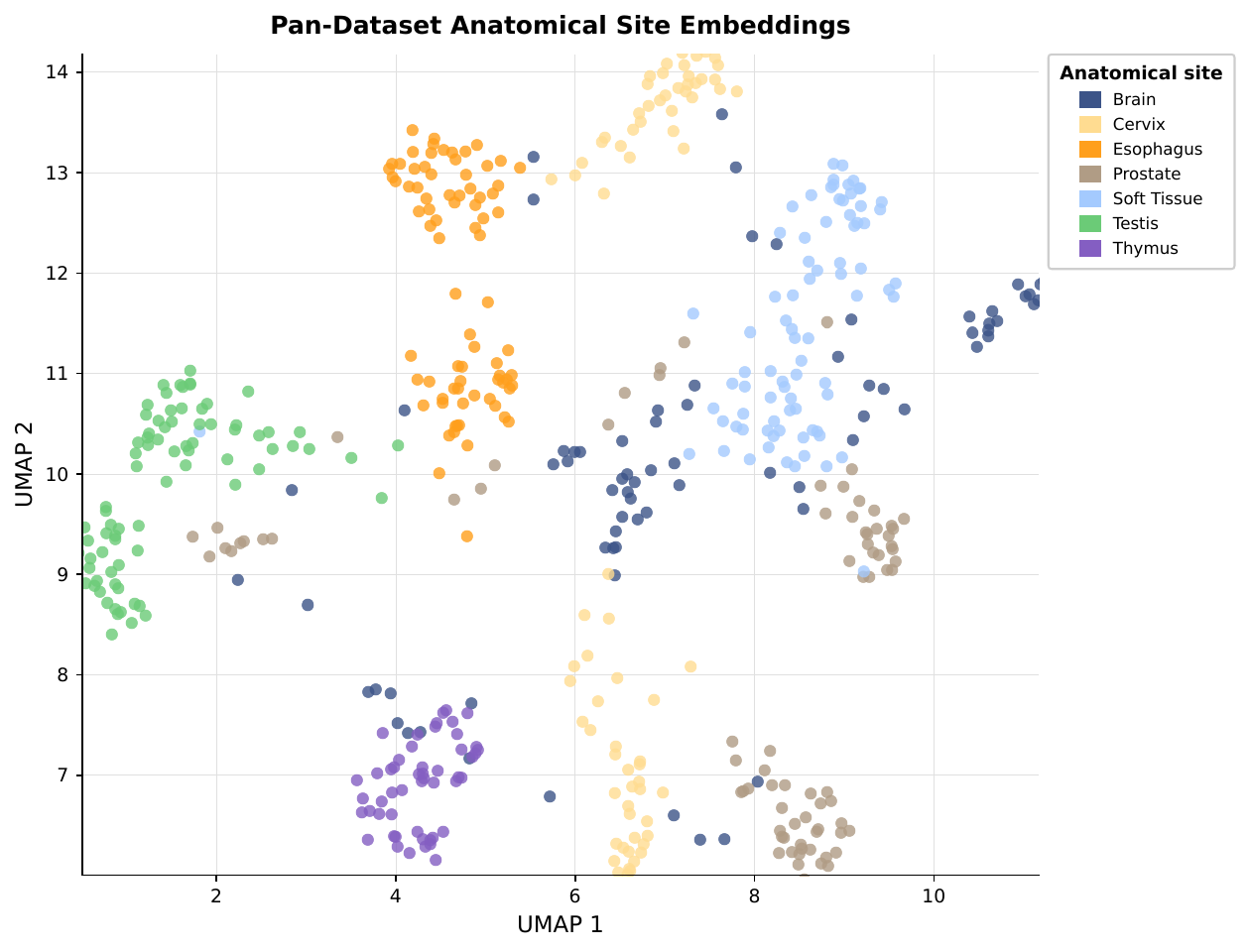} &
      \embcell{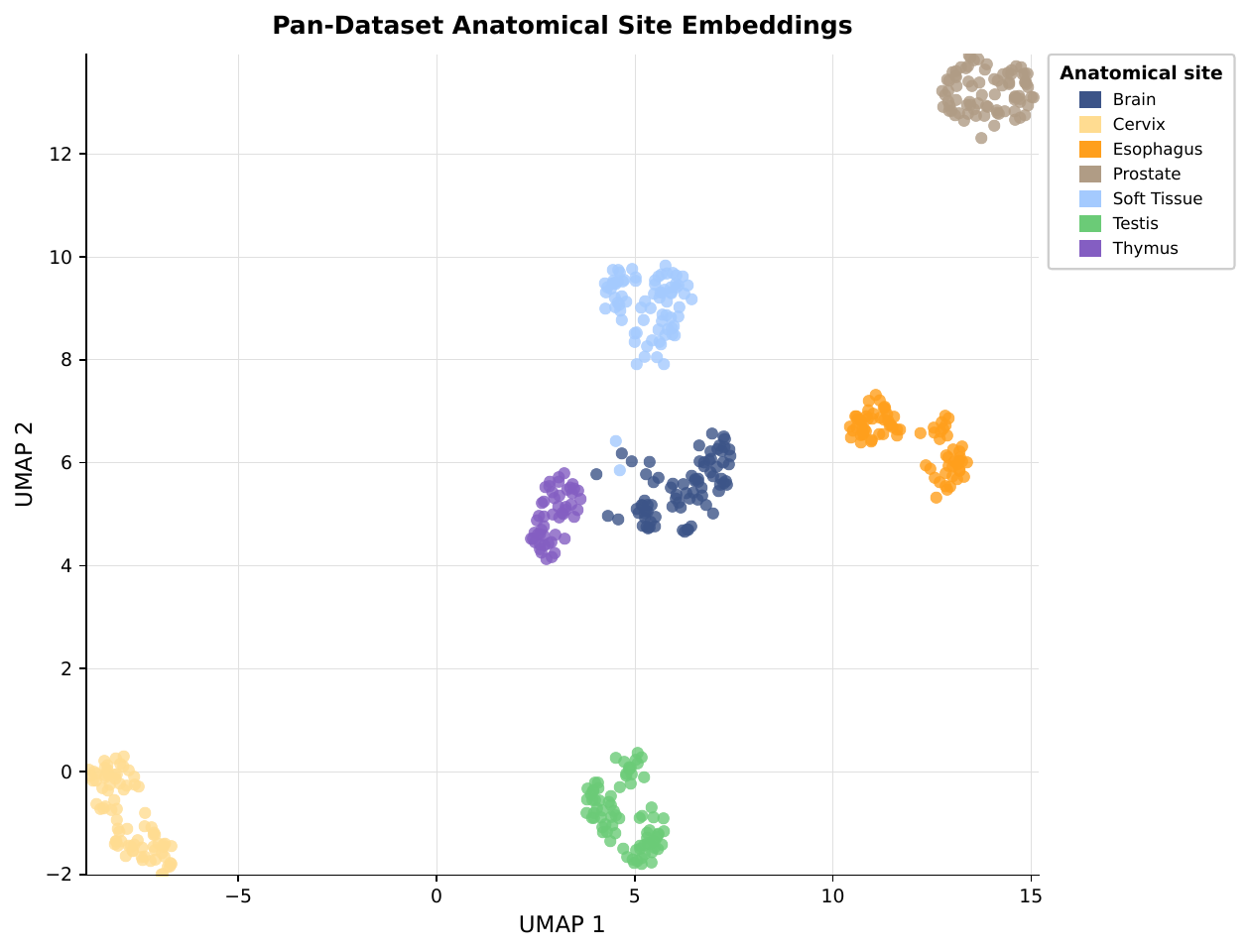} &
      \embcell{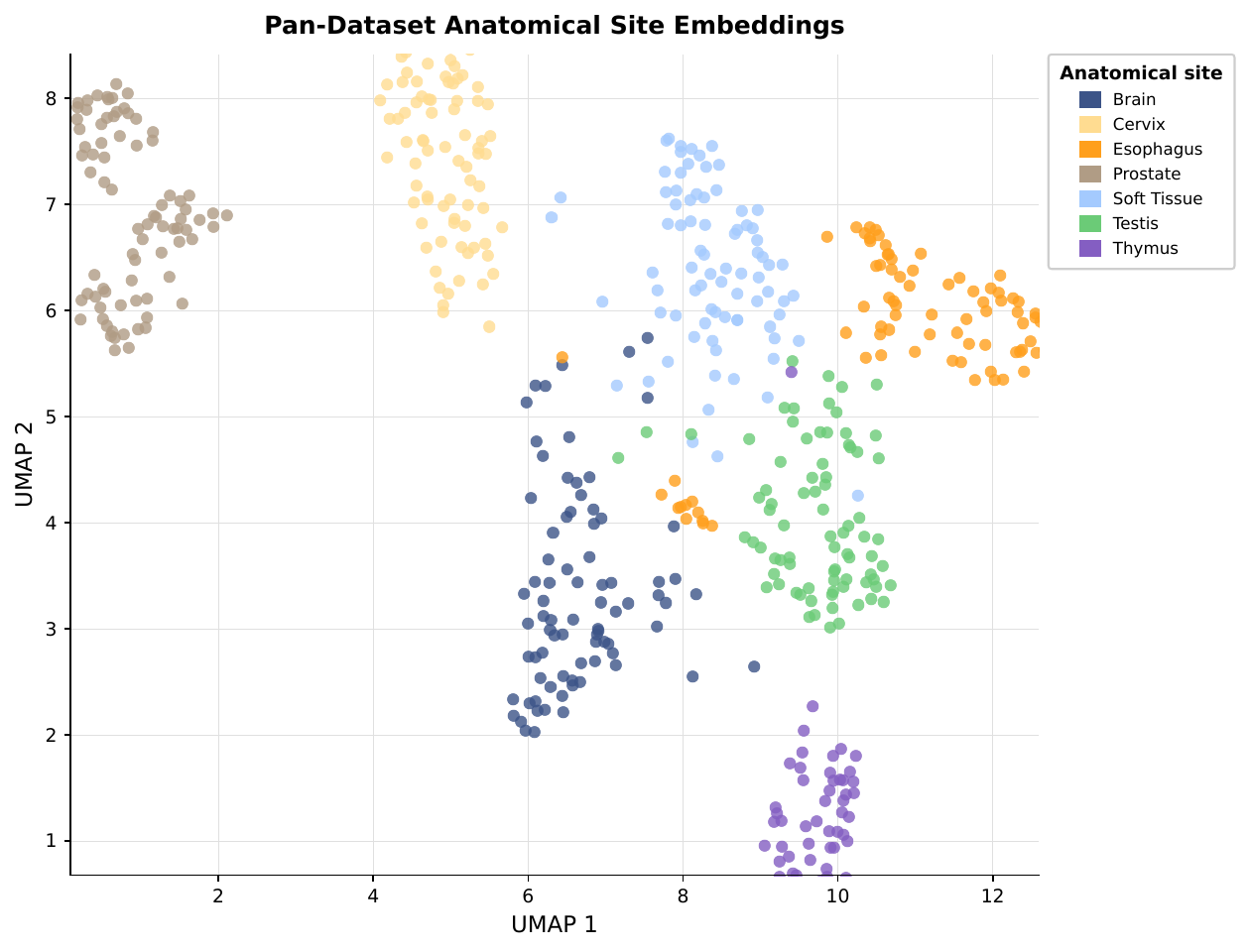} &
      \embcell{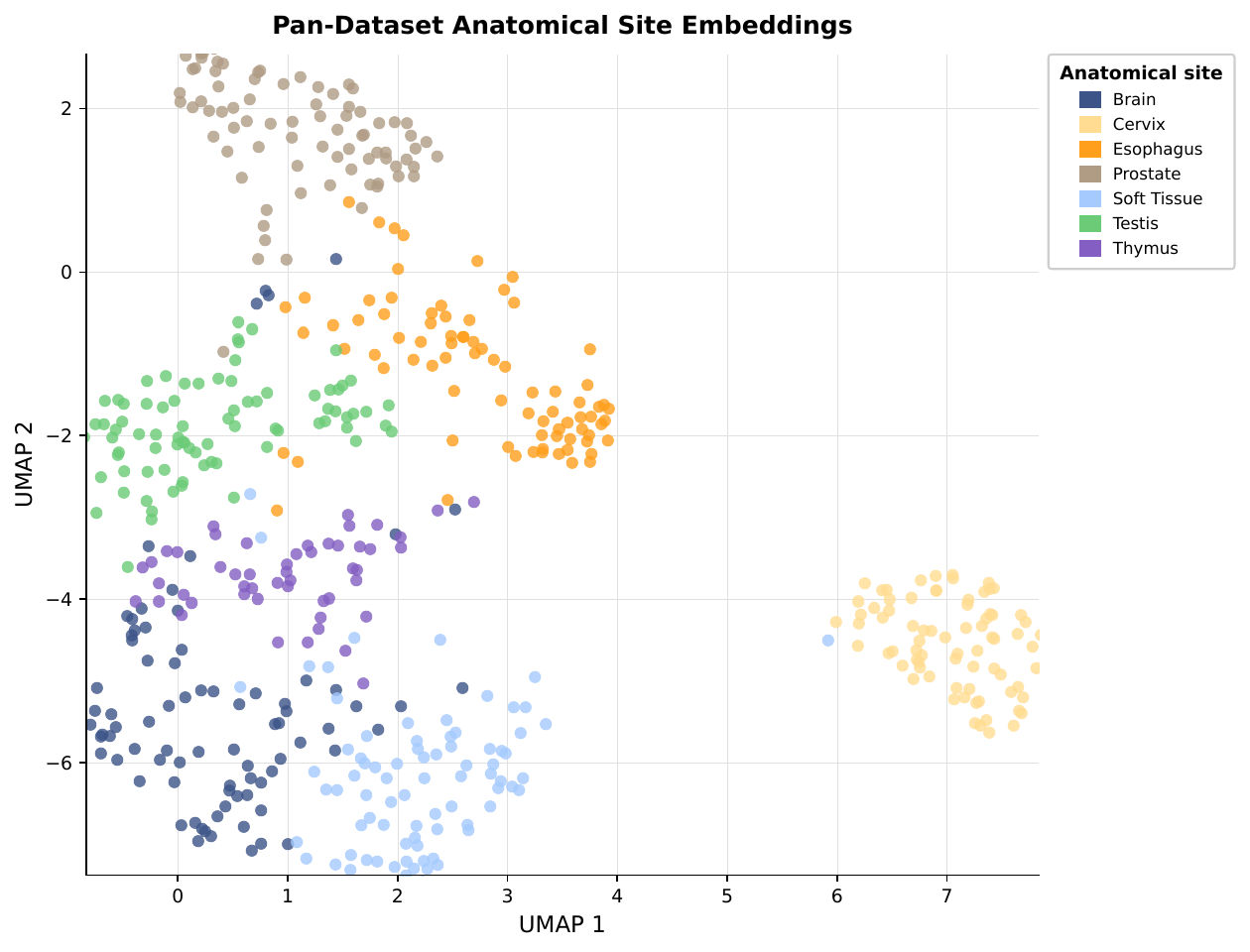} \\
    \embrowhdr{TCGA cancer type} &
      \embcellfirst{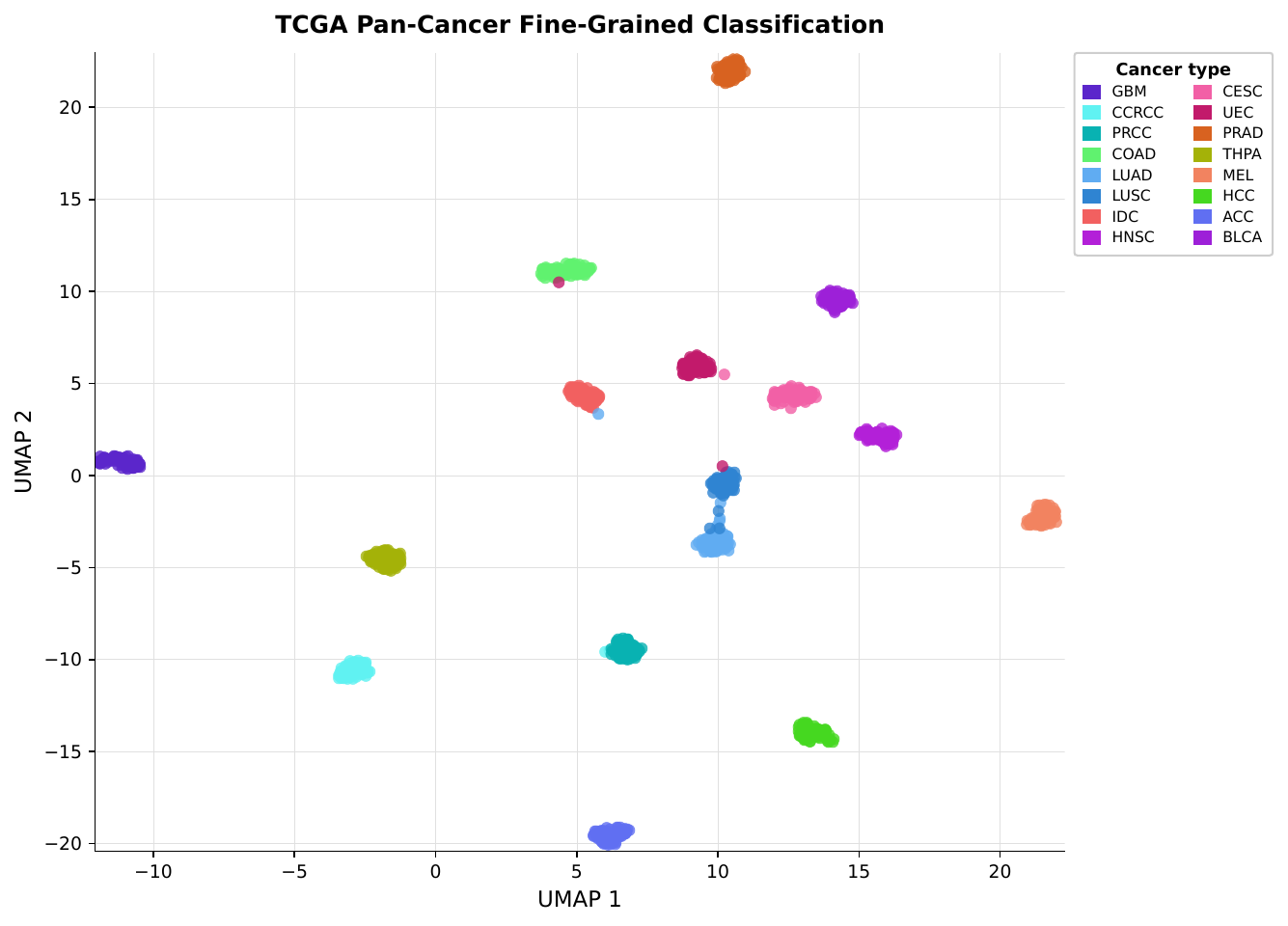} &
      \embcell{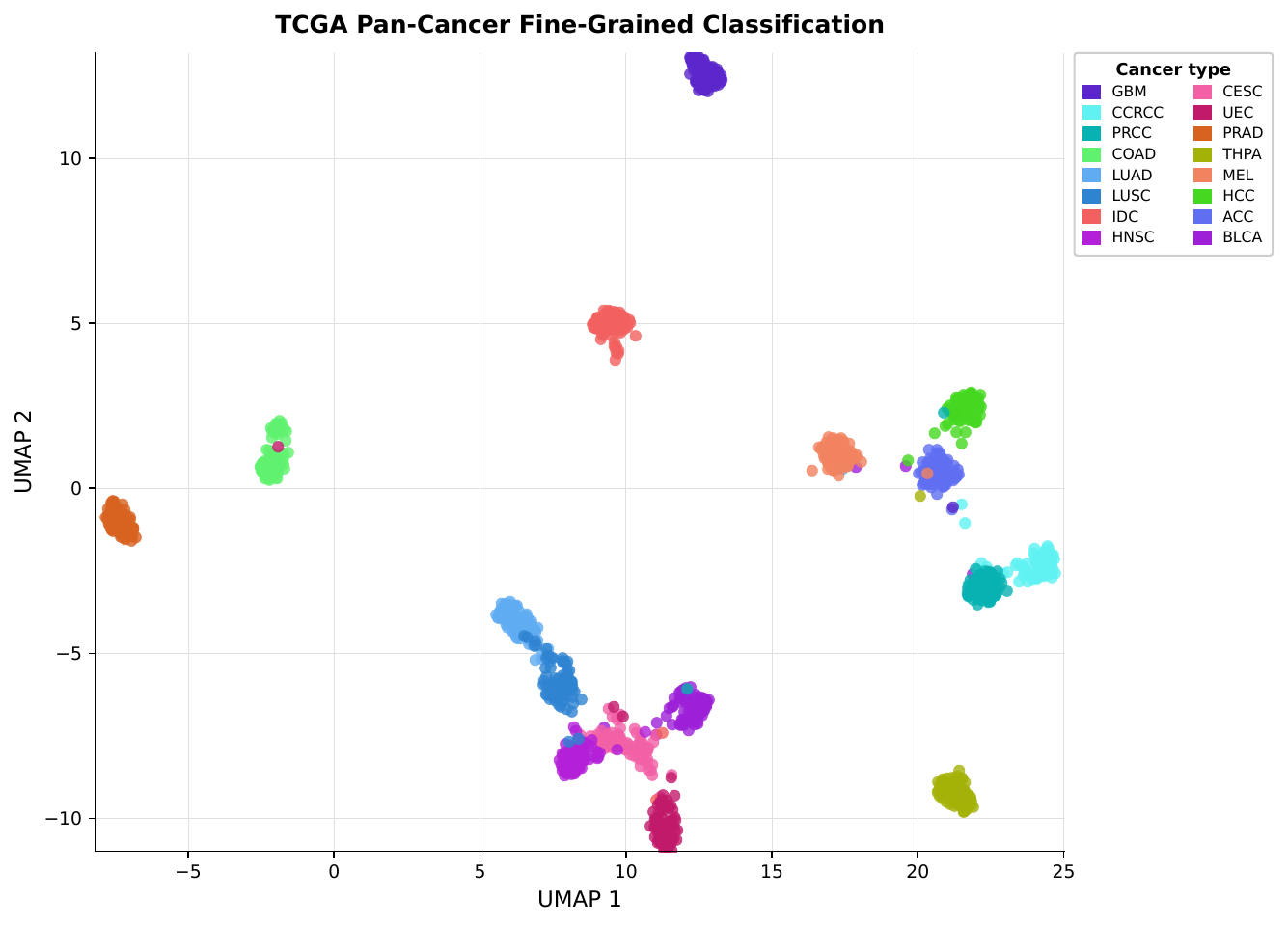} &
      \embcell{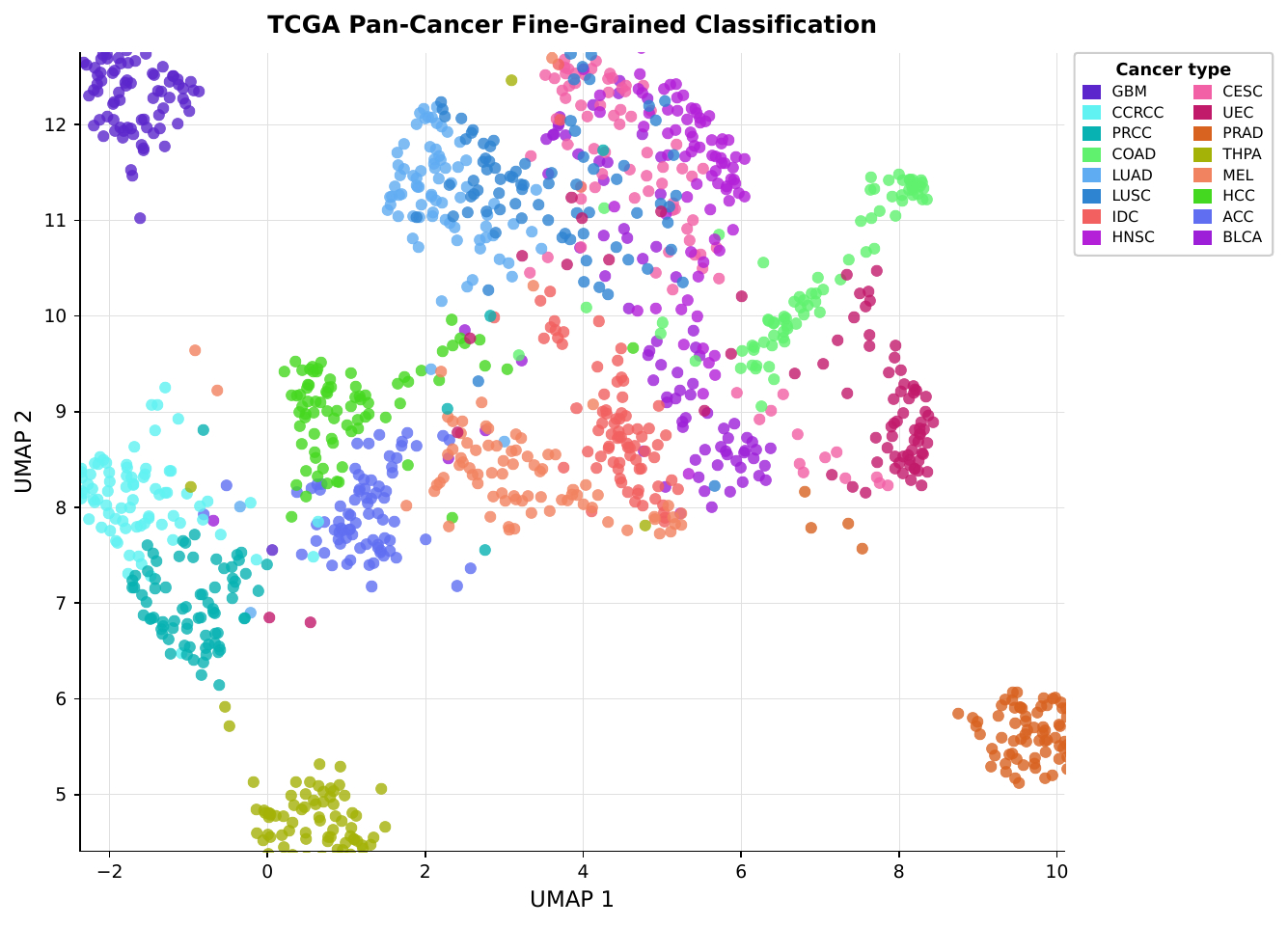} &
      \embcell{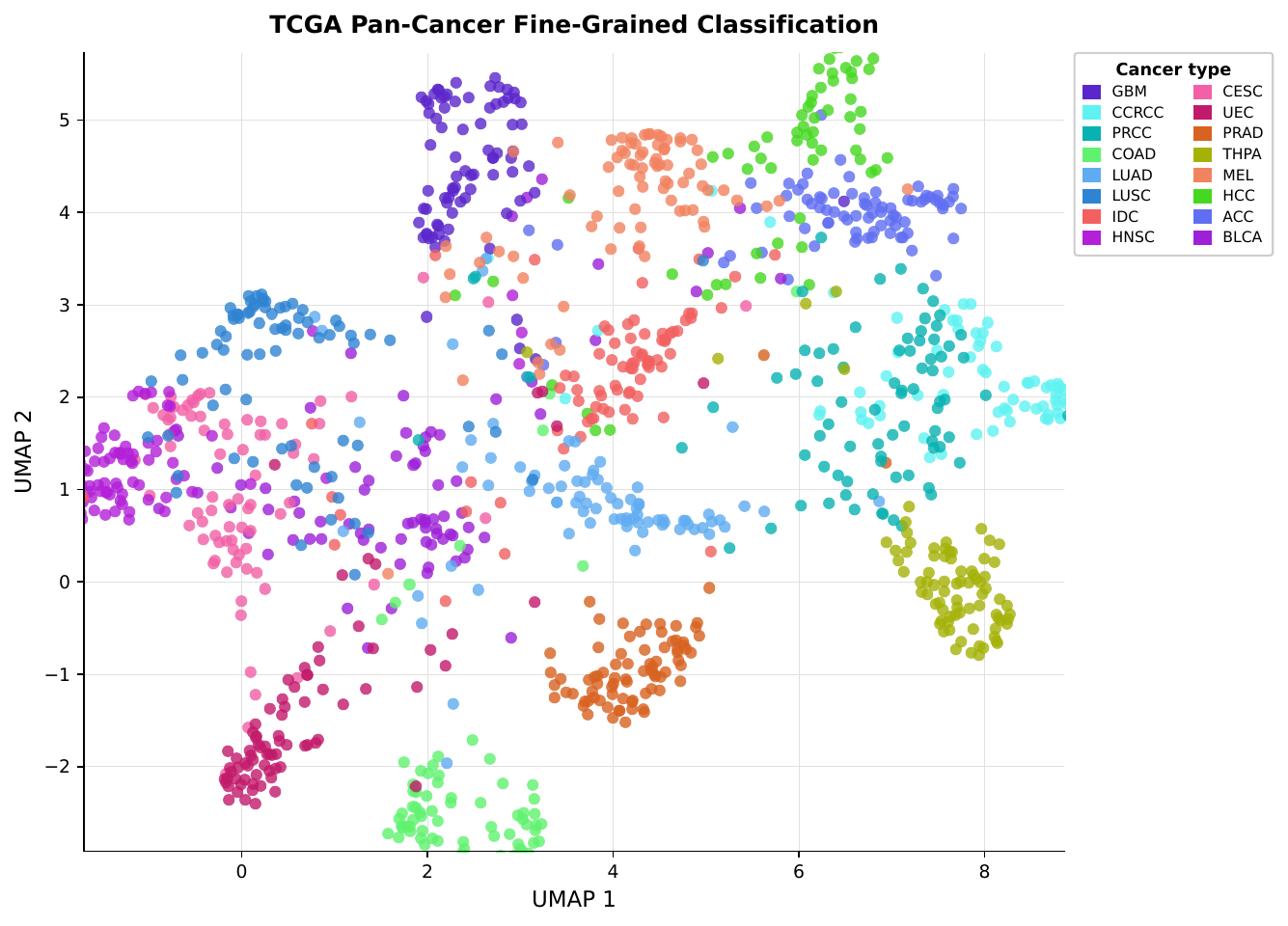} \\
  \end{tabular}
  \captionsetup{skip=2pt}
  \caption{UMAP qualitative comparison across four slide encoders (columns) and three tasks (rows), using matched class-balanced sampling and identical reduction settings.}
  \label{fig:embedding_qualitative_umap}
\end{figure*}

\section{Conclusion}
\label{sec:conclusion}

We presented MOOZY, a two-stage patient-first framework that decouples vision-only SSL pretraining from case-aware multi-task alignment replacing the naive early/late multi-slide fusion with explicit case-level inter-slide dependency modeling, showing competitive representations without proprietary slides, paired clinical reports, or billion-parameter architectures. Across sixteen held-out tasks, MOOZY achieves the strongest macro weighted F1 and balanced accuracy among slide encoders and exceeds all trained MIL baselines across the three macro metrics, while using fewer parameters than the compared slide encoders. Like other slide encoders, MOOZY targets global case-level predictions, so dense or segmentation predictions are not supported. The benefit of explicit cross-slide aggregation also remains to be quantified on tasks specifically designed to require deeper multi-slide reasoning, alongside natural extensions toward case-level retrieval, slide-report co-training, and genomic fusion.

\section*{Acknowledgements}
This work was supported by NSERC-DG RGPIN-2022-05378 [M.S.H], Amazon Research Award [M.S.H], and Gina Cody RIF [M.S.H], FRQNT scholarship [Y.K]. Computational resources were provided in part by Calcul Qu\'{e}bec (\url{www.calculquebec.ca}) and the Digital Research Alliance of Canada (\url{www.alliancecan.ca}).

\clearpage
\bibliographystyle{splncs04}
\bibliography{main}
\newpage

\appendix

\section{Dataset Statistics}
\label[appendix]{appendix_sec:dataset_statistics}

\Cref{tab:patches_extracted_appendix} reports slide counts and extracted patch totals at $20\times$ and $40\times$ magnifications after segmenting tissues from the collected slides. \Cref{tab:class_distribution_appendix} reports class-cardinality across classification tasks, where most tasks are low-cardinality and a smaller subset has higher-cardinality label spaces.

\begin{table}[tb]
  \caption{Total number of patches extracted from the full dataset at each magnification level using non-overlapping $224 \times 224$ tiling.}
  \label{tab:patches_extracted_appendix}
  \centering
  \begin{tabular}{c|c|c}
    \toprule
    Magnification Level & Number of Slides & Number of Patches \\
    \midrule
    20X & 53,286 & 449,943,195 \\
    40X & 23,848 & 1,224,330,326 \\
    \midrule
    Total & 77,134 & 1,674,273,521 \\
    \bottomrule
  \end{tabular}
\end{table}

\begin{table}[tb]
\centering
\caption{Distribution of the number of classes (\eg, labels) per classification task.}
\label{tab:class_distribution_appendix}
\setlength{\tabcolsep}{6pt}
\begin{tabular}{l @{\hspace{10pt}} cccccccccccc}
\toprule
\textbf{Number of Classes} & 2 & 3 & 4 & 5 & 6 & 7 & 8 & 9 & 10 & 12 & 30 & 46 \\
\midrule
\textbf{Number of Tasks}   & 148 & 36 & 5 & 5 & 1 & 2 & 1 & 2 & 2 & 1 & 1 & 1 \\
\bottomrule
\end{tabular}
\end{table}

\section{Task Preparation}
\label[appendix]{appendix_sec:task_preparation}

\subsection{Training Task Distribution by Anatomical Site and Category}
\label[appendix]{appendix_sec:organ_task_distribution}

\Cref{tab:organ_distribution_appendix} reports the full distribution of training tasks across anatomical sites and associated task categories.

\begin{table}[tb]
\centering
\caption{Distribution of training tasks across anatomical sites and task categories.}
\label{tab:organ_distribution_appendix}
\scriptsize
\setlength{\tabcolsep}{3pt}
\renewcommand{\arraystretch}{0.9}
\resizebox{\textwidth}{!}{%
\begin{tabular}{lrrrrp{6cm}}
\toprule
\textbf{Anatomical Site} & \textbf{Datasets} & \textbf{Tasks} & \textbf{Slides} & \textbf{Supervised Cases} & \textbf{Task Categories} \\
\midrule
Multi-organ & 3 & 5 & 21,832 & 18,702 & Cancer/Organ Classification, Malignancy Detection, Procedure Classification \\
Prostate & 4 & 11 & 13,282 & 12,823 & Grading, Survival Prediction, Treatment Response, Tumor Detection \\
Colon & 5 & 35 & 6,990 & 6,977 & Grading, Immune Classification, MSI Status, Malignancy Detection, Mutation Prediction, Subtype Classification, Survival Prediction \\
Breast & 8 & 27 & 4,524 & 4,062 & Biomarker Status, Grading, Immune Classification, Invasion Detection, Lesion Detection, Metastasis Detection, Mutation Prediction, Procedure Classification, Subtype Classification, Survival Prediction, Tumor Classification \\
Brain & 4 & 19 & 4,266 & 3,126 & Grading, Immune Classification, Mutation Prediction, Subtype Classification, Survival Prediction \\
Kidney & 6 & 27 & 3,037 & 2,853 & Grading, Immune Classification, Mutation Prediction, Subtype Classification, Survival Prediction \\
Lung & 8 & 33 & 2,801 & 2,280 & Histologic Pattern, Immune Classification, Malignancy Detection, Mutation Prediction, Subtype Classification, Survival Prediction \\
Stomach & 2 & 20 & 1,912 & 1,886 & Grading, Lesion Detection, Malignancy Detection, Mutation Prediction, Subtype Classification, Survival Prediction \\
Cervix & 3 & 10 & 1,500 & 1,490 & Grading, Invasion Detection, Procedure Classification, Survival Prediction \\
Bladder & 2 & 17 & 1,325 & 1,254 & Grading, Invasion Detection, Mutation Prediction, Subtype Classification, Survival Prediction, Tumor Detection \\
Uterus & 3 & 35 & 751 & 656 & Grading, Immune Classification, Mutation Prediction, Subtype Classification, Survival Prediction \\
Head \& Neck & 2 & 11 & 731 & 558 & Grading, Immune Classification, Mutation Prediction, Survival Prediction \\
Soft Tissue & 1 & 6 & 600 & 254 & Mutation Prediction, Subtype Classification, Survival Prediction \\
Thyroid & 1 & 5 & 519 & 506 & Mutation Prediction, Survival Prediction \\
Skin & 1 & 14 & 475 & 433 & Mutation Prediction, Survival Prediction \\
Pancreas & 2 & 10 & 451 & 288 & Grading, Immune Classification, Mutation Prediction, Survival Prediction \\
Adrenal Gland & 2 & 8 & 423 & 232 & Survival Prediction \\
Liver & 2 & 12 & 418 & 404 & Grading, Mutation Prediction, Survival Prediction \\
Testis & 1 & 5 & 303 & 225 & Subtype Classification, Survival Prediction \\
Ovary & 2 & 5 & 266 & 156 & Immune Classification, Survival Prediction \\
Thymus & 1 & 4 & 181 & 121 & Subtype Classification, Survival Prediction \\
Esophagus & 1 & 6 & 158 & 156 & Grading, Subtype Classification, Survival Prediction \\
Eye & 1 & 4 & 80 & 80 & Mutation Prediction, Survival Prediction \\
Lymph Node & 1 & 4 & 44 & 44 & Survival Prediction \\
\bottomrule
\end{tabular}%
}
\end{table}

\subsection{Stage~2 Sparse Supervision Structure}
\label[appendix]{appendix_sec:sparse_supervision_structure}

\begin{figure}[tb]
  \centering
  \includegraphics[width=0.62\linewidth]{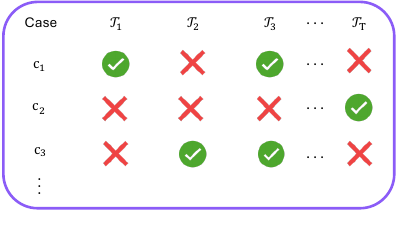}
  \caption{Schematic of sparse case-task supervision in Stage~2. Rows denote cases and columns denote tasks. A check mark indicates an available supervision target for that case-task pair; a cross indicates missing supervision.}
  \label{fig:stage2_sparse_supervision}
\end{figure}

Stage~2 supervision is inherently sparse because each supervised case is only labeled for the subset of tasks available in its source cohort. Let $M \in \{0,1\}^{N \times T}$ denote the case-task supervision matrix, where $M_{i,t}=1$ if supervised case $c_i$ has a valid label for task $\mathcal{T}_t$ and $M_{i,t}=0$ otherwise. For slide-level datasets without case identifiers, the corresponding supervised case contains a single slide. For each task during training, loss is computed only on labeled cases, and per-batch optimization averages only over active tasks with usable labels (as defined in \Cref{sec:semantic_alignment}).

\subsection{TCGA Task Preparation}
WSIs were collected from The Cancer Genome Atlas (TCGA) through the Genomic Data Commons (GDC) portal~\cite{TCGA_Program_NCI}. We first linked every TCGA slide to its case identifier, then harmonized case-level clinical and molecular attributes. We include 240 TCGA tasks (117 classification, 123 survival), corresponding to 72.1\% of all Stage~2 tasks and 96.1\% of all survival tasks. TCGA supervision spans 32 TCGA projects plus one pan-cancer pooled task, with 8 slide-level tasks and 232 case-level tasks. Across all TCGA tasks, the labeled union covers 9{,}732 unique cases and 11{,}857 unique slides. Survival endpoints include overall survival (OS), disease-specific survival (DSS), disease-free interval (DFI), and progression-free interval (PFI). \Cref{tab:tcga_task_family_summary} summarizes task families and cohort coverage.

\begin{table}[tb]
\centering
\caption{TCGA task families used in the supervised training run. Unique case/slide counts are union counts within each family and are not additive across rows.}
\label{tab:tcga_task_family_summary}
\small
\setlength{\tabcolsep}{4pt}
\resizebox{\textwidth}{!}{%
\begin{tabular}{l r c c l r r}
\toprule
\textbf{Task Family} & \textbf{Tasks} & \textbf{Level} & \textbf{Cohorts} & \textbf{Label Space} & \textbf{Unique Cases} & \textbf{Unique Slides} \\
\midrule
Pan-cancer cancer type & 1 & Slide & Pooled TCGA & 46 classes & 9,148 & 11,185 \\
Primary diagnosis & 7 & Slide & 7 cohorts & 2 to 3 classes & 2,284 & 2,284 \\
Tumor grade & 10 & Case & 10 cohorts & G1 to G4 (2 to 4 observed classes) & 3,274 & 3,790 \\
Mutation status & 99 & Case & 20 cohorts & Binary (wildtype vs mutant) & 7,955 & 9,619 \\
Survival endpoints & 123 & Case & 32 cohorts (DFI in 27) & Adaptive discrete-hazard bins (2 to 16 bins) & 9,578 & 11,675 \\
\midrule
TCGA union (all families) & 240 & Mixed & 32 + pooled & Mixed & 9,732 & 11,857 \\
\bottomrule
\end{tabular}%
}
\end{table}

\textbf{Pan-cancer cancer-type classification (slide-level).} We built a pooled TCGA benchmark with 11{,}185 slides and got 46 cancer subtypes labels from~\cite{ding2024multimodal}. Labels are defined at slide level and retain cohort-specific subtype granularity.

\textbf{Primary diagnosis classification (slide-level).} We derived cohort-specific primary-diagnosis tasks from diagnosis records marked as primary disease, excluding missing and non-informative values. We required at least two classes with at least 25 cases per class. Seven cohorts satisfied these criteria and are detailed in Table~\ref{tab:tcga_cohorts}.

\textbf{Tumor grade classification (case-level).} We extracted cohort-specific tumor-grade labels from TCGA diagnosis metadata after harmonizing grade strings into canonical G1 to G4 categories. A task was retained only when at least two classes remained after cleaning and each class had sufficient support (minimum 10 cases per class).

\textbf{Survival prediction (case-level).} We used four TCGA-CDR endpoints: overall survival (OS), disease-specific survival (DSS), disease-free interval (DFI), and progression-free interval (PFI). For each cohort-endpoint pair, event indicators and follow-up times were filtered to valid entries, then modeled with a discrete-hazard objective using task-specific quantile time bins. The bin count was adaptive with target 8 bins and bounds of 2 to 16 bins. OS, DSS, and PFI were available in all 32 cohorts, while DFI was available in 27 cohorts (missing in TCGA-GBM, TCGA-MESO, TCGA-SKCM, TCGA-THYM, and TCGA-UVM).

\textbf{Mutation status prediction (case-level).} We constructed binary wildtype/mutant tasks for 25 driver genes (ALK, APC, ARID1A, ATRX, BAP1, BRAF, CDKN2A, CTNNB1, EGFR, ERBB2, FBXW7, IDH1, IDH2, KRAS, MET, NF1, PBRM1, PIK3CA, PTEN, RB1, SETD2, SMAD4, TERT, TP53, VHL). Cohort-gene tasks were retained only when both classes met minimum support (10 cases per class), yielding 99 mutation tasks across 20 cohorts.

For case-level tasks (tumor grade, survival, mutation), labels are defined at case level and linked to all slides from that case.

\begin{table}[tb]
\centering
\caption{Included TCGA cohorts and retained primary-diagnosis classes.}
\label{tab:tcga_cohorts}
\small
\setlength{\tabcolsep}{4pt}
\begin{tabularx}{\linewidth}{@{} l l Y r @{}}
\toprule
\textbf{Cohort} & \textbf{Organ} & \textbf{Primary diagnosis classes (count)} & \textbf{Samples} \\
\midrule
TCGA-BRCA & Breast & Infiltrating duct carcinoma (710), Lobular carcinoma (182) & 892 \\
TCGA-COAD & Colorectal & Adenocarcinoma (378), Mucinous adenocarcinoma (59) & 437 \\
TCGA-ESCA & Esophagus & Squamous cell carcinoma (82), Adenocarcinoma (65) & 147 \\
TCGA-SARC & Soft tissue & Leiomyosarcoma (76), Dedifferentiated liposarcoma (41), Undifferentiated sarcoma (15) & 132 \\
TCGA-TGCT & Testis & Seminoma (130), Mixed germ cell tumor (25), Embryonal carcinoma (15) & 170 \\
TCGA-THYM & Thymus & Thymoma type AB (29), Thymoma type B2 (23) & 52 \\
TCGA-UCEC & Uterus & Endometrioid adenocarcinoma (350), Serous cystadenocarcinoma (104) & 454 \\
\bottomrule
\end{tabularx}
\end{table}

\begin{table}[tb]
\centering
\caption{Cohort-level coverage of TCGA tasks used in training. Columns report the number of tasks per family and the union of labeled cases/slides within each cohort across all included TCGA tasks.}
\label{tab:tcga_cohort_task_coverage}
\small
\setlength{\tabcolsep}{3pt}
\resizebox{\textwidth}{!}{%
\begin{tabular}{lrrrrrrr}
\toprule
\textbf{Cohort} & \textbf{Surv.} & \textbf{Mut.} & \textbf{Grade} & \textbf{Prim. Dx} & \textbf{Total} & \textbf{Cases} & \textbf{Slides} \\
\midrule
TCGA-ACC & 4 & 0 & 0 & 0 & 4 & 56 & 227 \\
TCGA-BLCA & 4 & 8 & 0 & 0 & 12 & 386 & 457 \\
TCGA-BRCA & 4 & 5 & 0 & 1 & 10 & 1,062 & 1,133 \\
TCGA-CESC & 4 & 0 & 1 & 0 & 5 & 269 & 279 \\
TCGA-CHOL & 4 & 0 & 1 & 0 & 5 & 39 & 39 \\
TCGA-COAD & 4 & 11 & 0 & 1 & 16 & 451 & 459 \\
TCGA-DLBC & 4 & 0 & 0 & 0 & 4 & 44 & 44 \\
TCGA-ESCA & 4 & 0 & 1 & 1 & 6 & 156 & 158 \\
TCGA-GBM & 3 & 1 & 0 & 0 & 4 & 389 & 860 \\
TCGA-HNSC & 4 & 2 & 1 & 0 & 7 & 450 & 472 \\
TCGA-KICH & 4 & 0 & 0 & 0 & 4 & 108 & 120 \\
TCGA-KIRC & 4 & 4 & 1 & 0 & 9 & 513 & 519 \\
TCGA-KIRP & 4 & 2 & 0 & 0 & 6 & 275 & 299 \\
TCGA-LGG & 4 & 5 & 1 & 0 & 10 & 491 & 844 \\
TCGA-LIHC & 4 & 2 & 1 & 0 & 7 & 365 & 379 \\
TCGA-LUAD & 4 & 6 & 0 & 0 & 10 & 478 & 541 \\
TCGA-LUSC & 4 & 2 & 0 & 0 & 6 & 478 & 512 \\
TCGA-MESO & 3 & 1 & 0 & 0 & 4 & 75 & 87 \\
TCGA-OV & 4 & 0 & 0 & 0 & 4 & 105 & 106 \\
TCGA-PAAD & 4 & 2 & 1 & 0 & 7 & 183 & 209 \\
TCGA-PCPG & 4 & 0 & 0 & 0 & 4 & 176 & 196 \\
TCGA-PRAD & 4 & 0 & 0 & 0 & 4 & 403 & 449 \\
TCGA-READ & 4 & 2 & 0 & 0 & 6 & 165 & 166 \\
TCGA-SARC & 4 & 1 & 0 & 1 & 6 & 254 & 600 \\
TCGA-SKCM & 3 & 11 & 0 & 0 & 14 & 433 & 475 \\
TCGA-STAD & 4 & 10 & 1 & 0 & 15 & 416 & 442 \\
TCGA-TGCT & 4 & 0 & 0 & 1 & 5 & 225 & 303 \\
TCGA-THCA & 4 & 1 & 0 & 0 & 5 & 506 & 519 \\
TCGA-THYM & 3 & 0 & 0 & 1 & 4 & 121 & 181 \\
TCGA-UCEC & 4 & 22 & 1 & 1 & 28 & 505 & 566 \\
TCGA-UCS & 4 & 0 & 0 & 0 & 4 & 57 & 91 \\
TCGA-UVM & 3 & 1 & 0 & 0 & 4 & 80 & 80 \\
\bottomrule
\end{tabular}%
}
\end{table}

\subsection{REG task preparation}
REG dataset pairs whole-slide image identifiers with short, templated pathology report text. Each report follows a consistent schema of the form \emph{``organ, procedure; histologic diagnosis[, grade]''}, which enables a deterministic decomposition into \textit{organ}, \textit{procedure}, and \textit{diagnostic content}. The corpus contains 8{,}494 report--slide pairs spanning breast, prostate, stomach, lung, bladder, colorectal, and cervix specimens. After validation by checking if all of them follow the same structure, 8{,}493 records were retained, a single malformed entry lacking the required delimiter was excluded. This standardized structure permits rule-based label extraction, avoiding variability and potential bias introduced by LLM-based parsing, and yielding stable label definitions anchored to the original clinical phrasing.

{\textbf{Normalization and parsing.}}
Before extracting labels, we unified organ names to a single vocabulary (e.g., \textit{urinary bladder}$\rightarrow$\textit{bladder}, \textit{uterine cervix}$\rightarrow$\textit{cervix}, \textit{nipple}$\rightarrow$\textit{breast}). For tasks spanning multiple organs, colon and rectum were merged into a single \textit{colorectal} category, as they share the same diagnostic criteria. Each report was then split into its three fields using the fixed delimiters, and all labels were derived solely from explicit wording in those fields, never inferred.

\begin{figure}[tb]
  \centering
  \includegraphics[width=\linewidth]{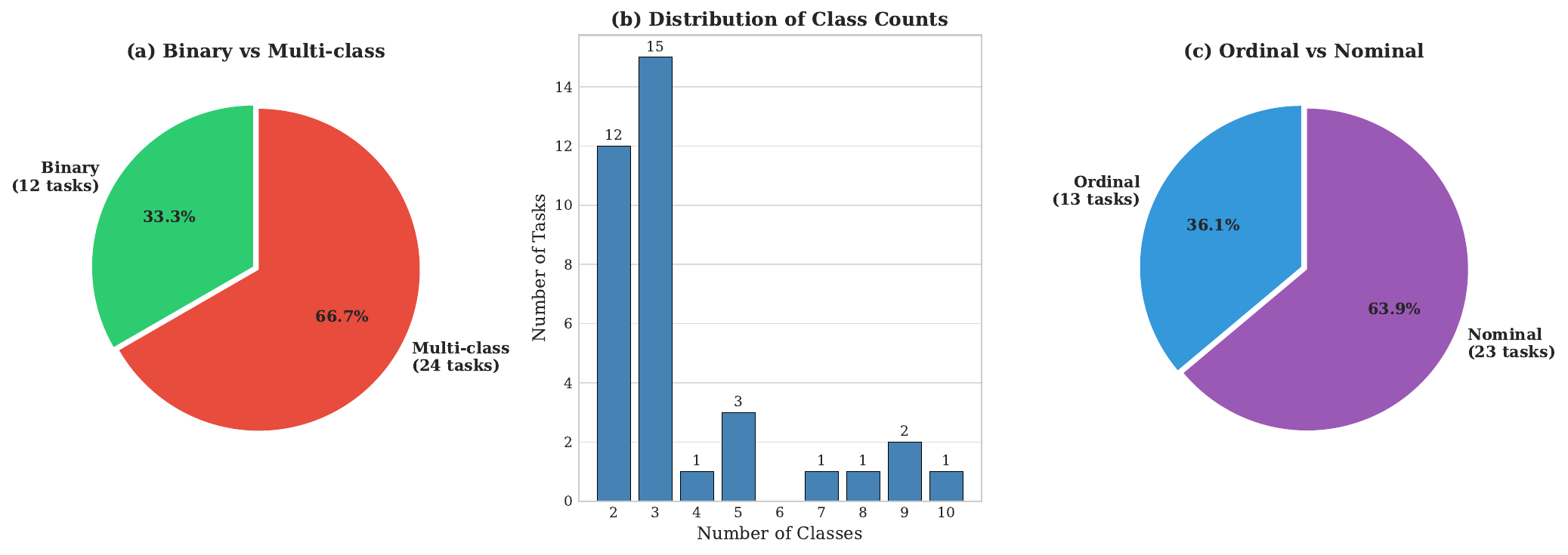}
  \caption{Characterization of REG classification tasks. \textbf{(a)}~Proportion of binary (33.3\%, 12 tasks) versus multi-class (66.7\%, 24 tasks) tasks. \textbf{(b)}~Distribution of class counts per task, with the majority having 2 or 3 classes. \textbf{(c)}~Proportion of ordinal tasks (36.1\%, 13 tasks, \eg, grading) versus nominal tasks (63.9\%, 23 tasks, \eg, subtype classification).}
  \label{fig:reg_tasks_complexity}
\end{figure}

{\textbf{Labeling principle.}}
Label assignment was deliberately conservative: a label was assigned only when the report contained clear, unambiguous textual evidence, and the sample was excluded from that task otherwise. For yes/no attributes, ``Present'' required an explicit positive statement, and ``Absent'' required either an explicit negation or a diagnosis that logically rules out the attribute. This conservative policy means each task has its own subset of samples. Slides whose reports lack sufficient evidence for a given task simply do not appear in that task's dataset. \Cref{fig:reg_tasks_complexity} summarizes the resulting 36 tasks by class count, task type, and label structure. Below we describe the tasks constructed for each organ.

\textbf{Breast (7 tasks).}
We extracted seven tasks from breast reports.
\emph{Histologic type} identifies the specific category of breast tissue abnormality (e.g., invasive carcinoma of no special type, invasive lobular carcinoma, ductal carcinoma in situ (DCIS), fibroadenoma, phyllodes tumor). When both an invasive component and an in-situ component are mentioned, the invasive component takes precedence.
\emph{DCIS presence} indicates whether a pre-invasive lesion confined to the milk ducts is mentioned. It is only marked \emph{Absent} for cases where invasive carcinoma is confirmed with no DCIS mentioned.
\emph{Overall histologic grade}, \emph{tubule formation}, \emph{nuclear grade}, and \emph{mitotic score} are the four components of the Nottingham grading system, a standardized scoring scheme that quantifies how abnormal the cancer cells look and how fast they are dividing. Each is labeled only when explicitly stated in the report.
\emph{Procedure type} records the biopsy method (core-needle, sono-guided core, mammotome, or biopsy NOS) from the report header.

\textbf{Prostate (5 tasks).}
We extracted five tasks from prostate reports.
\emph{Gleason score} is the sum of the two most prevalent cancer growth patterns in the sample, each rated 1--5 by the pathologist, where higher scores indicate more aggressive cancer.
\emph{Primary} and \emph{secondary Gleason pattern} are the individual pattern scores that make up the total.
\emph{ISUP grade group} is a simplified 1--5 scale derived from the Gleason score and used internationally to communicate prostate cancer severity.
\emph{Tumor presence} is marked \emph{Present} when cancer terminology (carcinoma/adenocarcinoma) appears and \emph{Absent} only when the report explicitly states no tumor was found.
All four grading labels are assigned only when the report explicitly provides the values.

\textbf{Stomach (5 tasks).}
We extracted five tasks from stomach reports.
\emph{Histologic type} classifies the tissue finding (e.g., adenocarcinoma, tubular adenoma, chronic gastritis, MALT lymphoma, gastrointestinal stromal tumor).
\emph{Differentiation grade} captures how closely cancer cells resemble normal stomach cells (well / moderately / poorly differentiated). It is assigned only for adenocarcinoma cases where the report explicitly states the grade.
\emph{Adenoma dysplasia grade} rates how abnormal the cells in a benign polyp look (low vs.\ high grade), assigned only for explicitly stated adenomas.
\emph{Intestinal metaplasia} flags a specific pre-cancerous change in the stomach lining, labeled \emph{Present} only when explicitly mentioned and \emph{Absent} only when gastritis is stated without it.
\emph{Malignancy status} uses a three-way scheme where \emph{malignant} is assigned for carcinoma/lymphoma, \emph{pre-malignant} for adenoma or high-grade dysplasia, and \emph{benign} for gastritis or polyp, all based on explicit wording. Cases without clear cues are excluded.

\textbf{Lung (3 tasks).}
We extracted three tasks from lung reports.
\emph{Histologic type} distinguishes the three most common lung cancer subtypes (adenocarcinoma, squamous cell carcinoma, small-cell carcinoma) when explicitly named.
\emph{Small-cell vs.\ non-small-cell} separates small-cell lung cancer, a fast-growing subtype requiring different treatment, from all other types, based on explicit terminology.
\emph{Malignancy status} is labeled only from explicit malignant or benign cues.

\textbf{Bladder (5 tasks).}
We extracted five tasks from bladder reports.
\emph{Tumor presence} is labeled \emph{Present} when carcinoma terminology appears and \emph{Absent} when the report explicitly states no tumor.
\emph{Invasiveness} distinguishes whether tumor cells are confined to the inner lining (non-invasive or in situ) versus having grown into deeper tissue layers (invasive).
\emph{Invasion depth} adds finer granularity: no invasion, invasion into the connective tissue just below the lining (subepithelial), or invasion into the muscle wall (muscularis propria), a critical distinction for treatment decisions.
\emph{Papillary grade} classifies non-invasive papillary tumors (finger-like growths from the bladder wall) as low or high grade based on explicit wording.
\emph{Consolidated tumor type} integrates all of the above into a single label, with an explicit ``no tumor'' statement taking precedence.

\textbf{Colorectal (4 tasks).}
We extracted four tasks from colon and rectum reports.
\emph{Histologic type} identifies the tissue finding (e.g., adenocarcinoma, tubular adenoma, sessile serrated lesion, hyperplastic polyp, chronic colitis).
\emph{Differentiation grade} and \emph{adenoma dysplasia grade} follow the same rules as stomach: assigned only for the relevant tissue type and only when explicitly stated.
\emph{Malignancy status} uses the same malignant/pre-malignant/benign scheme as stomach.

\textbf{Cervix (4 tasks).}
We extracted four tasks from cervix reports.
\emph{CIN grade} (cervical intraepithelial neoplasia, stages 1--3) and \emph{SIL grade} (squamous intraepithelial lesion, low or high) are two overlapping grading systems for pre-cancerous changes in the cervix. Both are assigned only in non-invasive contexts and are suppressed if invasive cancer is mentioned.
\emph{Invasive vs.\ pre-invasive} distinguishes cancer that has breached the cervical lining from changes still confined within it.
\emph{Procedure type} records the biopsy method (e.g., colposcopic or punch biopsy) from the report header.

\begin{figure}[tb]
  \centering
  \includegraphics[width=\linewidth]{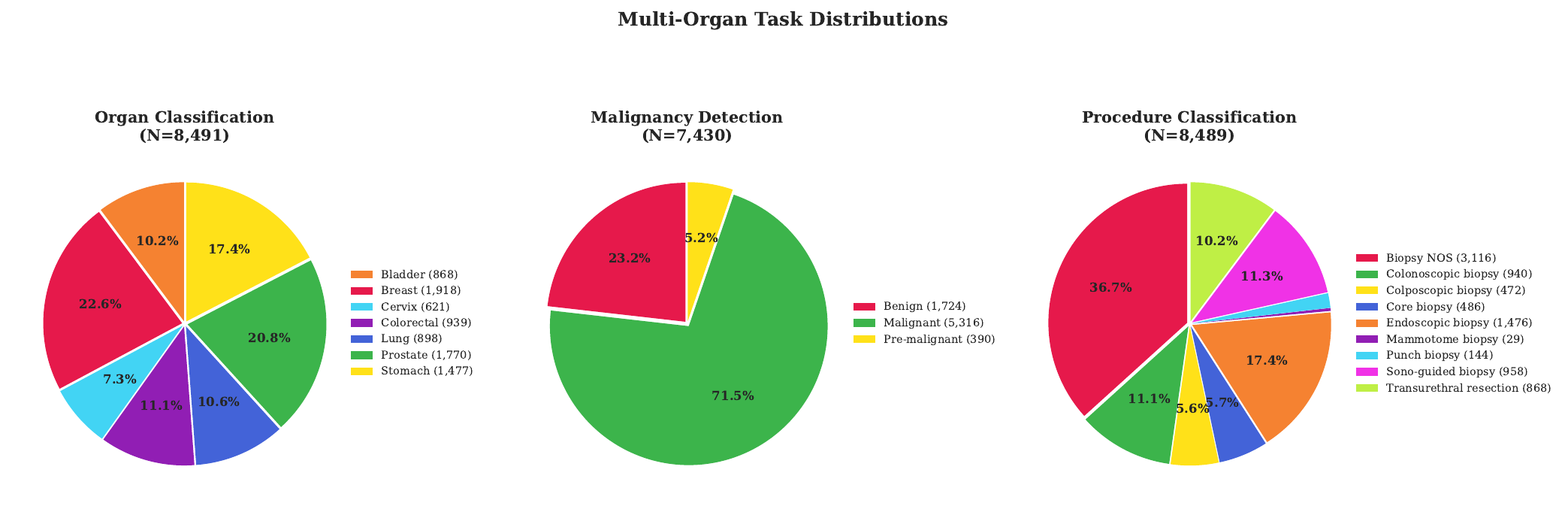}
  \caption{Class distributions for the three cross-organ REG tasks. \textit{Organ Classification} ($N{=}8{,}491$) spans seven organs, with breast (22.6\%) and prostate (20.8\%) as the largest groups. \textit{Malignancy Detection} ($N{=}7{,}430$) exhibits a strong class imbalance, with 71.5\% malignant cases. \textit{Procedure Classification} ($N{=}8{,}489$) covers nine biopsy types, with biopsy NOS (36.7\%) being the most common.}
  \label{fig:reg_multi_organ_details}
\end{figure}

{\textbf{Cross-organ tasks (3 tasks).}}
Beyond the organ-specific tasks, we constructed three tasks that pool samples across all organs.
\emph{Organ classification} predicts which of the seven tissue sites a slide comes from, using the normalized organ vocabulary (with colon and rectum merged into colorectal).
\emph{Procedure classification} predicts the biopsy method using a unified taxonomy spanning all organs (endoscopic, transurethral resection, colposcopic, colonoscopic, sono-guided, core, punch, mammotome, biopsy NOS). A label is assigned only when the procedure is explicitly named in the report header.
\emph{Global malignancy detection} applies the three-way malignant/pre-malignant/benign scheme across all organs: \emph{malignant} for carcinoma, adenocarcinoma, or lymphoma, \emph{pre-malignant} for adenoma or carcinoma in situ (unless invasion is explicitly mentioned), and \emph{benign} for explicit benign or no-tumor statements and non-malignant findings such as inflammation or hyperplasia. As with all other tasks, samples without unambiguous textual evidence are excluded. \Cref{fig:reg_multi_organ_details} shows the class distributions for these three tasks.

\section{Implementation Details}
\label[appendix]{appendix_sec:implementation_details}

\subsection{Spatial Grid Construction}
\label[appendix]{appendix_subsec:spatial_grid}

For patches at level-0 coordinates $(x_i, y_i)$, let $(x_{\min}, y_{\min})$ be the minimum coordinates and $\Delta$ the uniform patch spacing. The grid position $(r, c)$ for patch $i$ is:
\begin{equation}
    r = \left\lfloor \frac{y_i - y_{\min}}{\Delta} + 0.5 \right\rfloor, \quad
    c = \left\lfloor \frac{x_i - x_{\min}}{\Delta} + 0.5 \right\rfloor.
\end{equation}
Grid positions without tissue are filled with zero vectors, and a binary validity mask $\mathbf{V} \in \{0, 1\}^{H \times W}$ tracks tissue presence.

\subsection{Multi-Scale Window Validity Constraint}

For a crop $\mathbf{C}$ of size $S \times S$ with corresponding validity mask $\mathbf{V}_\mathbf{C}$, the minimum valid-token constraint is:
\begin{equation}
    \frac{\sum_{r,c} V_{\mathbf{C},r,c}}{S^2} \geq \rho_{\min}.
\end{equation}
Crops failing this criterion are resampled up to a fixed maximum number of attempts.

\subsection{Block Masking Algorithm}
\label[appendix]{appendix_subsec:block_masking}

Within each batch, a fraction $p_\text{mask}$ of global crops are selected for masking. Their mask ratios are distributed as $\gamma_1, \ldots, \gamma_n = \mathrm{linspace}(\gamma_{\min}, \gamma_{\max}, n)$ and randomly shuffled across the selected crops, ensuring uniform coverage of the masking spectrum within each batch.

For each selected global crop $\mathbf{C}^g_j$ with assigned ratio $\gamma_j$, a binary mask $\mathbf{M}_j \in \{0,1\}^{S_g \times S_g}$ is constructed by iteratively placing rectangular blocks with aspect ratios drawn log-uniformly from $[\alpha_{\min}, \alpha_{\max}]$ at random spatial positions, until the target count of masked valid tokens is reached:
\begin{equation}
    |\{(r,c) : M_{j,r,c} = 1 \land V_{j,r,c} = 1\}| = \lfloor \gamma_j \cdot |\{(r,c) : V_{j,r,c} = 1\}| \rfloor.
\end{equation}
Any remaining budget after block placement is filled by randomly selecting individual valid tokens.

\subsection{Adaptive Token Capping (Stage 2)}
\label[appendix]{appendix_subsec:token_capping}

When the number of remaining valid tokens $V_{i,j}$ (after dropout) exceeds $K_\text{max}$, we apply stratified random sampling to cap the token count while preserving whole-slide spatial coverage. Let $K = K_\text{max}$ and partition the valid-rank indices $\{0, \ldots, V_{i,j}-1\}$ (in flattened raster order) into $K$ equal-width bins, where $s_b$ and $e_b$ denote the start and end rank of bin $b$, respectively:
\begin{equation}
    s_b = \left\lfloor \frac{b \, V_{i,j}}{K} \right\rfloor, \quad
    e_b = \left\lfloor \frac{(b+1) V_{i,j}}{K} \right\rfloor - 1, \quad b = 0, \ldots, K-1.
\end{equation}
For each bin $b$, one offset is drawn uniformly:
\begin{equation}
    u_b \sim \mathrm{Unif}\!\left\{0, \ldots, \max(1, e_b - s_b + 1) - 1\right\},
\end{equation}
and rank $r_b = s_b + u_b$ is retained. The final retained set $\{r_b\}_{b=0}^{K-1}$ is mapped back to original valid-token indices, yielding exactly one sampled token per bin.

\subsection{Survival Bin Selection and Loss}
\label[appendix]{appendix_subsec:survival_bins}

For each survival task $t$, let $E_t$ denote the number of observed (uncensored) events in the training set. The provisional number of discrete time bins is chosen adaptively from $E_t$ using the configured bounds $(B_{\min}, B_{\text{target}}, B_{\max})$:
\begin{equation}
    \hat{B}_t = \begin{cases}
        \max(B_{\min},\ \max(1, E_t)), & E_t < B_{\text{target}}, \\[4pt]
        \min\!\left(B_{\max},\ B_{\text{target}} + \left\lfloor \dfrac{E_t - B_{\text{target}}}{3\,B_{\text{target}}} \right\rfloor\right), & E_t \geq B_{\text{target}}.
    \end{cases}
\end{equation}
This rule limits the number of bins when few events are available and allows the discretization to grow gradually as the event count increases.

To construct the time discretization, we place $\hat{B}_t - 1$ equally-spaced quantile cut-points of the observed event-time distribution in the training set, partitioning the time axis into $\hat{B}_t$ candidate intervals. When event times are tied, multiple cut-points may coincide and are merged, so the effective bin count $B_t \leq \hat{B}_t$ reflects the number of distinct intervals that remain.

For sample $i$, let $\tau_i^{(t)}$ denote the observed survival time and let $\delta_i^{(t)} \in \{0,1\}$ indicate whether the event was observed ($\delta_i^{(t)}=1$) or censored ($\delta_i^{(t)}=0$). Let $j_i^{(t)} \in \{1, \ldots, B_t\}$ denote the index of the time bin containing $\tau_i^{(t)}$. The model outputs one logit $a_{i,k}^{(t)}$ per bin, which is converted to a discrete hazard
\begin{equation}
    h_{i,k}^{(t)} = \sigma(a_{i,k}^{(t)}), \qquad k = 1, \ldots, B_t.
\end{equation}
Here, $h_{i,k}^{(t)}$ represents the conditional probability of experiencing the event in bin $k$, given survival through all preceding bins. The per-sample negative log-likelihood is then
\begin{equation}
    \ell_i^{(t)} = \begin{cases}
        -\left(\displaystyle\sum_{k < j_i^{(t)}} \log(1 - h_{i,k}^{(t)}) + \log h_{i,j_i^{(t)}}^{(t)}\right), & \delta_i^{(t)} = 1, \\[6pt]
        -\displaystyle\sum_{k \leq j_i^{(t)}} \log(1 - h_{i,k}^{(t)}), & \delta_i^{(t)} = 0,
    \end{cases}
\end{equation}
where the first case corresponds to an observed event in bin $j_i^{(t)}$, and the second corresponds to right censoring at bin $j_i^{(t)}$.

\subsection{Projection Head Formulation}
\label[appendix]{appendix_sec:projection_head}

We provide the exact Stage-1 projection-head parameterization used in \cref{sec:ssl_pretraining}. For an input token embedding $\mathbf{h}\in\mathbb{R}^{d}$, the head computes:
\begin{align}
    \mathbf{u}_1 &= \text{GELU}(\mathbf{W}_1^\text{proj} \mathbf{h} + \mathbf{b}_1^\text{proj}), \\
    \mathbf{u}_2 &= \text{GELU}(\mathbf{W}_2^\text{proj} \mathbf{u}_1 + \mathbf{b}_2^\text{proj}), \\
    \mathbf{u}_3 &= \frac{\mathbf{W}_3^\text{proj} \mathbf{u}_2 + \mathbf{b}_3^\text{proj}}{\|\mathbf{W}_3^\text{proj} \mathbf{u}_2 + \mathbf{b}_3^\text{proj}\|_2}, \\
    \mathbf{z} &= \mathbf{W}_4^\text{proj} \mathbf{u}_3.
\end{align}
Here $\mathbf{W}_1^\text{proj}, \mathbf{W}_2^\text{proj} \in \mathbb{R}^{d_h \times d}$ project to hidden dimension $d_h$, $\mathbf{W}_3^\text{proj} \in \mathbb{R}^{d_b \times d_h}$ maps to bottleneck dimension $d_b$ before L2 normalization, and $\mathbf{W}_4^\text{proj} \in \mathbb{R}^{K \times d_b}$ is the weight-normalized prototype layer.

\subsection{Task Head Formulations}
\label[appendix]{appendix_sec:task_heads}

For a case embedding $\tilde{\mathbf{h}}\in\mathbb{R}^{d}$, task heads use one of two parameterizations.

\textbf{Linear head.} The linear head applies feature dropout with rate $p_\text{head}$ before projection:
\begin{equation}
    g_t(\tilde{\mathbf{h}})=\mathbf{W}_t\,\mathrm{Dropout}_{p_\text{head}}(\tilde{\mathbf{h}})+\mathbf{b}_t.
\end{equation}

\textbf{MLP head.} The MLP head uses LayerNorm, two hidden linear layers with GELU, and fixed internal dropout:
\begin{equation}
    g_t(\tilde{\mathbf{h}})=\mathbf{W}_3\phi\!\left(\mathbf{W}_2\phi\!\left(\mathbf{W}_1\mathrm{LN}(\tilde{\mathbf{h}})\right)\right),\quad
    \phi(\mathbf{u})=\mathrm{Dropout}_{0.25}(\mathrm{GELU}(\mathbf{u})).
\end{equation}

\section{Augmentation Strategy Visualizations}
\label[appendix]{appendix_sec:augmentation_strategies}

\begin{table}[tb]
\centering
\caption{Augmentation strategies used in each training stage.}
\label{tab:aug_strategies}
\setlength{\tabcolsep}{3pt}
\small
\begin{tabularx}{\linewidth}{@{}C c c C@{}}
\toprule
\hdrone{Strategy} & \hdrtwo{Stage 1}{SSL} & \hdrtwo{Stage 2}{Alignment} & \hdrone{Applied to} \\
\midrule
Spatial Aug.\ (Flip, Rotation) & \cmark & \cmark & All crops / full grids \\
Block Masking                  & \cmark & \xmark & Global crops only \\
Token Dropout                  & \xmark & \cmark & Full slide grids \\
\bottomrule
\end{tabularx}
\end{table}

The augmentation strategies summarized in \Cref{tab:aug_strategies} are visualized in \Cref{fig:appendix_augmentation_strategies}. Spatial augmentation improves orientation robustness while preserving morphology, token dropout regularizes full-slide inputs during Stage~2, and block-based masking defines the masked prediction signal during Stage~1.

\newlength{\augsubfigH}
\setlength{\augsubfigH}{3.0cm}

\begin{figure}[tb]
  \centering
  \begin{subfigure}[b]{0.32\linewidth}
    \centering
    \includegraphics[width=\linewidth,height=\augsubfigH,keepaspectratio]{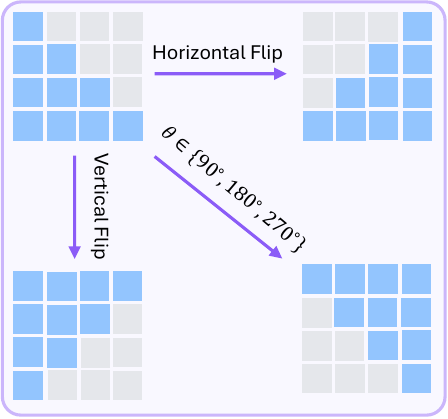}
    \caption{Spatial Augmentation}
  \end{subfigure}\hfill
  \begin{subfigure}[b]{0.32\linewidth}
    \centering
    \includegraphics[width=\linewidth,height=\augsubfigH,keepaspectratio]{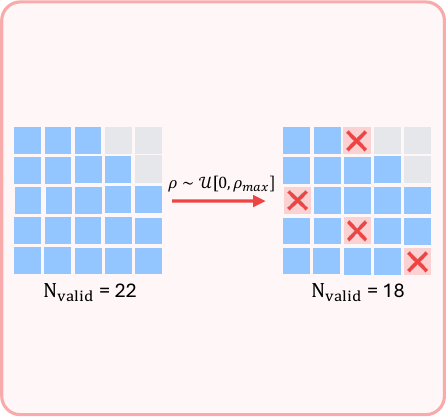}
    \caption{Token Dropout}
  \end{subfigure}\hfill
  \begin{subfigure}[b]{0.32\linewidth}
    \centering
    \includegraphics[width=\linewidth,height=\augsubfigH,keepaspectratio]{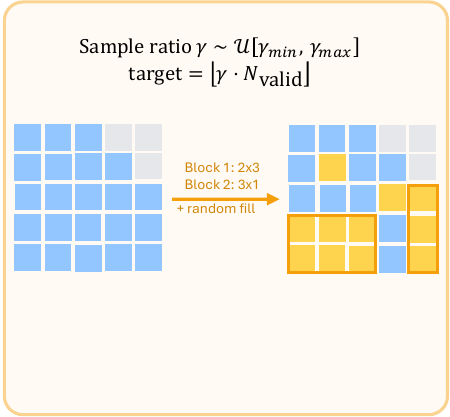}
    \caption{Block-based Masking}
  \end{subfigure}
  \caption{Visual examples of augmentation strategies across training stages.}
  \label{fig:appendix_augmentation_strategies}
\end{figure}

\section{SSL Pretraining Hyperparameters}
\label[appendix]{appendix_sec:ssl_hyperparameters}

\Cref{tab:ssl_hyperparams_architecture,tab:ssl_hyperparams_multicrop,tab:ssl_hyperparams_masking,tab:ssl_hyperparams_optimization,tab:ssl_hyperparams_distillation} provide the complete set of hyperparameters used for self-supervised slide encoder pretraining (Stage~1).

\begin{table}[tb]
\centering
\caption{Encoder architecture hyperparameters.}
\label{tab:ssl_hyperparams_architecture}
\begin{tabular}{lc}
\toprule
\textbf{Hyperparameter} & \textbf{Value} \\
\midrule
Input feature dimension ($d_\text{patch}$) & 384 \\
Model dimension ($d$) & 768 \\
Number of attention heads ($H$) & 12 \\
Number of transformer layers ($D$) & 6 \\
Feed-forward dimension & 3072 \\
Number of register tokens ($R$) & 4 \\
MLP dropout rate & 0.1 \\
Attention dropout rate & 0.0 \\
Stochastic depth max rate & 0.1 \\
LayerScale initialization & Disabled \\
QK normalization & Disabled \\
Learnable ALiBi slopes & No (fixed) \\
\midrule
\multicolumn{2}{l}{\textit{Projection Head}} \\
Hidden dimension & 2048 \\
Bottleneck dimension & 256 \\
Output dimension ($K$) & 8192 \\
Weight normalization & Enabled (frozen gain) \\
\bottomrule
\end{tabular}
\end{table}

\begin{table}[tb]
\centering
\caption{Multi-crop sampling hyperparameters.}
\label{tab:ssl_hyperparams_multicrop}
\begin{tabular}{lc}
\toprule
\textbf{Hyperparameter} & \textbf{Value} \\
\midrule
Number of global crops ($G$) & 2 \\
Global crop size ($S_g \times S_g$) & $20 \times 20$ tokens \\
Number of local crops ($L$) & 4 \\
Local crop size ($S_l \times S_l$) & $12 \times 12$ tokens \\
Minimum valid token ratio ($\rho_{\min}$) & 0.25 \\
Maximum resampling attempts & 3 \\
\midrule
\multicolumn{2}{l}{\textit{Spatial Augmentations}} \\
Horizontal flip probability ($p_h$) & 0.5 \\
Vertical flip probability ($p_v$) & 0.5 \\
Rotation probability ($p_r$) & 0.5 \\
Rotation angles & $\{90^\circ, 180^\circ, 270^\circ\}$ \\
\bottomrule
\end{tabular}
\end{table}

\begin{table}[tb]
\centering
\caption{Block masking hyperparameters.}
\label{tab:ssl_hyperparams_masking}
\begin{tabular}{lc}
\toprule
\textbf{Hyperparameter} & \textbf{Value} \\
\midrule
Masking strategy & Block \\
Mask ratio minimum ($\gamma_{\min}$) & 0.1 \\
Mask ratio maximum ($\gamma_{\max}$) & 0.5 \\
Minimum patches per block & 4 \\
Maximum patches per block & Unlimited \\
Minimum aspect ratio ($\alpha_{\min}$) & 0.3 \\
Maximum aspect ratio ($\alpha_{\max}$) & $1/0.3 \approx 3.33$ \\
Per-crop masking probability & 0.5 \\
Masking applied to & Global crops only \\
\bottomrule
\end{tabular}
\end{table}

\begin{table}[tb]
\centering
\caption{Optimization hyperparameters.}
\label{tab:ssl_hyperparams_optimization}
\begin{tabular}{lc}
\toprule
\textbf{Hyperparameter} & \textbf{Value} \\
\midrule
Optimizer & AdamW \\
Base learning rate & $5 \times 10^{-4}$ \\
Reference batch size for LR scaling & 256 \\
Minimum learning rate & $2 \times 10^{-6}$ \\
Learning rate schedule & Cosine decay \\
Learning rate warmup epochs & 5 \\
Weight decay (start) & 0.04 \\
Weight decay (end) & 0.4 \\
Weight decay schedule & Cosine \\
Gradient clipping (max norm) & 0.3 \\
Mixed precision & BFloat16 \\
\midrule
\multicolumn{2}{l}{\textit{Training Scale}} \\
Micro batch size per GPU & 64 \\
Number of GPUs & 8 \\
Gradient accumulation steps & 2 \\
Effective batch size & 1024 \\
Number of epochs & 200 \\
Total optimizer steps & 14{,}400 \\
Training time (GPU-hours) & ${\approx}436$ \\
\bottomrule
\end{tabular}
\end{table}

\begin{table}[tb]
\centering
\caption{Self-distillation hyperparameters.}
\label{tab:ssl_hyperparams_distillation}
\begin{tabular}{lc}
\toprule
\textbf{Hyperparameter} & \textbf{Value} \\
\midrule
\multicolumn{2}{l}{\textit{EMA Teacher}} \\
Initial momentum ($\mu_0$) & 0.996 \\
Final momentum ($\mu_T$) & 1.0 \\
Momentum schedule & Cosine \\
\midrule
\multicolumn{2}{l}{\textit{Temperature}} \\
Student temperature ($\tau_s$) & 0.1 \\
Teacher CLS temperature (start) & 0.04 \\
Teacher CLS temperature (end, $\tau_t$) & 0.07 \\
Teacher patch temperature (start) & 0.04 \\
Teacher patch temperature (end, $\tau_t^\text{patch}$) & 0.07 \\
Temperature warmup epochs & 30 \\
\midrule
\multicolumn{2}{l}{\textit{Centering}} \\
Center momentum ($\lambda$) & 0.9 \\
\midrule
\multicolumn{2}{l}{\textit{Stabilization}} \\
Freeze final projection layer (epochs) & 3 \\
\bottomrule
\end{tabular}
\end{table}

\section{Semantic Alignment Hyperparameters}
\label[appendix]{appendix_sec:semantic_alignment_hyperparameters}

\Cref{tab:semantic_alignment_architecture,tab:semantic_alignment_optimization,tab:semantic_alignment_augmentation,tab:semantic_alignment_loss} provide the complete set of hyperparameters used for case-aware semantic alignment (Stage~2).

\begin{table}[tb]
\centering
\caption{Architecture hyperparameters for semantic alignment. The slide encoder uses the same architecture as SSL pretraining (\cref{tab:ssl_hyperparams_architecture}), initialized from the pretrained teacher weights.}
\label{tab:semantic_alignment_architecture}
\begin{tabular}{lc}
\toprule
\textbf{Hyperparameter} & \textbf{Value} \\
\midrule
\multicolumn{2}{l}{\textit{Case Transformer}} \\
Number of layers ($D_\text{case}$) & 3 \\
Number of attention heads & 12 \\
Feed-forward dimension & $3072$ \\
Dropout rate & 0.1 \\
LayerScale initialization & $10^{-5}$ \\
DropPath max rate & Disabled \\
\texttt{[CASE]} token initialization std & 0.02 \\
\midrule
\multicolumn{2}{l}{\textit{Task Heads}} \\
Head type & MLP \\
Head dropout rate & 0.1 \\
\bottomrule
\end{tabular}
\end{table}

\begin{table}[tb]
\centering
\caption{Optimization hyperparameters for semantic alignment.}
\label{tab:semantic_alignment_optimization}
\begin{tabular}{lc}
\toprule
\textbf{Hyperparameter} & \textbf{Value} \\
\midrule
Optimizer & AdamW \\
Base learning rate & $5 \times 10^{-5}$ \\
Minimum learning rate & $2 \times 10^{-7}$ \\
Learning rate schedule & Cosine annealing \\
Warmup steps & 0 \\
Weight decay & 0.4 \\
Gradient clipping (max norm) & 0.3 \\
Mixed precision & BFloat16 \\
\midrule
\multicolumn{2}{l}{\textit{Training Scale}} \\
Micro batch size per GPU & 1 case \\
Number of GPUs & 8 \\
Gradient accumulation steps & 128 \\
Effective batch size & 1024 cases \\
Number of epochs & 20 \\
Total optimizer steps & 1{,}000 \\
Training time (GPU-hours) & ${\approx}512$ \\
\bottomrule
\end{tabular}
\end{table}

\begin{table}[tb]
\centering
\caption{Data augmentation hyperparameters for semantic alignment.}
\label{tab:semantic_alignment_augmentation}
\begin{tabular}{lc}
\toprule
\textbf{Hyperparameter} & \textbf{Value} \\
\midrule
\multicolumn{2}{l}{\textit{Spatial Augmentations}} \\
Horizontal flip probability ($p_h$) & 0.5 \\
Vertical flip probability ($p_v$) & 0.5 \\
Rotation probability ($p_r$) & 0.5 \\
Rotation angles & $\{90^\circ, 180^\circ, 270^\circ\}$ \\
\midrule
\multicolumn{2}{l}{\textit{Token Dropout}} \\
Maximum dropout ratio ($\rho_\text{max}$) & 0.1 \\
\bottomrule
\end{tabular}
\end{table}

\begin{table}[tb]
\centering
\caption{Loss function hyperparameters for semantic alignment.}
\label{tab:semantic_alignment_loss}
\begin{tabular}{lc}
\toprule
\textbf{Hyperparameter} & \textbf{Value} \\
\midrule
\multicolumn{2}{l}{\textit{Classification Tasks}} \\
Label smoothing ($\epsilon$) & 0.03 \\
Class weighting & Inverse frequency \\
\midrule
\multicolumn{2}{l}{\textit{Survival Tasks}} \\
Loss function & Discrete-time NLL (hazard) \\
Time bins (min / target / max) & 2 / 8 / 16 \\
Class weighting & None \\
\midrule
\multicolumn{2}{l}{\textit{Multi-Task Aggregation}} \\
Task loss weighting & Equal (average) \\
\midrule
\multicolumn{2}{l}{\textit{Validation}} \\
Validation split ratio & 0.05 (5\%) \\
Stratification & By task (class/event) \\
\bottomrule
\end{tabular}
\end{table}

\section{MLP Probe Evaluation Protocol}
\label[appendix]{appendix_sec:mlp_probe_setup}
\label[appendix]{appendix_sec:mil_benchmark_setup}

Both the slide encoder and MIL comparisons share the same five-fold evaluation protocol. Folds are constructed with label stratification. Each fold applies an 80\%/20\% train-validation split. Model selection is based on best validation weighted F1 score. We report mean $\pm$ standard deviation across folds for weighted F1, weighted ROC-AUC, and balanced accuracy.

For MLP probe setup in slide encoders comparison (\Cref{tab:encoder_comparison}), we use a three-layer MLP classifier on frozen slide representations:
\begin{center}
\texttt{LayerNorm(D)} $\rightarrow$ \texttt{Linear(D, h1)} $\rightarrow$ \texttt{GELU} $\rightarrow$ \texttt{Dropout} $\rightarrow$\\
\texttt{Linear(h1, h2)} $\rightarrow$ \texttt{GELU} $\rightarrow$ \texttt{Dropout} $\rightarrow$ \texttt{Linear(h2, C)},
\end{center}
with adaptive hidden sizes $h_1=\max(4,\mathrm{round}(0.66D))$ and $h_2=\max(2,\mathrm{round}(0.5h_1))$.
Optimization uses AdamW with learning rate $1\times10^{-3}$ and weight decay $1\times10^{-2}$.
Training runs for 200 epochs with batch size 64, cross-entropy loss, and dropout 0.25 in both hidden blocks. Class-balanced sampling is applied during training.

The MLP prob setup in the MIL comparison (\Cref{tab:mil_comparison_macro}) uses the same MLP head used in slide encoder comparison, and follows the MIL-Lab implementation \cite{shao2025miltransfer} for different MILs. Each patch encoder is paired with five MIL architectures (MeanMIL, ABMIL, CLAM, DSMIL, and TransMIL). Each reported encoder entry is the arithmetic mean over the five architectures in table \Cref{tab:mil_comparison}.
Optimization uses AdamW with learning rate $1\times10^{-3}$ and weight decay $1\times10^{-2}$ for 100 epochs with one bag per iteration and class-balanced sampling.
For multi-slide cases, case-level predictions are obtained by averaging per-slide logits (late fusion).

\section{Linear Probe Setup and Results}
\label[appendix]{appendix_sec:linear_probe_setup}

For completeness, we additionally evaluate different slide encoders and MILs with a multinomial logistic-regression classifier.
We use the same five-fold splits as the MLP comparison, with label stratification and case-level grouping and an 80\% to 20\% train validation split in each fold.
The classifier uses L2 regularization and selects regularization strength by minimizing validation loss over 45 logarithmically spaced values from $10^{-6}$ to $10^{5}$.
Optimization uses LBFGS with a maximum of 500 iterations and class-balanced weighting.
Performance is reported as mean and standard deviation across folds for weighted F1, weighted ROC-AUC, and balanced accuracy.
Task-level linear-probe results across slide encoders and MIL baselines are shown in \Cref{tab:linear_probe_results,tab:linear_probe_mil_results}. For the MIL table, each encoder entry is averaged over MeanMIL, ABMIL, CLAM, DSMIL, and TransMIL.

\begin{table}[tb]
\caption{Linear-probe slide encoder comparison across sixteen held-out tasks.}
\label{tab:linear_probe_results}
\centering
\scriptsize
\setlength{\tabcolsep}{2pt}
\renewcommand{\arraystretch}{0.90}
\resizebox{\linewidth}{!}{%
\begin{tabular}{%
  >{\raggedright\arraybackslash}p{1.9cm}%
  c%
  c%
  c%
  c%
  c%
  c%
  c}
\toprule
Task & Metric & CHIEF & \shortstack{Giga-\\Path} & PRISM & \shortstack{Made-\\leine} & TITAN & \shortstack{MOOZY\\(Ours)} \\
\midrule
\multirow{3}{*}{\parbox{1.9cm}{\raggedright Residual Cancer Burden}}
  & F1        & $0.34_{\pm0.06}$ & $0.32_{\pm0.05}$ & $0.33_{\pm0.09}$ & $\underline{0.36}_{\pm0.07}$ & $0.34_{\pm0.09}$ & $\mathbf{0.38}_{\pm0.11}$ \\
  & AUC       & $0.58_{\pm0.05}$ & $0.57_{\pm0.07}$ & $0.61_{\pm0.03}$ & $\underline{0.62}_{\pm0.03}$ & $0.57_{\pm0.06}$ & $\mathbf{0.66}_{\pm0.05}$ \\
  & Bal.\ Acc & $0.39_{\pm0.04}$ & $0.30_{\pm0.06}$ & $0.32_{\pm0.02}$ & $\underline{0.40}_{\pm0.05}$ & $0.38_{\pm0.08}$ & $\mathbf{0.44}_{\pm0.07}$ \\
\midrule
\multirow{3}{*}{\parbox{1.9cm}{\raggedright TP53 Mut.}}
  & F1        & $0.70_{\pm0.08}$ & $0.69_{\pm0.09}$ & $0.77_{\pm0.07}$ & $0.77_{\pm0.06}$ & $\mathbf{0.84}_{\pm0.04}$ & $\underline{0.80}_{\pm0.06}$ \\
  & AUC       & $0.80_{\pm0.09}$ & $0.75_{\pm0.06}$ & $0.85_{\pm0.06}$ & $0.85_{\pm0.06}$ & $\mathbf{0.88}_{\pm0.04}$ & $\underline{0.87}_{\pm0.03}$ \\
  & Bal.\ Acc & $0.70_{\pm0.07}$ & $0.71_{\pm0.08}$ & $0.78_{\pm0.08}$ & $0.78_{\pm0.05}$ & $\mathbf{0.84}_{\pm0.03}$ & $\underline{0.81}_{\pm0.05}$ \\
\midrule
\multirow{3}{*}{\parbox{1.9cm}{\raggedright BAP1 Mut.}}
  & F1        & $0.75_{\pm0.06}$ & $0.75_{\pm0.10}$ & $0.71_{\pm0.11}$ & $0.73_{\pm0.09}$ & $\underline{0.75}_{\pm0.10}$ & $\mathbf{0.78}_{\pm0.05}$ \\
  & AUC       & $0.67_{\pm0.14}$ & $0.66_{\pm0.15}$ & $0.63_{\pm0.16}$ & $\underline{0.69}_{\pm0.14}$ & $0.68_{\pm0.09}$ & $\mathbf{0.74}_{\pm0.11}$ \\
  & Bal.\ Acc & $\underline{0.60}_{\pm0.19}$ & $0.56_{\pm0.15}$ & $0.54_{\pm0.18}$ & $0.60_{\pm0.19}$ & $0.59_{\pm0.16}$ & $\mathbf{0.64}_{\pm0.12}$ \\
\midrule
\multirow{3}{*}{\parbox{1.9cm}{\raggedright ACVR2A Mut.}}
  & F1        & $\mathbf{0.86}_{\pm0.03}$ & $0.69_{\pm0.13}$ & $0.77_{\pm0.06}$ & $0.78_{\pm0.09}$ & $0.81_{\pm0.06}$ & $\underline{0.83}_{\pm0.04}$ \\
  & AUC       & $0.79_{\pm0.09}$ & $0.69_{\pm0.15}$ & $0.81_{\pm0.11}$ & $0.76_{\pm0.16}$ & $\underline{0.82}_{\pm0.04}$ & $\mathbf{0.89}_{\pm0.07}$ \\
  & Bal.\ Acc & $\underline{0.76}_{\pm0.07}$ & $0.52_{\pm0.14}$ & $0.68_{\pm0.15}$ & $0.66_{\pm0.14}$ & $0.69_{\pm0.13}$ & $\mathbf{0.78}_{\pm0.10}$ \\
\midrule
\multirow{3}{*}{\parbox{1.9cm}{\raggedright Histologic Grade}}
  & F1        & $0.63_{\pm0.03}$ & $\mathbf{0.68}_{\pm0.04}$ & $0.56_{\pm0.14}$ & $0.62_{\pm0.07}$ & $0.62_{\pm0.06}$ & $\underline{0.66}_{\pm0.08}$ \\
  & AUC       & $0.65_{\pm0.05}$ & $\mathbf{0.77}_{\pm0.06}$ & $0.60_{\pm0.18}$ & $0.71_{\pm0.12}$ & $0.64_{\pm0.06}$ & $\underline{0.74}_{\pm0.11}$ \\
  & Bal.\ Acc & $0.64_{\pm0.04}$ & $\mathbf{0.69}_{\pm0.04}$ & $0.56_{\pm0.14}$ & $0.62_{\pm0.09}$ & $0.63_{\pm0.07}$ & $\underline{0.68}_{\pm0.08}$ \\
\midrule
\multirow{3}{*}{\parbox{1.9cm}{\raggedright KRAS Mut.}}
  & F1        & $0.59_{\pm0.09}$ & $0.71_{\pm0.07}$ & $0.59_{\pm0.07}$ & $0.66_{\pm0.15}$ & $\mathbf{0.72}_{\pm0.08}$ & $\underline{0.71}_{\pm0.08}$ \\
  & AUC       & $0.63_{\pm0.10}$ & $0.69_{\pm0.10}$ & $0.59_{\pm0.07}$ & $0.62_{\pm0.15}$ & $\mathbf{0.77}_{\pm0.06}$ & $\underline{0.73}_{\pm0.07}$ \\
  & Bal.\ Acc & $0.55_{\pm0.10}$ & $\mathbf{0.73}_{\pm0.07}$ & $0.58_{\pm0.06}$ & $0.63_{\pm0.18}$ & $\underline{0.73}_{\pm0.09}$ & $0.71_{\pm0.09}$ \\
\midrule
\multirow{3}{*}{\parbox{1.9cm}{\raggedright IDH Status}}
  & F1        & $0.92_{\pm0.02}$ & $\underline{0.93}_{\pm0.01}$ & $0.89_{\pm0.02}$ & $0.90_{\pm0.01}$ & $0.93_{\pm0.02}$ & $\mathbf{0.95}_{\pm0.02}$ \\
  & AUC       & $0.96_{\pm0.01}$ & $0.97_{\pm0.01}$ & $0.96_{\pm0.01}$ & $0.96_{\pm0.01}$ & $\underline{0.98}_{\pm0.01}$ & $\mathbf{0.99}_{\pm0.01}$ \\
  & Bal.\ Acc & $0.91_{\pm0.02}$ & $0.93_{\pm0.02}$ & $0.89_{\pm0.02}$ & $0.90_{\pm0.02}$ & $\underline{0.93}_{\pm0.02}$ & $\mathbf{0.95}_{\pm0.02}$ \\
\midrule
\multirow{3}{*}{\parbox{1.9cm}{\raggedright Treatment Response}}
  & F1        & $0.39_{\pm0.06}$ & $\underline{0.39}_{\pm0.10}$ & $0.39_{\pm0.12}$ & $0.36_{\pm0.11}$ & $0.34_{\pm0.04}$ & $\mathbf{0.47}_{\pm0.11}$ \\
  & AUC       & $0.64_{\pm0.06}$ & $0.64_{\pm0.08}$ & $\mathbf{0.69}_{\pm0.10}$ & $0.61_{\pm0.05}$ & $\underline{0.64}_{\pm0.04}$ & $0.60_{\pm0.08}$ \\
  & Bal.\ Acc & $\underline{0.37}_{\pm0.09}$ & $0.29_{\pm0.10}$ & $0.33_{\pm0.10}$ & $0.34_{\pm0.16}$ & $0.30_{\pm0.02}$ & $\mathbf{0.38}_{\pm0.10}$ \\
\midrule
\multirow{3}{*}{\parbox{1.9cm}{\raggedright BRCA PAM50 Subtype}}
  & F1        & $0.62_{\pm0.05}$ & $0.63_{\pm0.06}$ & $\underline{0.65}_{\pm0.04}$ & $0.63_{\pm0.06}$ & $\mathbf{0.66}_{\pm0.04}$ & $0.58_{\pm0.05}$ \\
  & AUC       & $0.85_{\pm0.03}$ & $0.84_{\pm0.05}$ & $\underline{0.86}_{\pm0.04}$ & $0.85_{\pm0.04}$ & $\mathbf{0.87}_{\pm0.03}$ & $0.82_{\pm0.03}$ \\
  & Bal.\ Acc & $0.55_{\pm0.04}$ & $0.51_{\pm0.08}$ & $0.55_{\pm0.05}$ & $\underline{0.56}_{\pm0.05}$ & $\mathbf{0.58}_{\pm0.05}$ & $0.50_{\pm0.06}$ \\
\midrule
\multirow{3}{*}{\parbox{1.9cm}{\raggedright HNSC mRNA Subtype}}
  & F1        & $0.51_{\pm0.02}$ & $0.53_{\pm0.04}$ & $\underline{0.57}_{\pm0.05}$ & $0.50_{\pm0.05}$ & $\mathbf{0.58}_{\pm0.06}$ & $0.45_{\pm0.02}$ \\
  & AUC       & $0.74_{\pm0.02}$ & $0.75_{\pm0.02}$ & $\underline{0.79}_{\pm0.04}$ & $0.77_{\pm0.02}$ & $\mathbf{0.80}_{\pm0.02}$ & $0.70_{\pm0.03}$ \\
  & Bal.\ Acc & $0.51_{\pm0.04}$ & $0.53_{\pm0.06}$ & $\mathbf{0.57}_{\pm0.04}$ & $0.48_{\pm0.06}$ & $\underline{0.54}_{\pm0.05}$ & $0.46_{\pm0.02}$ \\
\midrule
\multirow{3}{*}{\parbox{1.9cm}{\raggedright UCEC Genomic Subtype}}
  & F1        & $0.50_{\pm0.04}$ & $0.52_{\pm0.06}$ & $\underline{0.58}_{\pm0.03}$ & $0.50_{\pm0.04}$ & $\mathbf{0.62}_{\pm0.04}$ & $0.54_{\pm0.02}$ \\
  & AUC       & $0.75_{\pm0.03}$ & $0.77_{\pm0.03}$ & $\underline{0.82}_{\pm0.03}$ & $0.76_{\pm0.02}$ & $\mathbf{0.82}_{\pm0.02}$ & $0.77_{\pm0.03}$ \\
  & Bal.\ Acc & $0.49_{\pm0.05}$ & $0.51_{\pm0.07}$ & $\underline{0.55}_{\pm0.03}$ & $0.48_{\pm0.04}$ & $\mathbf{0.59}_{\pm0.04}$ & $0.52_{\pm0.02}$ \\
\midrule
\multirow{3}{*}{\parbox{1.9cm}{\raggedright Synaptophysin Grade}}
  & F1        & $\underline{0.71}_{\pm0.09}$ & $0.71_{\pm0.08}$ & $0.60_{\pm0.09}$ & $0.68_{\pm0.09}$ & $\mathbf{0.73}_{\pm0.02}$ & $0.63_{\pm0.08}$ \\
  & AUC       & $0.62_{\pm0.16}$ & $\mathbf{0.72}_{\pm0.11}$ & $0.46_{\pm0.06}$ & $0.67_{\pm0.09}$ & $\underline{0.67}_{\pm0.06}$ & $0.61_{\pm0.09}$ \\
  & Bal.\ Acc & $0.57_{\pm0.11}$ & $\underline{0.60}_{\pm0.08}$ & $0.46_{\pm0.04}$ & $0.56_{\pm0.11}$ & $\mathbf{0.63}_{\pm0.06}$ & $0.56_{\pm0.12}$ \\
\midrule
\multirow{3}{*}{\parbox{1.9cm}{\raggedright RAS/BRAF Status}}
  & F1        & $0.83_{\pm0.10}$ & $\underline{0.86}_{\pm0.08}$ & $0.73_{\pm0.14}$ & $0.80_{\pm0.14}$ & $0.85_{\pm0.06}$ & $\mathbf{0.88}_{\pm0.09}$ \\
  & AUC       & $0.66_{\pm0.11}$ & $0.56_{\pm0.15}$ & $0.36_{\pm0.15}$ & $0.38_{\pm0.23}$ & $\mathbf{0.74}_{\pm0.20}$ & $\underline{0.73}_{\pm0.23}$ \\
  & Bal.\ Acc & $0.61_{\pm0.24}$ & $0.61_{\pm0.21}$ & $0.47_{\pm0.24}$ & $0.60_{\pm0.23}$ & $\underline{0.67}_{\pm0.19}$ & $\mathbf{0.75}_{\pm0.20}$ \\
\midrule
\multirow{3}{*}{\parbox{1.9cm}{\raggedright Keratinizing SCC Grade}}
  & F1        & $\underline{0.68}_{\pm0.03}$ & $0.64_{\pm0.06}$ & $0.64_{\pm0.06}$ & $0.66_{\pm0.05}$ & $\mathbf{0.70}_{\pm0.04}$ & $0.66_{\pm0.03}$ \\
  & AUC       & $\underline{0.73}_{\pm0.03}$ & $0.70_{\pm0.07}$ & $0.71_{\pm0.07}$ & $0.72_{\pm0.05}$ & $\mathbf{0.75}_{\pm0.06}$ & $0.73_{\pm0.03}$ \\
  & Bal.\ Acc & $\underline{0.68}_{\pm0.03}$ & $0.63_{\pm0.06}$ & $0.63_{\pm0.06}$ & $0.66_{\pm0.05}$ & $\mathbf{0.69}_{\pm0.04}$ & $0.65_{\pm0.04}$ \\
\midrule
\multirow{3}{*}{\parbox{1.9cm}{\raggedright Non-Keratinizing SCC Grade}}
  & F1        & $\mathbf{0.70}_{\pm0.15}$ & $0.63_{\pm0.09}$ & $0.46_{\pm0.14}$ & $0.62_{\pm0.20}$ & $\underline{0.67}_{\pm0.16}$ & $0.55_{\pm0.09}$ \\
  & AUC       & $\underline{0.67}_{\pm0.12}$ & $0.56_{\pm0.13}$ & $0.37_{\pm0.22}$ & $0.61_{\pm0.08}$ & $\mathbf{0.69}_{\pm0.10}$ & $0.59_{\pm0.07}$ \\
  & Bal.\ Acc & $\mathbf{0.67}_{\pm0.16}$ & $0.58_{\pm0.03}$ & $0.41_{\pm0.14}$ & $0.58_{\pm0.16}$ & $\underline{0.62}_{\pm0.16}$ & $0.52_{\pm0.03}$ \\
\midrule
\multirow{3}{*}{\parbox{1.9cm}{\raggedright Primary vs.\ Metastasis}}
  & F1        & $\underline{0.88}_{\pm0.02}$ & $0.88_{\pm0.02}$ & $0.86_{\pm0.02}$ & $\mathbf{0.88}_{\pm0.04}$ & $0.87_{\pm0.03}$ & $0.87_{\pm0.06}$ \\
  & AUC       & $0.68_{\pm0.07}$ & $0.64_{\pm0.05}$ & $0.65_{\pm0.10}$ & $\underline{0.75}_{\pm0.05}$ & $\mathbf{0.78}_{\pm0.02}$ & $0.71_{\pm0.07}$ \\
  & Bal.\ Acc & $0.60_{\pm0.07}$ & $0.57_{\pm0.07}$ & $0.56_{\pm0.07}$ & $\underline{0.62}_{\pm0.03}$ & $\mathbf{0.62}_{\pm0.03}$ & $0.61_{\pm0.06}$ \\
\bottomrule
\end{tabular}
}
\end{table}

\begin{table}[tb]
\caption{Linear-probe MIL comparison across sixteen held-out tasks, averaged over MeanMIL, ABMIL, CLAM, DSMIL, and TransMIL.}
\label{tab:linear_probe_mil_results}
\centering
\scriptsize
\setlength{\tabcolsep}{2pt}
\renewcommand{\arraystretch}{0.96}
\resizebox{\linewidth}{!}{%
\begin{tabular}{%
  >{\raggedright\arraybackslash}p{2.1cm}%
  c%
  c%
  c%
  c%
  c%
  c%
  c}
\toprule
Task & Metric & Backbone & \shortstack{UNI\\v2} & \shortstack{Phikon\\v2} & \shortstack{CONCH\\v1.5} & MUSK & \shortstack{MOOZY\\(Ours)} \\
\midrule
\multirow{3}{*}{\parbox{2.1cm}{\raggedright Residual Cancer Burden}}
  & F1 (weighted)      & $\underline{0.46}_{\pm0.05}$ & $0.44_{\pm0.05}$ & $0.42_{\pm0.06}$ & $\mathbf{0.47}_{\pm0.05}$ & $0.44_{\pm0.05}$ & $0.38_{\pm0.11}$ \\
  & ROC-AUC (weighted) & $0.60_{\pm0.06}$ & $0.60_{\pm0.06}$ & $0.59_{\pm0.06}$ & $\underline{0.61}_{\pm0.06}$ & $0.59_{\pm0.07}$ & $\mathbf{0.66}_{\pm0.05}$ \\
  & Balanced Acc       & $\mathbf{0.44}_{\pm0.06}$ & $0.40_{\pm0.06}$ & $0.39_{\pm0.06}$ & $0.42_{\pm0.06}$ & $0.40_{\pm0.06}$ & $\underline{0.44}_{\pm0.07}$ \\
\midrule
\multirow{3}{*}{\parbox{2.1cm}{\raggedright TP53 mutation}}
  & F1 (weighted)      & $0.77_{\pm0.06}$ & $0.77_{\pm0.07}$ & $0.78_{\pm0.07}$ & $\mathbf{0.80}_{\pm0.06}$ & $0.79_{\pm0.06}$ & $\underline{0.80}_{\pm0.06}$ \\
  & ROC-AUC (weighted) & $0.79_{\pm0.07}$ & $0.75_{\pm0.09}$ & $0.78_{\pm0.08}$ & $\underline{0.81}_{\pm0.08}$ & $0.80_{\pm0.06}$ & $\mathbf{0.87}_{\pm0.03}$ \\
  & Balanced Acc       & $0.76_{\pm0.05}$ & $0.77_{\pm0.07}$ & $0.77_{\pm0.07}$ & $\underline{0.79}_{\pm0.07}$ & $0.79_{\pm0.04}$ & $\mathbf{0.81}_{\pm0.05}$ \\
\midrule
\multirow{3}{*}{\parbox{2.1cm}{\raggedright BAP1 mutation}}
  & F1 (weighted)      & $0.84_{\pm0.04}$ & $0.82_{\pm0.05}$ & $0.83_{\pm0.05}$ & $\underline{0.85}_{\pm0.05}$ & $\mathbf{0.86}_{\pm0.06}$ & $0.78_{\pm0.05}$ \\
  & ROC-AUC (weighted) & $0.66_{\pm0.19}$ & $0.65_{\pm0.16}$ & $0.67_{\pm0.12}$ & $\mathbf{0.75}_{\pm0.13}$ & $0.73_{\pm0.13}$ & $\underline{0.74}_{\pm0.11}$ \\
  & Balanced Acc       & $0.67_{\pm0.10}$ & $0.64_{\pm0.10}$ & $0.68_{\pm0.10}$ & $\mathbf{0.74}_{\pm0.09}$ & $\underline{0.73}_{\pm0.10}$ & $0.64_{\pm0.12}$ \\
\midrule
\multirow{3}{*}{\parbox{2.1cm}{\raggedright ACVR2A mutation}}
  & F1 (weighted)      & $\underline{0.83}_{\pm0.09}$ & $\mathbf{0.84}_{\pm0.07}$ & $0.81_{\pm0.11}$ & $0.82_{\pm0.07}$ & $0.79_{\pm0.09}$ & $0.83_{\pm0.04}$ \\
  & ROC-AUC (weighted) & $\underline{0.74}_{\pm0.10}$ & $0.72_{\pm0.16}$ & $0.73_{\pm0.13}$ & $0.70_{\pm0.14}$ & $0.60_{\pm0.19}$ & $\mathbf{0.89}_{\pm0.07}$ \\
  & Balanced Acc       & $0.73_{\pm0.09}$ & $\underline{0.73}_{\pm0.11}$ & $0.68_{\pm0.13}$ & $0.70_{\pm0.10}$ & $0.65_{\pm0.12}$ & $\mathbf{0.78}_{\pm0.10}$ \\
\midrule
\multirow{3}{*}{\parbox{2.1cm}{\raggedright Histologic Grade}}
  & F1 (weighted)      & $0.74_{\pm0.05}$ & $0.74_{\pm0.05}$ & $0.72_{\pm0.05}$ & $\mathbf{0.76}_{\pm0.05}$ & $\underline{0.75}_{\pm0.08}$ & $0.66_{\pm0.08}$ \\
  & ROC-AUC (weighted) & $0.73_{\pm0.09}$ & $0.73_{\pm0.05}$ & $0.72_{\pm0.06}$ & $\mathbf{0.76}_{\pm0.08}$ & $0.74_{\pm0.08}$ & $\underline{0.74}_{\pm0.11}$ \\
  & Balanced Acc       & $0.74_{\pm0.05}$ & $0.75_{\pm0.06}$ & $0.72_{\pm0.05}$ & $\mathbf{0.75}_{\pm0.05}$ & $\underline{0.75}_{\pm0.07}$ & $0.68_{\pm0.08}$ \\
\midrule
\multirow{3}{*}{\parbox{2.1cm}{\raggedright KRAS mutation}}
  & F1 (weighted)      & $\underline{0.78}_{\pm0.07}$ & $0.74_{\pm0.05}$ & $0.74_{\pm0.05}$ & $\mathbf{0.80}_{\pm0.08}$ & $0.76_{\pm0.07}$ & $0.71_{\pm0.08}$ \\
  & ROC-AUC (weighted) & $\mathbf{0.74}_{\pm0.10}$ & $0.71_{\pm0.09}$ & $0.68_{\pm0.07}$ & $\underline{0.74}_{\pm0.12}$ & $0.70_{\pm0.10}$ & $0.73_{\pm0.07}$ \\
  & Balanced Acc       & $\underline{0.76}_{\pm0.09}$ & $0.69_{\pm0.08}$ & $0.68_{\pm0.06}$ & $\mathbf{0.77}_{\pm0.09}$ & $0.72_{\pm0.08}$ & $0.71_{\pm0.09}$ \\
\midrule
\multirow{3}{*}{\parbox{2.1cm}{\raggedright IDH Status}}
  & F1 (weighted)      & $0.93_{\pm0.02}$ & $0.93_{\pm0.02}$ & $0.93_{\pm0.02}$ & $\underline{0.94}_{\pm0.02}$ & $0.92_{\pm0.02}$ & $\mathbf{0.95}_{\pm0.02}$ \\
  & ROC-AUC (weighted) & $0.96_{\pm0.01}$ & $0.97_{\pm0.01}$ & $\underline{0.97}_{\pm0.01}$ & $0.97_{\pm0.01}$ & $0.96_{\pm0.01}$ & $\mathbf{0.99}_{\pm0.01}$ \\
  & Balanced Acc       & $0.93_{\pm0.02}$ & $0.93_{\pm0.02}$ & $0.93_{\pm0.02}$ & $\underline{0.93}_{\pm0.02}$ & $0.92_{\pm0.02}$ & $\mathbf{0.95}_{\pm0.02}$ \\
\midrule
\multirow{3}{*}{\parbox{2.1cm}{\raggedright Treatment Response}}
  & F1 (weighted)      & $0.51_{\pm0.08}$ & $0.45_{\pm0.04}$ & $0.49_{\pm0.08}$ & $\mathbf{0.53}_{\pm0.06}$ & $\underline{0.52}_{\pm0.08}$ & $0.47_{\pm0.11}$ \\
  & ROC-AUC (weighted) & $0.66_{\pm0.10}$ & $0.62_{\pm0.04}$ & $0.65_{\pm0.08}$ & $\underline{0.67}_{\pm0.08}$ & $\mathbf{0.68}_{\pm0.07}$ & $0.60_{\pm0.08}$ \\
  & Balanced Acc       & $0.46_{\pm0.12}$ & $0.37_{\pm0.07}$ & $0.38_{\pm0.08}$ & $\mathbf{0.47}_{\pm0.11}$ & $\underline{0.47}_{\pm0.09}$ & $0.38_{\pm0.10}$ \\
\midrule
\multirow{3}{*}{\parbox{2.1cm}{\raggedright BRCA PAM50 Subtype}}
  & F1 (weighted)      & $0.64_{\pm0.03}$ & $\underline{0.67}_{\pm0.04}$ & $0.65_{\pm0.04}$ & $\mathbf{0.67}_{\pm0.03}$ & $0.64_{\pm0.04}$ & $0.58_{\pm0.05}$ \\
  & ROC-AUC (weighted) & $0.81_{\pm0.03}$ & $\underline{0.82}_{\pm0.03}$ & $0.80_{\pm0.04}$ & $\mathbf{0.83}_{\pm0.03}$ & $0.80_{\pm0.04}$ & $0.82_{\pm0.03}$ \\
  & Balanced Acc       & $0.48_{\pm0.04}$ & $\underline{0.51}_{\pm0.05}$ & $0.48_{\pm0.04}$ & $\mathbf{0.52}_{\pm0.04}$ & $0.47_{\pm0.04}$ & $0.50_{\pm0.06}$ \\
\midrule
\multirow{3}{*}{\parbox{2.1cm}{\raggedright HNSC mRNA Subtype}}
  & F1 (weighted)      & $0.53_{\pm0.04}$ & $\underline{0.54}_{\pm0.04}$ & $0.53_{\pm0.05}$ & $\mathbf{0.58}_{\pm0.05}$ & $0.51_{\pm0.03}$ & $0.45_{\pm0.02}$ \\
  & ROC-AUC (weighted) & $0.71_{\pm0.04}$ & $\underline{0.72}_{\pm0.04}$ & $0.71_{\pm0.04}$ & $\mathbf{0.75}_{\pm0.04}$ & $0.69_{\pm0.04}$ & $0.70_{\pm0.03}$ \\
  & Balanced Acc       & $0.50_{\pm0.05}$ & $\underline{0.53}_{\pm0.05}$ & $0.52_{\pm0.06}$ & $\mathbf{0.57}_{\pm0.05}$ & $0.51_{\pm0.04}$ & $0.46_{\pm0.02}$ \\
\midrule
\multirow{3}{*}{\parbox{2.1cm}{\raggedright UCEC Genomic Subtype}}
  & F1 (weighted)      & $0.52_{\pm0.04}$ & $\underline{0.58}_{\pm0.04}$ & $0.53_{\pm0.04}$ & $\mathbf{0.58}_{\pm0.04}$ & $0.51_{\pm0.03}$ & $0.54_{\pm0.02}$ \\
  & ROC-AUC (weighted) & $0.72_{\pm0.04}$ & $\underline{0.77}_{\pm0.03}$ & $0.74_{\pm0.04}$ & $\mathbf{0.77}_{\pm0.03}$ & $0.71_{\pm0.03}$ & $0.77_{\pm0.03}$ \\
  & Balanced Acc       & $0.50_{\pm0.04}$ & $\underline{0.55}_{\pm0.05}$ & $0.51_{\pm0.04}$ & $\mathbf{0.56}_{\pm0.04}$ & $0.49_{\pm0.03}$ & $0.52_{\pm0.02}$ \\
\midrule
\multirow{3}{*}{\parbox{2.1cm}{\raggedright Synaptophysin Grade}}
  & F1 (weighted)      & $\underline{0.80}_{\pm0.07}$ & $0.78_{\pm0.08}$ & $0.78_{\pm0.06}$ & $0.79_{\pm0.06}$ & $\mathbf{0.80}_{\pm0.05}$ & $0.63_{\pm0.08}$ \\
  & ROC-AUC (weighted) & $\underline{0.65}_{\pm0.10}$ & $0.63_{\pm0.10}$ & $0.65_{\pm0.12}$ & $0.65_{\pm0.13}$ & $\mathbf{0.66}_{\pm0.09}$ & $0.61_{\pm0.09}$ \\
  & Balanced Acc       & $\underline{0.65}_{\pm0.07}$ & $0.64_{\pm0.08}$ & $0.64_{\pm0.08}$ & $0.65_{\pm0.09}$ & $\mathbf{0.67}_{\pm0.07}$ & $0.56_{\pm0.12}$ \\
\midrule
\multirow{3}{*}{\parbox{2.1cm}{\raggedright RAS/BRAF Status}}
  & F1 (weighted)      & $\underline{0.89}_{\pm0.08}$ & $0.87_{\pm0.09}$ & $0.87_{\pm0.08}$ & $\mathbf{0.89}_{\pm0.07}$ & $0.87_{\pm0.08}$ & $0.88_{\pm0.09}$ \\
  & ROC-AUC (weighted) & $\underline{0.62}_{\pm0.19}$ & $0.59_{\pm0.17}$ & $0.52_{\pm0.16}$ & $0.49_{\pm0.20}$ & $0.58_{\pm0.16}$ & $\mathbf{0.73}_{\pm0.23}$ \\
  & Balanced Acc       & $\underline{0.68}_{\pm0.20}$ & $0.65_{\pm0.20}$ & $0.63_{\pm0.19}$ & $0.67_{\pm0.19}$ & $0.63_{\pm0.20}$ & $\mathbf{0.75}_{\pm0.20}$ \\
\midrule
\multirow{3}{*}{\parbox{2.1cm}{\raggedright Keratinizing SCC Grade}}
  & F1 (weighted)      & $\underline{0.68}_{\pm0.03}$ & $\mathbf{0.69}_{\pm0.04}$ & $0.66_{\pm0.04}$ & $0.67_{\pm0.03}$ & $0.68_{\pm0.04}$ & $0.66_{\pm0.03}$ \\
  & ROC-AUC (weighted) & $0.68_{\pm0.05}$ & $\underline{0.69}_{\pm0.05}$ & $0.66_{\pm0.05}$ & $0.66_{\pm0.05}$ & $0.67_{\pm0.06}$ & $\mathbf{0.73}_{\pm0.03}$ \\
  & Balanced Acc       & $\underline{0.67}_{\pm0.04}$ & $\mathbf{0.69}_{\pm0.04}$ & $0.66_{\pm0.04}$ & $0.67_{\pm0.03}$ & $0.67_{\pm0.05}$ & $0.65_{\pm0.04}$ \\
\midrule
\multirow{3}{*}{\parbox{2.1cm}{\raggedright Non-Keratinizing SCC Grade}}
  & F1 (weighted)      & $0.74_{\pm0.14}$ & $0.76_{\pm0.14}$ & $0.76_{\pm0.13}$ & $\mathbf{0.78}_{\pm0.13}$ & $\underline{0.77}_{\pm0.14}$ & $0.55_{\pm0.09}$ \\
  & ROC-AUC (weighted) & $0.60_{\pm0.10}$ & $\underline{0.69}_{\pm0.13}$ & $0.64_{\pm0.12}$ & $\mathbf{0.70}_{\pm0.13}$ & $0.61_{\pm0.12}$ & $0.59_{\pm0.07}$ \\
  & Balanced Acc       & $0.72_{\pm0.15}$ & $0.74_{\pm0.14}$ & $0.74_{\pm0.14}$ & $\mathbf{0.78}_{\pm0.14}$ & $\underline{0.74}_{\pm0.15}$ & $0.52_{\pm0.03}$ \\
\midrule
\multirow{3}{*}{\parbox{2.1cm}{\raggedright Primary vs.\ Metastasis}}
  & F1 (weighted)      & $0.92_{\pm0.02}$ & $\underline{0.92}_{\pm0.02}$ & $\mathbf{0.92}_{\pm0.02}$ & $0.92_{\pm0.02}$ & $0.92_{\pm0.02}$ & $0.87_{\pm0.06}$ \\
  & ROC-AUC (weighted) & $0.64_{\pm0.08}$ & $0.64_{\pm0.10}$ & $0.64_{\pm0.10}$ & $\underline{0.67}_{\pm0.07}$ & $0.60_{\pm0.08}$ & $\mathbf{0.71}_{\pm0.07}$ \\
  & Balanced Acc       & $0.59_{\pm0.04}$ & $0.59_{\pm0.05}$ & $0.60_{\pm0.05}$ & $\underline{0.61}_{\pm0.05}$ & $0.58_{\pm0.04}$ & $\mathbf{0.61}_{\pm0.06}$ \\
\bottomrule
\end{tabular}
}
\end{table}

\section{Encoder Parameter Comparison}
\label[appendix]{appendix_sec:encoder_parameter_comparison}

\Cref{tab:encoder_parameter_comparison} reports the parameter counts for all compared slide encoders, split into slide-encoder parameters, patch-encoder parameters, and total parameters. This breakdown makes the capacity allocation explicit: most baselines place the majority of parameters in the patch encoder, while MOOZY uses a comparatively lightweight patch encoder and a compact case-level slide stack. In absolute model size, MOOZY (85.77M total) is substantially smaller than GigaPath (1.22B), PRISM (742.06M), Madeleine (400.23M), and TITAN (354.65M). CHIEF has fewer total parameters, while MOOZY provides stronger macro weighted F1, weighted ROC-AUC, and balanced accuracy.

\begin{table}[tb]
\caption{Parameter count comparison across slide encoders. MOOZY has 85.77M parameters and achieves the strongest macro weighted F1 and balanced accuracy (\Cref{tab:encoder_comparison}).}
\label{tab:encoder_parameter_comparison}
\centering
\fontsize{6.6}{7.1}\selectfont
\setlength{\tabcolsep}{2.2pt}
\renewcommand{\arraystretch}{0.86}
\resizebox{0.96\linewidth}{!}{%
\begin{tabular}{lccc}
\toprule
Encoder & Slide Encoder Parameters & Patch Encoder Parameters & Total Parameters \\
\midrule
CHIEF    & 1.19M  & 27.52M  & 28.71M  \\ 
GigaPath & 85.15M & 1.13B   & 1.22B   \\ 
PRISM    & 110.83M & 631.23M & 742.06M \\ 
Madeleine & 5.00M & 395.23M & 400.23M \\ 
TITAN    & 48.54M & 306.11M & 354.65M \\ 
MOOZY (Ours) & 42.8M (slide) + 21.3M (case) & 21.67M & 85.77M \\ 
\bottomrule
\end{tabular}%
}
\end{table}

\section{MIL-Specific Task-wise Comparison}
\label[appendix]{appendix_sec:mil_specific_taskwise}

\Cref{tab:mil_comparison_macro} in the main paper reports macro-average results over sixteen held-out tasks. \Cref{tab:mil_comparison} provides the full per-task breakdown averaged over all five MIL architectures. \Cref{tab:mil_meanmil_taskwise,tab:mil_abmil_taskwise,tab:mil_clam_taskwise,tab:mil_dsmil_taskwise,tab:mil_transmil_taskwise} provide the full per-method per-task breakdown. MOOZY remains strongest on the macro metrics, while task-wise results show architecture-dependent variation across the benchmark, particularly for Synaptophysin Grade and Non-Keratinizing SCC Grade.

\begin{table}[tb]
\caption{Comparison of MOOZY against patch encoder baselines with trained MIL aggregators on sixteen held-out tasks. MIL baselines train a task-specific aggregator from scratch on frozen patch features; MOOZY is entirely frozen and evaluated with an MLP probe. Each patch encoder entry is the arithmetic mean over five MIL architectures (MeanMIL, ABMIL, CLAM, DSMIL, TransMIL). \textbf{Bold}: best; \underline{underline}: second best.}
\label{tab:mil_comparison}
\centering
\scriptsize
\setlength{\tabcolsep}{2pt}
\renewcommand{\arraystretch}{0.94}
\resizebox{\linewidth}{!}{%
\begin{tabular}{%
  >{\raggedright\arraybackslash}p{2.1cm}%
  c%
  c%
  c%
  c%
  c%
  c%
  c}
\toprule
Task & Metric & Backbone & \shortstack{UNI\\v2} & \shortstack{Phikon\\v2} & \shortstack{CONCH\\v1.5} & MUSK & \shortstack{MOOZY\\(Ours)} \\
\midrule
\multirow{3}{*}{\parbox{2.1cm}{\raggedright Residual Cancer Burden}}
  & F1 (weighted)      & $0.46_{\pm0.05}$ & $0.44_{\pm0.05}$ & $0.42_{\pm0.06}$ & $\underline{0.47}_{\pm0.05}$ & $0.44_{\pm0.05}$ & $\mathbf{0.56}_{\pm0.05}$ \\
  & ROC-AUC (weighted) & $0.60_{\pm0.06}$ & $0.60_{\pm0.06}$ & $0.59_{\pm0.06}$ & $\underline{0.61}_{\pm0.06}$ & $0.59_{\pm0.07}$ & $\mathbf{0.74}_{\pm0.04}$ \\
  & Balanced Acc       & $\underline{0.44}_{\pm0.06}$ & $0.40_{\pm0.06}$ & $0.39_{\pm0.06}$ & $0.42_{\pm0.06}$ & $0.40_{\pm0.06}$ & $\mathbf{0.51}_{\pm0.06}$ \\
\midrule
\multirow{3}{*}{\parbox{2.1cm}{\raggedright TP53 mutation}}
  & F1 (weighted)      & $0.77_{\pm0.06}$ & $0.77_{\pm0.07}$ & $0.78_{\pm0.07}$ & $\underline{0.80}_{\pm0.06}$ & $0.79_{\pm0.06}$ & $\mathbf{0.87}_{\pm0.04}$ \\
  & ROC-AUC (weighted) & $0.79_{\pm0.07}$ & $0.75_{\pm0.09}$ & $0.78_{\pm0.08}$ & $\underline{0.81}_{\pm0.08}$ & $0.80_{\pm0.06}$ & $\mathbf{0.86}_{\pm0.06}$ \\
  & Balanced Acc       & $0.76_{\pm0.05}$ & $0.77_{\pm0.07}$ & $0.77_{\pm0.07}$ & $\underline{0.79}_{\pm0.07}$ & $0.79_{\pm0.04}$ & $\mathbf{0.86}_{\pm0.05}$ \\
\midrule
\multirow{3}{*}{\parbox{2.1cm}{\raggedright BAP1 mutation}}
  & F1 (weighted)      & $0.84_{\pm0.04}$ & $0.82_{\pm0.05}$ & $0.83_{\pm0.05}$ & $0.85_{\pm0.05}$ & $\underline{0.86}_{\pm0.06}$ & $\mathbf{0.89}_{\pm0.06}$ \\
\bottomrule
\end{tabular}%
}
\end{table}

\begin{table}[tb]
\caption{Comparison of MOOZY against patch encoder baselines using MeanMIL on sixteen held-out tasks. MIL baselines train a task-specific MeanMIL aggregator from scratch on frozen patch features; MOOZY is entirely frozen and evaluated with an MLP probe. Values are mean $\pm$ standard deviation across five folds. \textbf{Bold}: best; \underline{underline}: second best.}
\label{tab:mil_meanmil_taskwise}
\centering
\scriptsize
\setlength{\tabcolsep}{2pt}
\renewcommand{\arraystretch}{0.96}
\resizebox{\linewidth}{!}{%
\begin{tabular}{%
  >{\raggedright\arraybackslash}p{2.1cm}%
  c%
  c%
  c%
  c%
  c%
  c%
  c}
\toprule
Task & Metric & Backbone & \shortstack{UNI\\v2} & \shortstack{Phikon\\v2} & \shortstack{CONCH\\v1.5} & MUSK & \shortstack{MOOZY\\(Ours)} \\
\midrule
\multirow{3}{*}{\parbox{2.1cm}{\raggedright Residual Cancer Burden}}
  & F1 (weighted)      & $0.46_{\pm0.04}$ & $0.45_{\pm0.07}$ & $0.42_{\pm0.06}$ & $0.45_{\pm0.06}$ & $\underline{0.46}_{\pm0.08}$ & $\mathbf{0.56}_{\pm0.05}$ \\
  & ROC-AUC (weighted) & $0.58_{\pm0.07}$ & $0.58_{\pm0.09}$ & $0.59_{\pm0.08}$ & $\underline{0.60}_{\pm0.06}$ & $0.60_{\pm0.08}$ & $\mathbf{0.74}_{\pm0.04}$ \\
  & Balanced Acc       & $\underline{0.44}_{\pm0.04}$ & $0.39_{\pm0.09}$ & $0.38_{\pm0.07}$ & $0.39_{\pm0.04}$ & $0.42_{\pm0.09}$ & $\mathbf{0.51}_{\pm0.06}$ \\
\midrule
\multirow{3}{*}{\parbox{2.1cm}{\raggedright TP53 mutation}}
  & F1 (weighted)      & $0.80_{\pm0.06}$ & $0.80_{\pm0.05}$ & $0.82_{\pm0.06}$ & $\underline{0.84}_{\pm0.03}$ & $0.79_{\pm0.04}$ & $\mathbf{0.87}_{\pm0.04}$ \\
  & ROC-AUC (weighted) & $0.80_{\pm0.07}$ & $0.78_{\pm0.07}$ & $0.82_{\pm0.04}$ & $\underline{0.85}_{\pm0.03}$ & $0.80_{\pm0.02}$ & $\mathbf{0.86}_{\pm0.06}$ \\
  & Balanced Acc       & $0.80_{\pm0.04}$ & $0.80_{\pm0.04}$ & $0.81_{\pm0.06}$ & $\underline{0.83}_{\pm0.03}$ & $0.78_{\pm0.02}$ & $\mathbf{0.86}_{\pm0.05}$ \\
\midrule
\multirow{3}{*}{\parbox{2.1cm}{\raggedright BAP1 mutation}}
  & F1 (weighted)      & $0.87_{\pm0.06}$ & $0.84_{\pm0.06}$ & $0.84_{\pm0.06}$ & $0.83_{\pm0.07}$ & $\mathbf{0.89}_{\pm0.04}$ & $\underline{0.89}_{\pm0.06}$ \\
\bottomrule
\end{tabular}%
}
\end{table}

\begin{table}[tb]
\caption{Comparison of MOOZY against patch encoder baselines using ABMIL on sixteen held-out tasks. MIL baselines train a task-specific ABMIL aggregator from scratch on frozen patch features; MOOZY is entirely frozen and evaluated with an MLP probe. Values are mean $\pm$ standard deviation across five folds. \textbf{Bold}: best; \underline{underline}: second best.}
\label{tab:mil_abmil_taskwise}
\centering
\scriptsize
\setlength{\tabcolsep}{2pt}
\renewcommand{\arraystretch}{0.96}
\resizebox{\linewidth}{!}{%
\begin{tabular}{%
  >{\raggedright\arraybackslash}p{2.1cm}%
  c%
  c%
  c%
  c%
  c%
  c%
  c}
\toprule
Task & Metric & Backbone & \shortstack{UNI\\v2} & \shortstack{Phikon\\v2} & \shortstack{CONCH\\v1.5} & MUSK & \shortstack{MOOZY\\(Ours)} \\
\midrule
\multirow{3}{*}{\parbox{2.1cm}{\raggedright Residual Cancer Burden}}
  & F1 (weighted)      & $0.47_{\pm0.05}$ & $\underline{0.49}_{\pm0.05}$ & $0.42_{\pm0.09}$ & $0.49_{\pm0.03}$ & $0.47_{\pm0.03}$ & $\mathbf{0.56}_{\pm0.05}$ \\
  & ROC-AUC (weighted) & $0.65_{\pm0.04}$ & $0.65_{\pm0.03}$ & $0.60_{\pm0.07}$ & $0.62_{\pm0.05}$ & $\underline{0.65}_{\pm0.06}$ & $\mathbf{0.74}_{\pm0.04}$ \\
  & Balanced Acc       & $\underline{0.47}_{\pm0.05}$ & $0.45_{\pm0.05}$ & $0.39_{\pm0.06}$ & $0.42_{\pm0.06}$ & $0.44_{\pm0.04}$ & $\mathbf{0.51}_{\pm0.06}$ \\
\midrule
\multirow{3}{*}{\parbox{2.1cm}{\raggedright TP53 mutation}}
  & F1 (weighted)      & $0.80_{\pm0.08}$ & $\underline{0.82}_{\pm0.08}$ & $0.79_{\pm0.04}$ & $0.80_{\pm0.11}$ & $0.82_{\pm0.04}$ & $\mathbf{0.87}_{\pm0.04}$ \\
  & ROC-AUC (weighted) & $0.80_{\pm0.05}$ & $0.81_{\pm0.09}$ & $0.82_{\pm0.06}$ & $0.82_{\pm0.12}$ & $\underline{0.82}_{\pm0.03}$ & $\mathbf{0.86}_{\pm0.06}$ \\
  & Balanced Acc       & $0.80_{\pm0.06}$ & $\underline{0.83}_{\pm0.09}$ & $0.79_{\pm0.04}$ & $0.79_{\pm0.11}$ & $0.81_{\pm0.04}$ & $\mathbf{0.86}_{\pm0.05}$ \\
\midrule
\multirow{3}{*}{\parbox{2.1cm}{\raggedright BAP1 mutation}}
  & F1 (weighted)      & $0.84_{\pm0.03}$ & $0.82_{\pm0.05}$ & $0.82_{\pm0.03}$ & $\underline{0.88}_{\pm0.04}$ & $0.87_{\pm0.05}$ & $\mathbf{0.89}_{\pm0.06}$ \\
\bottomrule
\end{tabular}%
}
\end{table}

\begin{table}[tb]
\caption{Comparison of MOOZY against patch encoder baselines using CLAM on sixteen held-out tasks. MIL baselines train a task-specific CLAM aggregator from scratch on frozen patch features; MOOZY is entirely frozen and evaluated with an MLP probe. Values are mean $\pm$ standard deviation across five folds. \textbf{Bold}: best; \underline{underline}: second best.}
\label{tab:mil_clam_taskwise}
\centering
\scriptsize
\setlength{\tabcolsep}{2pt}
\renewcommand{\arraystretch}{0.96}
\resizebox{\linewidth}{!}{%
\begin{tabular}{%
  >{\raggedright\arraybackslash}p{2.1cm}%
  c%
  c%
  c%
  c%
  c%
  c%
  c}
\toprule
Task & Metric & Backbone & \shortstack{UNI\\v2} & \shortstack{Phikon\\v2} & \shortstack{CONCH\\v1.5} & MUSK & \shortstack{MOOZY\\(Ours)} \\
\midrule
\multirow{3}{*}{\parbox{2.1cm}{\raggedright Residual Cancer Burden}}
  & F1 (weighted)      & $0.45_{\pm0.04}$ & $\underline{0.48}_{\pm0.05}$ & $0.43_{\pm0.08}$ & $0.45_{\pm0.05}$ & $0.45_{\pm0.04}$ & $\mathbf{0.56}_{\pm0.05}$ \\
  & ROC-AUC (weighted) & $0.60_{\pm0.09}$ & $\underline{0.66}_{\pm0.06}$ & $0.59_{\pm0.05}$ & $0.60_{\pm0.06}$ & $0.60_{\pm0.05}$ & $\mathbf{0.74}_{\pm0.04}$ \\
  & Balanced Acc       & $\underline{0.43}_{\pm0.08}$ & $0.42_{\pm0.04}$ & $0.43_{\pm0.07}$ & $0.41_{\pm0.08}$ & $0.43_{\pm0.06}$ & $\mathbf{0.51}_{\pm0.06}$ \\
\midrule
\multirow{3}{*}{\parbox{2.1cm}{\raggedright TP53 mutation}}
  & F1 (weighted)      & $0.83_{\pm0.07}$ & $0.81_{\pm0.07}$ & $0.82_{\pm0.06}$ & $\underline{0.85}_{\pm0.06}$ & $0.84_{\pm0.05}$ & $\mathbf{0.87}_{\pm0.04}$ \\
  & ROC-AUC (weighted) & $0.83_{\pm0.07}$ & $0.79_{\pm0.10}$ & $0.85_{\pm0.05}$ & $\mathbf{0.88}_{\pm0.06}$ & $0.83_{\pm0.02}$ & $\underline{0.86}_{\pm0.06}$ \\
  & Balanced Acc       & $0.83_{\pm0.07}$ & $0.81_{\pm0.07}$ & $0.82_{\pm0.05}$ & $\underline{0.86}_{\pm0.07}$ & $0.84_{\pm0.04}$ & $\mathbf{0.86}_{\pm0.05}$ \\
\midrule
\multirow{3}{*}{\parbox{2.1cm}{\raggedright BAP1 mutation}}
  & F1 (weighted)      & $0.86_{\pm0.05}$ & $0.83_{\pm0.07}$ & $0.85_{\pm0.05}$ & $0.85_{\pm0.05}$ & $\underline{0.88}_{\pm0.08}$ & $\mathbf{0.89}_{\pm0.06}$ \\
\bottomrule
\end{tabular}%
}
\end{table}

\begin{table}[tb]
\caption{Comparison of MOOZY against patch encoder baselines using DSMIL on sixteen held-out tasks. MIL baselines train a task-specific DSMIL aggregator from scratch on frozen patch features; MOOZY is entirely frozen and evaluated with an MLP probe. Values are mean $\pm$ standard deviation across five folds. \textbf{Bold}: best; \underline{underline}: second best.}
\label{tab:mil_dsmil_taskwise}
\centering
\scriptsize
\setlength{\tabcolsep}{2pt}
\renewcommand{\arraystretch}{0.96}
\resizebox{\linewidth}{!}{%
\begin{tabular}{%
  >{\raggedright\arraybackslash}p{2.1cm}%
  c%
  c%
  c%
  c%
  c%
  c%
  c}
\toprule
Task & Metric & Backbone & \shortstack{UNI\\v2} & \shortstack{Phikon\\v2} & \shortstack{CONCH\\v1.5} & MUSK & \shortstack{MOOZY\\(Ours)} \\
\midrule
\multirow{3}{*}{\parbox{2.1cm}{\raggedright Residual Cancer Burden}}
  & F1 (weighted)      & $0.46_{\pm0.06}$ & $0.45_{\pm0.02}$ & $0.45_{\pm0.05}$ & $\underline{0.47}_{\pm0.04}$ & $0.45_{\pm0.04}$ & $\mathbf{0.56}_{\pm0.05}$ \\
  & ROC-AUC (weighted) & $0.61_{\pm0.06}$ & $\underline{0.63}_{\pm0.04}$ & $0.59_{\pm0.06}$ & $0.60_{\pm0.07}$ & $0.58_{\pm0.09}$ & $\mathbf{0.74}_{\pm0.04}$ \\
  & Balanced Acc       & $\underline{0.46}_{\pm0.05}$ & $0.45_{\pm0.06}$ & $0.42_{\pm0.04}$ & $0.43_{\pm0.05}$ & $0.37_{\pm0.10}$ & $\mathbf{0.51}_{\pm0.06}$ \\
\midrule
\multirow{3}{*}{\parbox{2.1cm}{\raggedright TP53 mutation}}
  & F1 (weighted)      & $0.58_{\pm0.04}$ & $0.70_{\pm0.06}$ & $0.79_{\pm0.10}$ & $0.78_{\pm0.02}$ & $\underline{0.79}_{\pm0.07}$ & $\mathbf{0.87}_{\pm0.04}$ \\
  & ROC-AUC (weighted) & $0.67_{\pm0.08}$ & $0.66_{\pm0.06}$ & $0.78_{\pm0.14}$ & $0.78_{\pm0.08}$ & $\underline{0.84}_{\pm0.08}$ & $\mathbf{0.86}_{\pm0.06}$ \\
  & Balanced Acc       & $0.56_{\pm0.05}$ & $0.70_{\pm0.06}$ & $0.78_{\pm0.10}$ & $0.77_{\pm0.02}$ & $\underline{0.78}_{\pm0.06}$ & $\mathbf{0.86}_{\pm0.05}$ \\
\midrule
\multirow{3}{*}{\parbox{2.1cm}{\raggedright BAP1 mutation}}
  & F1 (weighted)      & $0.82_{\pm0.03}$ & $0.78_{\pm0.02}$ & $0.80_{\pm0.06}$ & $0.83_{\pm0.05}$ & $\underline{0.85}_{\pm0.03}$ & $\mathbf{0.89}_{\pm0.06}$ \\
\bottomrule
\end{tabular}%
}
\end{table}

\begin{table}[tb]
\caption{Comparison of MOOZY against patch encoder baselines using TransMIL on sixteen held-out tasks. MIL baselines train a task-specific TransMIL aggregator from scratch on frozen patch features; MOOZY is entirely frozen and evaluated with an MLP probe. Values are mean $\pm$ standard deviation across five folds. \textbf{Bold}: best; \underline{underline}: second best.}
\label{tab:mil_transmil_taskwise}
\centering
\scriptsize
\setlength{\tabcolsep}{2pt}
\renewcommand{\arraystretch}{0.96}
\resizebox{\linewidth}{!}{%
\begin{tabular}{%
  >{\raggedright\arraybackslash}p{2.1cm}%
  c%
  c%
  c%
  c%
  c%
  c%
  c}
\toprule
Task & Metric & Backbone & \shortstack{UNI\\v2} & \shortstack{Phikon\\v2} & \shortstack{CONCH\\v1.5} & MUSK & \shortstack{MOOZY\\(Ours)} \\
\midrule
\multirow{3}{*}{\parbox{2.1cm}{\raggedright Residual Cancer Burden}}
  & F1 (weighted)      & $0.45_{\pm0.04}$ & $0.36_{\pm0.04}$ & $0.38_{\pm0.05}$ & $\underline{0.49}_{\pm0.05}$ & $0.36_{\pm0.08}$ & $\mathbf{0.56}_{\pm0.05}$ \\
  & ROC-AUC (weighted) & $0.57_{\pm0.05}$ & $0.47_{\pm0.07}$ & $0.57_{\pm0.05}$ & $\underline{0.62}_{\pm0.05}$ & $0.53_{\pm0.06}$ & $\mathbf{0.74}_{\pm0.04}$ \\
  & Balanced Acc       & $0.41_{\pm0.09}$ & $0.30_{\pm0.05}$ & $0.35_{\pm0.04}$ & $\underline{0.46}_{\pm0.09}$ & $0.33_{\pm0.03}$ & $\mathbf{0.51}_{\pm0.06}$ \\
\midrule
\multirow{3}{*}{\parbox{2.1cm}{\raggedright TP53 mutation}}
  & F1 (weighted)      & $\underline{0.84}_{\pm0.04}$ & $0.74_{\pm0.09}$ & $0.70_{\pm0.09}$ & $0.74_{\pm0.08}$ & $0.73_{\pm0.09}$ & $\mathbf{0.87}_{\pm0.04}$ \\
  & ROC-AUC (weighted) & $\underline{0.83}_{\pm0.07}$ & $0.74_{\pm0.15}$ & $0.65_{\pm0.13}$ & $0.73_{\pm0.12}$ & $0.72_{\pm0.13}$ & $\mathbf{0.86}_{\pm0.06}$ \\
  & Balanced Acc       & $\underline{0.82}_{\pm0.05}$ & $0.72_{\pm0.07}$ & $0.67_{\pm0.09}$ & $0.72_{\pm0.10}$ & $0.72_{\pm0.06}$ & $\mathbf{0.86}_{\pm0.05}$ \\
\midrule
\multirow{3}{*}{\parbox{2.1cm}{\raggedright BAP1 mutation}}
  & F1 (weighted)      & $0.84_{\pm0.04}$ & $0.83_{\pm0.07}$ & $0.86_{\pm0.06}$ & $\underline{0.86}_{\pm0.04}$ & $0.83_{\pm0.07}$ & $\mathbf{0.89}_{\pm0.06}$ \\
\bottomrule
\end{tabular}%
}
\end{table}

\section{Stage 1 Only and MOOZY Task-wise Comparison}
\label[appendix]{appendix_sec:stage1_moozy_comparison}

\Cref{tab:stage1_moozy_taskwise} provides the complete task-level breakdown for the multi-stage ablation discussed in \Cref{sec:results}. The table reports five-fold mean and standard deviation for Stage~1 and MOOZY, together with relative gains for weighted F1, weighted ROC-AUC, and balanced accuracy. Across sixteen tasks, MOOZY improves weighted F1 on eleven, AUC on fifteen, and balanced accuracy on fourteen; eleven tasks improve on all three metrics. The largest gains are UCEC Genomic Subtype F1 (+12.00\%), BAP1 mutation AUC (+19.70\%), and ACVR2A mutation balanced accuracy (+30.43\%). KRAS mutation ties in F1 and favors Stage~1 in AUC and balanced accuracy. BRCA PAM50 Subtype, HNSC mRNA Subtype, Synaptophysin Grade, and Non-Keratinizing SCC Grade show small F1 regressions, with Non-Keratinizing SCC Grade also lower in balanced accuracy. Macro averages improve by 3.50\% weighted F1, 6.64\% AUC, and 5.98\% balanced accuracy.

\begin{table}[tb]
\centering
\caption{Task-wise comparison between Stage~1 and MOOZY (Ours) across the sixteen slide-encoder evaluation tasks. Values are mean $\pm$ standard deviation across five folds. Relative improvement is computed as $(\text{MOOZY} - \text{Stage~1}) / \text{Stage~1} \times 100$ using fold means.}
\label{tab:stage1_moozy_taskwise}
\small
\setlength{\tabcolsep}{3pt}
\resizebox{\textwidth}{!}{%
\begin{tabular}{l c c c c c c c c c}
\toprule
Task & F1 (S1) & F1 (MOOZY) & $\Delta$F1 (\%) & AUC (S1) & AUC (MOOZY) & $\Delta$AUC (\%) & Bal.\ Acc (S1) & Bal.\ Acc (MOOZY) & $\Delta$Bal.\ Acc (\%) \\
\midrule
Residual Cancer Burden & $0.51_{\pm0.05}$ & $0.56_{\pm0.05}$ & +9.80 & $0.63_{\pm0.06}$ & $0.74_{\pm0.04}$ & +17.46 & $0.48_{\pm0.06}$ & $0.51_{\pm0.06}$ & +6.25 \\
TP53 mutation & $0.81_{\pm0.07}$ & $0.87_{\pm0.04}$ & +7.41 & $0.79_{\pm0.07}$ & $0.86_{\pm0.06}$ & +8.86 & $0.80_{\pm0.06}$ & $0.86_{\pm0.05}$ & +7.50 \\
BAP1 mutation & $0.83_{\pm0.07}$ & $0.89_{\pm0.06}$ & +7.23 & $0.66_{\pm0.23}$ & $0.79_{\pm0.12}$ & +19.70 & $0.66_{\pm0.12}$ & $0.78_{\pm0.11}$ & +18.18 \\
ACVR2A mutation & $0.84_{\pm0.08}$ & $0.91_{\pm0.05}$ & +8.33 & $0.77_{\pm0.12}$ & $0.91_{\pm0.09}$ & +18.18 & $0.69_{\pm0.13}$ & $0.90_{\pm0.10}$ & +30.43 \\
Histologic Grade & $0.76_{\pm0.05}$ & $0.78_{\pm0.08}$ & +2.63 & $0.73_{\pm0.05}$ & $0.75_{\pm0.15}$ & +2.74 & $0.76_{\pm0.06}$ & $0.77_{\pm0.08}$ & +1.32 \\
KRAS mutation & $0.85_{\pm0.08}$ & $0.85_{\pm0.04}$ & +0.00 & $0.82_{\pm0.10}$ & $0.80_{\pm0.06}$ & -2.44 & $0.83_{\pm0.07}$ & $0.79_{\pm0.10}$ & -4.82 \\
IDH Status & $0.93_{\pm0.01}$ & $0.97_{\pm0.02}$ & +4.30 & $0.96_{\pm0.01}$ & $0.99_{\pm0.01}$ & +3.13 & $0.92_{\pm0.01}$ & $0.97_{\pm0.02}$ & +5.43 \\
Treatment Response & $0.55_{\pm0.09}$ & $0.58_{\pm0.14}$ & +5.45 & $0.66_{\pm0.09}$ & $0.68_{\pm0.07}$ & +3.03 & $0.47_{\pm0.14}$ & $0.48_{\pm0.17}$ & +2.13 \\
BRCA PAM50 Subtype & $0.64_{\pm0.03}$ & $0.63_{\pm0.03}$ & -1.56 & $0.79_{\pm0.03}$ & $0.80_{\pm0.01}$ & +1.27 & $0.47_{\pm0.06}$ & $0.50_{\pm0.05}$ & +6.38 \\
HNSC mRNA Subtype & $0.56_{\pm0.08}$ & $0.55_{\pm0.03}$ & -1.79 & $0.71_{\pm0.04}$ & $0.72_{\pm0.01}$ & +1.41 & $0.53_{\pm0.09}$ & $0.54_{\pm0.04}$ & +1.89 \\
UCEC Genomic Subtype & $0.50_{\pm0.04}$ & $0.56_{\pm0.05}$ & +12.00 & $0.70_{\pm0.05}$ & $0.74_{\pm0.04}$ & +5.71 & $0.47_{\pm0.03}$ & $0.52_{\pm0.05}$ & +10.64 \\
Synaptophysin Grade & $0.79_{\pm0.05}$ & $0.79_{\pm0.08}$ & -0.11 & $0.57_{\pm0.18}$ & $0.61_{\pm0.15}$ & +7.01 & $0.64_{\pm0.04}$ & $0.65_{\pm0.08}$ & +2.07 \\
RAS/BRAF Status & $0.93_{\pm0.05}$ & $0.94_{\pm0.08}$ & +1.65 & $0.71_{\pm0.18}$ & $0.75_{\pm0.26}$ & +5.82 & $0.81_{\pm0.15}$ & $0.85_{\pm0.20}$ & +4.79 \\
Keratinizing SCC Grade & $0.70_{\pm0.05}$ & $0.73_{\pm0.03}$ & +3.43 & $0.71_{\pm0.06}$ & $0.72_{\pm0.06}$ & +2.08 & $0.69_{\pm0.05}$ & $0.72_{\pm0.03}$ & +3.71 \\
Non-Keratinizing SCC Grade & $0.77_{\pm0.12}$ & $0.77_{\pm0.14}$ & -1.07 & $0.59_{\pm0.11}$ & $0.66_{\pm0.07}$ & +10.56 & $0.75_{\pm0.13}$ & $0.73_{\pm0.15}$ & -1.69 \\
Primary vs.\ Metastasis & $0.92_{\pm0.01}$ & $0.93_{\pm0.02}$ & +1.24 & $0.65_{\pm0.08}$ & $0.69_{\pm0.13}$ & +6.19 & $0.62_{\pm0.08}$ & $0.65_{\pm0.10}$ & +4.67 \\
\midrule
Macro average & 0.743 & 0.769 & +3.50 & 0.715 & 0.763 & +6.64 & 0.662 & 0.702 & +5.98 \\
\bottomrule
\end{tabular}%
}
\end{table}

\section{Stage 2 Only and MOOZY Task-wise Comparison}
\label[appendix]{appendix_sec:stage2_moozy_comparison}

\Cref{tab:stage2_moozy_taskwise} provides the complete task-level breakdown for the Stage~2 only versus MOOZY comparison discussed in \Cref{sec:results}. Stage~2 only trains the slide encoder with multi-task supervision from scratch, without Stage~1 SSL pretraining. MOOZY improves all three metrics on fourteen of sixteen tasks. The largest gains are Treatment Response F1 (+18.37\%) and balanced accuracy (+17.07\%), and RAS/BRAF Status AUC (+32.76\%). Stage~2 only remains stronger across all three metrics on UCEC Genomic Subtype and Synaptophysin Grade. Macro averages improve by 5.23\% weighted F1, 9.45\% AUC, and 6.47\% balanced accuracy.

\begin{table}[tb]
\centering
\caption{Task-wise comparison between Stage~2 only and MOOZY (Ours) across the sixteen slide-encoder evaluation tasks. Stage~2 only trains the slide encoder with multi-task supervision but without Stage~1 SSL pretraining. Values are mean $\pm$ standard deviation across five folds. Relative improvement is computed as $(\text{MOOZY} - \text{Stage~2 only}) / \text{Stage~2 only} \times 100$ using fold means.}
\label{tab:stage2_moozy_taskwise}
\small
\setlength{\tabcolsep}{3pt}
\resizebox{\textwidth}{!}{%
\begin{tabular}{l c c c c c c c c c}
\toprule
Task & F1 (S2) & F1 (MOOZY) & $\Delta$F1 (\%) & AUC (S2) & AUC (MOOZY) & $\Delta$AUC (\%) & Bal.\ Acc (S2) & Bal.\ Acc (MOOZY) & $\Delta$Bal.\ Acc (\%) \\
\midrule
Residual Cancer Burden & $0.50_{\pm0.02}$ & $0.56_{\pm0.05}$ & +12.00 & $0.65_{\pm0.02}$ & $0.74_{\pm0.04}$ & +13.85 & $0.47_{\pm0.05}$ & $0.51_{\pm0.06}$ & +8.51 \\
TP53 mutation          & $0.77_{\pm0.07}$ & $0.87_{\pm0.04}$ & +12.99 & $0.75_{\pm0.08}$ & $0.86_{\pm0.06}$ & +14.67 & $0.76_{\pm0.07}$ & $0.86_{\pm0.05}$ & +13.16 \\
BAP1 mutation          & $0.85_{\pm0.04}$ & $0.89_{\pm0.06}$ & +4.71  & $0.72_{\pm0.17}$ & $0.79_{\pm0.12}$ & +9.72  & $0.74_{\pm0.13}$ & $0.78_{\pm0.11}$ & +5.41 \\
ACVR2A mutation        & $0.89_{\pm0.04}$ & $0.91_{\pm0.05}$ & +2.25  & $0.79_{\pm0.12}$ & $0.91_{\pm0.09}$ & +15.19 & $0.81_{\pm0.16}$ & $0.90_{\pm0.10}$ & +11.11 \\
Histologic Grade       & $0.76_{\pm0.05}$ & $0.78_{\pm0.08}$ & +2.63  & $0.71_{\pm0.09}$ & $0.75_{\pm0.15}$ & +5.63  & $0.75_{\pm0.05}$ & $0.77_{\pm0.08}$ & +2.67 \\
KRAS mutation          & $0.77_{\pm0.08}$ & $0.85_{\pm0.04}$ & +10.39 & $0.62_{\pm0.07}$ & $0.80_{\pm0.06}$ & +29.03 & $0.72_{\pm0.07}$ & $0.79_{\pm0.10}$ & +9.72 \\
IDH Status             & $0.95_{\pm0.01}$ & $0.97_{\pm0.02}$ & +2.11  & $0.97_{\pm0.01}$ & $0.99_{\pm0.01}$ & +2.06  & $0.95_{\pm0.01}$ & $0.97_{\pm0.02}$ & +2.11 \\
Treatment Response     & $0.49_{\pm0.07}$ & $0.58_{\pm0.14}$ & +18.37 & $0.59_{\pm0.06}$ & $0.68_{\pm0.07}$ & +15.25 & $0.41_{\pm0.11}$ & $0.48_{\pm0.17}$ & +17.07 \\
BRCA PAM50 Subtype     & $0.58_{\pm0.02}$ & $0.63_{\pm0.03}$ & +8.62  & $0.78_{\pm0.03}$ & $0.80_{\pm0.01}$ & +2.56  & $0.47_{\pm0.05}$ & $0.50_{\pm0.05}$ & +6.38 \\
HNSC mRNA Subtype      & $0.50_{\pm0.02}$ & $0.55_{\pm0.03}$ & +10.00 & $0.69_{\pm0.04}$ & $0.72_{\pm0.01}$ & +4.35  & $0.50_{\pm0.04}$ & $0.54_{\pm0.04}$ & +8.00 \\
UCEC Genomic Subtype   & $0.57_{\pm0.05}$ & $0.56_{\pm0.05}$ & -1.75  & $0.75_{\pm0.04}$ & $0.74_{\pm0.04}$ & -1.33  & $0.54_{\pm0.04}$ & $0.52_{\pm0.05}$ & -3.70 \\
Synaptophysin Grade & $0.79_{\pm0.08}$ & $0.79_{\pm0.08}$ & -1.07 & $0.66_{\pm0.16}$ & $0.61_{\pm0.15}$ & -7.09 & $0.70_{\pm0.12}$ & $0.65_{\pm0.08}$ & -7.08 \\
RAS/BRAF Status & $0.90_{\pm0.08}$ & $0.94_{\pm0.08}$ & +4.51 & $0.57_{\pm0.26}$ & $0.75_{\pm0.26}$ & +32.76 & $0.75_{\pm0.21}$ & $0.85_{\pm0.20}$ & +12.60 \\
Keratinizing SCC Grade & $0.70_{\pm0.02}$ & $0.73_{\pm0.03}$ & +4.07 & $0.70_{\pm0.03}$ & $0.72_{\pm0.06}$ & +3.50 & $0.69_{\pm0.01}$ & $0.72_{\pm0.03}$ & +3.99 \\
Non-Keratinizing SCC Grade & $0.75_{\pm0.12}$ & $0.77_{\pm0.14}$ & +1.44 & $0.62_{\pm0.13}$ & $0.66_{\pm0.07}$ & +6.50 & $0.71_{\pm0.12}$ & $0.73_{\pm0.15}$ & +2.86 \\
Primary vs.\ Metastasis & $0.91_{\pm0.02}$ & $0.93_{\pm0.02}$ & +2.55 & $0.60_{\pm0.10}$ & $0.69_{\pm0.13}$ & +15.25 & $0.57_{\pm0.05}$ & $0.65_{\pm0.10}$ & +14.50 \\
\midrule
Macro average & 0.731 & 0.769 & +5.23 & 0.697 & 0.763 & +9.45 & 0.659 & 0.702 & +6.47 \\
\bottomrule
\end{tabular}%
}
\end{table}

\section{Stage 2 Only Without Case Aggregator and MOOZY Task-wise Comparison}
\label[appendix]{appendix_sec:stage2_nocaseagg_moozy_comparison}

\Cref{tab:stage2_nocaseagg_moozy_taskwise} provides the complete task-level breakdown for the Stage~2 only without case aggregator configuration versus MOOZY, discussed in \Cref{sec:results}. This configuration trains the slide encoder with multi-task supervision from scratch, without Stage~1 SSL pretraining, and aggregates per-slide embeddings by mean slide pooling rather than the case transformer. Across sixteen tasks, MOOZY improves weighted F1 and balanced accuracy on fourteen, and AUC on twelve; twelve tasks improve on all three metrics. The largest gains are Treatment Response F1 (+18.37\%) and balanced accuracy (+23.08\%), and RAS/BRAF Status AUC (+28.99\%). BAP1 mutation loses AUC, HNSC mRNA Subtype ties in AUC, and BRCA PAM50 and UCEC Genomic Subtype favor the ablated configuration across all three metrics. Macro averages improve by 3.50\% weighted F1, 5.86\% AUC, and 5.72\% balanced accuracy.

\begin{table}[tb]
\centering
\caption{Task-wise comparison between Stage~2 only without the case aggregator (slide encoder trained with multi-task supervision from scratch, no Stage~1 SSL pretraining, mean slide pooling) and MOOZY (Ours) across the sixteen slide-encoder evaluation tasks. Values are mean $\pm$ standard deviation across five folds. Relative improvement is computed as $(\text{MOOZY} - \text{Stage~2 w/o CA}) / \text{Stage~2 w/o CA} \times 100$ using fold means.}
\label{tab:stage2_nocaseagg_moozy_taskwise}
\small
\setlength{\tabcolsep}{3pt}
\resizebox{\textwidth}{!}{%
\begin{tabular}{l c c c c c c c c c}
\toprule
Task & F1 (S2 w/o CA) & F1 (MOOZY) & $\Delta$F1 (\%) & AUC (S2 w/o CA) & AUC (MOOZY) & $\Delta$AUC (\%) & Bal.\ Acc (S2 w/o CA) & Bal.\ Acc (MOOZY) & $\Delta$Bal.\ Acc (\%) \\
\midrule
Residual Cancer Burden & $0.51_{\pm0.01}$ & $0.56_{\pm0.05}$ & +9.80 & $0.63_{\pm0.02}$ & $0.74_{\pm0.04}$ & +17.46 & $0.48_{\pm0.06}$ & $0.51_{\pm0.06}$ & +6.25 \\
TP53 mutation & $0.80_{\pm0.07}$ & $0.87_{\pm0.04}$ & +8.75 & $0.79_{\pm0.09}$ & $0.86_{\pm0.06}$ & +8.86 & $0.79_{\pm0.07}$ & $0.86_{\pm0.05}$ & +8.86 \\
BAP1 mutation & $0.86_{\pm0.07}$ & $0.89_{\pm0.06}$ & +3.49 & $0.85_{\pm0.10}$ & $0.79_{\pm0.12}$ & -7.06 & $0.76_{\pm0.15}$ & $0.78_{\pm0.11}$ & +2.63 \\
ACVR2A mutation & $0.87_{\pm0.05}$ & $0.91_{\pm0.05}$ & +4.60 & $0.77_{\pm0.19}$ & $0.91_{\pm0.09}$ & +18.18 & $0.78_{\pm0.17}$ & $0.90_{\pm0.10}$ & +15.38 \\
Histologic Grade & $0.75_{\pm0.09}$ & $0.78_{\pm0.08}$ & +4.00 & $0.72_{\pm0.13}$ & $0.75_{\pm0.15}$ & +4.17 & $0.74_{\pm0.09}$ & $0.77_{\pm0.08}$ & +4.05 \\
KRAS mutation & $0.80_{\pm0.12}$ & $0.85_{\pm0.04}$ & +6.25 & $0.74_{\pm0.14}$ & $0.80_{\pm0.06}$ & +8.11 & $0.75_{\pm0.12}$ & $0.79_{\pm0.10}$ & +5.33 \\
IDH Status & $0.95_{\pm0.01}$ & $0.97_{\pm0.02}$ & +2.11 & $0.97_{\pm0.01}$ & $0.99_{\pm0.01}$ & +2.06 & $0.95_{\pm0.02}$ & $0.97_{\pm0.02}$ & +2.11 \\
Treatment Response & $0.49_{\pm0.06}$ & $0.58_{\pm0.14}$ & +18.37 & $0.63_{\pm0.07}$ & $0.68_{\pm0.07}$ & +7.94 & $0.39_{\pm0.06}$ & $0.48_{\pm0.17}$ & +23.08 \\
BRCA PAM50 Subtype & $0.65_{\pm0.02}$ & $0.63_{\pm0.03}$ & -3.08 & $0.82_{\pm0.03}$ & $0.80_{\pm0.01}$ & -2.44 & $0.51_{\pm0.03}$ & $0.50_{\pm0.05}$ & -1.96 \\
HNSC mRNA Subtype & $0.53_{\pm0.05}$ & $0.55_{\pm0.03}$ & +3.77 & $0.72_{\pm0.04}$ & $0.72_{\pm0.01}$ & +0.00 & $0.52_{\pm0.06}$ & $0.54_{\pm0.04}$ & +3.85 \\
UCEC Genomic Subtype & $0.60_{\pm0.05}$ & $0.56_{\pm0.05}$ & -6.67 & $0.77_{\pm0.03}$ & $0.74_{\pm0.04}$ & -3.90 & $0.56_{\pm0.05}$ & $0.52_{\pm0.05}$ & -7.14 \\
Synaptophysin Grade & $0.78_{\pm0.07}$ & $0.79_{\pm0.08}$ & +0.43 & $0.53_{\pm0.17}$ & $0.61_{\pm0.15}$ & +14.57 & $0.63_{\pm0.11}$ & $0.65_{\pm0.08}$ & +3.24 \\
RAS/BRAF Status & $0.90_{\pm0.09}$ & $0.94_{\pm0.08}$ & +4.95 & $0.58_{\pm0.24}$ & $0.75_{\pm0.26}$ & +28.99 & $0.73_{\pm0.23}$ & $0.85_{\pm0.20}$ & +15.91 \\
Keratinizing SCC Grade & $0.72_{\pm0.03}$ & $0.73_{\pm0.03}$ & +0.50 & $0.72_{\pm0.04}$ & $0.72_{\pm0.06}$ & +1.04 & $0.72_{\pm0.02}$ & $0.72_{\pm0.03}$ & +0.44 \\
Non-Keratinizing SCC Grade & $0.76_{\pm0.11}$ & $0.77_{\pm0.14}$ & +0.98 & $0.64_{\pm0.11}$ & $0.66_{\pm0.07}$ & +3.18 & $0.72_{\pm0.12}$ & $0.73_{\pm0.15}$ & +1.62 \\
Primary vs.\ Metastasis & $0.92_{\pm0.01}$ & $0.93_{\pm0.02}$ & +1.72 & $0.65_{\pm0.08}$ & $0.69_{\pm0.13}$ & +5.94 & $0.59_{\pm0.06}$ & $0.65_{\pm0.10}$ & +10.13 \\
\midrule
Macro average & 0.743 & 0.769 & +3.50 & 0.721 & 0.763 & +5.86 & 0.664 & 0.702 & +5.72 \\
\bottomrule
\end{tabular}%
}
\end{table}

\section{Case Aggregator Ablation: Mean Slide Pooling vs MOOZY}
\label[appendix]{appendix_sec:case_aggregator_ablation}

\Cref{tab:meanpool_moozy_taskwise} provides the complete task-level breakdown for the case-aggregator ablation under the MLP probe protocol. The table reports five-fold mean and standard deviation for MOOZY without the case aggregator (slide encoder alone with mean slide pooling) and full MOOZY, together with relative gains for weighted F1, weighted ROC-AUC, and balanced accuracy. Across sixteen tasks, the aggregator improves weighted F1 on fourteen, AUC on twelve, and balanced accuracy on thirteen. Eleven tasks improve on all three metrics. KRAS mutation, BRCA PAM50 Subtype, UCEC Genomic Subtype, and Synaptophysin Grade show metric-specific regressions, while the slide-level IDH Status task remains effectively unchanged. Macro averages improve by 2.69\% weighted F1, 3.50\% AUC, and 2.96\% balanced accuracy.

\begin{table}[tb]
\centering
\caption{Task-wise comparison between MOOZY w/o case aggregator (Stage~2 slide encoder alone with mean slide pooling) and MOOZY (Ours) across the sixteen slide-encoder evaluation tasks. Values are mean $\pm$ standard deviation across five folds. Relative improvement is computed as $(\text{MOOZY} - \text{MOOZY w/o case aggregator}) / \text{MOOZY w/o case aggregator} \times 100$ using fold means.}
\label{tab:meanpool_moozy_taskwise}
\small
\setlength{\tabcolsep}{3pt}
\resizebox{\textwidth}{!}{%
\begin{tabular}{l c c c c c c c c c}
\toprule
Task & F1 (w/o Case Agg.) & F1 (MOOZY) & $\Delta$F1 (\%) & AUC (w/o Case Agg.) & AUC (MOOZY) & $\Delta$AUC (\%) & Bal.\ Acc (w/o Case Agg.) & Bal.\ Acc (MOOZY) & $\Delta$Bal.\ Acc (\%) \\
\midrule
Residual Cancer Burden & $0.49_{\pm0.04}$ & $0.56_{\pm0.05}$ & +14.29 & $0.64_{\pm0.08}$ & $0.74_{\pm0.04}$ & +15.62 & $0.45_{\pm0.06}$ & $0.51_{\pm0.06}$ & +13.33 \\
TP53 mutation & $0.84_{\pm0.04}$ & $0.87_{\pm0.04}$ & +3.57 & $0.84_{\pm0.06}$ & $0.86_{\pm0.06}$ & +2.38 & $0.85_{\pm0.05}$ & $0.86_{\pm0.05}$ & +1.18 \\
BAP1 mutation & $0.87_{\pm0.05}$ & $0.89_{\pm0.06}$ & +2.30 & $0.76_{\pm0.14}$ & $0.79_{\pm0.12}$ & +3.95 & $0.76_{\pm0.12}$ & $0.78_{\pm0.11}$ & +2.63 \\
ACVR2A mutation & $0.88_{\pm0.05}$ & $0.91_{\pm0.05}$ & +3.41 & $0.89_{\pm0.10}$ & $0.91_{\pm0.09}$ & +2.25 & $0.80_{\pm0.12}$ & $0.90_{\pm0.10}$ & +12.50 \\
Histologic Grade & $0.76_{\pm0.06}$ & $0.78_{\pm0.08}$ & +2.63 & $0.74_{\pm0.10}$ & $0.75_{\pm0.15}$ & +1.35 & $0.76_{\pm0.06}$ & $0.77_{\pm0.08}$ & +1.32 \\
KRAS mutation & $0.84_{\pm0.05}$ & $0.85_{\pm0.04}$ & +1.19 & $0.79_{\pm0.08}$ & $0.80_{\pm0.06}$ & +1.27 & $0.81_{\pm0.06}$ & $0.79_{\pm0.10}$ & -2.47 \\
IDH Status & $0.96_{\pm0.01}$ & $0.97_{\pm0.02}$ & +1.04 & $0.99_{\pm0.01}$ & $0.99_{\pm0.01}$ & +0.00 & $0.96_{\pm0.01}$ & $0.97_{\pm0.02}$ & +1.04 \\
Treatment Response & $0.53_{\pm0.06}$ & $0.58_{\pm0.14}$ & +9.43 & $0.66_{\pm0.04}$ & $0.68_{\pm0.07}$ & +3.03 & $0.44_{\pm0.05}$ & $0.48_{\pm0.17}$ & +9.09 \\
BRCA PAM50 Subtype & $0.64_{\pm0.04}$ & $0.63_{\pm0.03}$ & -1.56 & $0.80_{\pm0.03}$ & $0.80_{\pm0.01}$ & +0.00 & $0.48_{\pm0.03}$ & $0.50_{\pm0.05}$ & +4.17 \\
HNSC mRNA Subtype & $0.54_{\pm0.05}$ & $0.55_{\pm0.03}$ & +1.85 & $0.70_{\pm0.04}$ & $0.72_{\pm0.01}$ & +2.86 & $0.53_{\pm0.03}$ & $0.54_{\pm0.04}$ & +1.89 \\
UCEC Genomic Subtype & $0.56_{\pm0.05}$ & $0.56_{\pm0.05}$ & +0.00 & $0.74_{\pm0.03}$ & $0.74_{\pm0.04}$ & +0.00 & $0.53_{\pm0.05}$ & $0.52_{\pm0.05}$ & -1.89 \\
Synaptophysin Grade & $0.78_{\pm0.07}$ & $0.79_{\pm0.08}$ & +0.78 & $0.66_{\pm0.14}$ & $0.61_{\pm0.15}$ & -7.88 & $0.69_{\pm0.07}$ & $0.65_{\pm0.08}$ & -6.46 \\
RAS/BRAF Status & $0.91_{\pm0.08}$ & $0.94_{\pm0.08}$ & +3.86 & $0.65_{\pm0.27}$ & $0.75_{\pm0.26}$ & +16.53 & $0.80_{\pm0.19}$ & $0.85_{\pm0.20}$ & +6.39 \\
Keratinizing SCC Grade & $0.72_{\pm0.02}$ & $0.73_{\pm0.03}$ & +1.60 & $0.71_{\pm0.06}$ & $0.72_{\pm0.06}$ & +1.95 & $0.71_{\pm0.03}$ & $0.72_{\pm0.03}$ & +1.54 \\
Non-Keratinizing SCC Grade & $0.74_{\pm0.12}$ & $0.77_{\pm0.14}$ & +2.77 & $0.56_{\pm0.07}$ & $0.66_{\pm0.07}$ & +16.64 & $0.72_{\pm0.11}$ & $0.73_{\pm0.15}$ & +1.52 \\
Primary vs.\ Metastasis & $0.92_{\pm0.02}$ & $0.93_{\pm0.02}$ & +0.78 & $0.67_{\pm0.11}$ & $0.69_{\pm0.13}$ & +2.91 & $0.61_{\pm0.11}$ & $0.65_{\pm0.10}$ & +6.93 \\
\midrule
Macro average & 0.749 & 0.769 & +2.69 & 0.737 & 0.763 & +3.50 & 0.682 & 0.702 & +2.96 \\
\bottomrule
\end{tabular}%
}
\end{table}

\section{Attention Map Generation and Additional Examples}
\label[appendix]{appendix_sec:attention_maps}

\phantomsection\label[appendix]{appendix_sec:attention_maps_generation}
Heatmaps are generated from WSIs randomly sampled across the held-out evaluation datasets. For each selected slide, we produce matched visualizations for all compared encoders using the same attribution objective, so all models are compared on a common relevance scale. Let $X=\{x_i\}_{i=1}^{N}$ denote patch tokens for one slide, and let $z^{(m)}(X)$ be the corresponding slide embedding from encoder $m$, where
\begin{equation}
m \in \{\text{CHIEF}, \text{Madeleine}, \text{PRISM}, \text{TITAN}, \text{MOOZY}\}.
\end{equation}
As the attribution target we use the squared $\ell_2$ norm of the slide embedding,
\begin{equation}
\phi^{(m)}(X)=\frac{1}{2}\left\lVert z^{(m)}(X)\right\rVert_2^2,
\end{equation}
which is an encoder-intrinsic scalar that requires no downstream task head, making comparisons across models fair. We assign each patch the Grad$\times$Input relevance
\begin{equation}
s_i^{(m)}=\left\lVert x_i \odot \frac{\partial \phi^{(m)}}{\partial x_i}\right\rVert_1.
\end{equation}
Intuitively, $s_i^{(m)}$ measures how much a small perturbation of patch $x_i$ shifts $\phi^{(m)}$, so high-score regions are those most influential to the encoder's slide-level representation. The relevance vector $\{s_i^{(m)}\}_{i=1}^{N}$ is then converted into a display map. We first apply rank normalization
\begin{equation}
\tilde{s}_i^{(m)}=\frac{\mathrm{rank}(s_i^{(m)})}{N},
\end{equation}
then project normalized patch scores to level-0 patch boxes on the thumbnail and average them per pixel where boxes overlap. The resulting maps are displayed as matched slide-level overlays.

\begin{figure*}[tb]
  \centering
  \includegraphics[width=\linewidth]{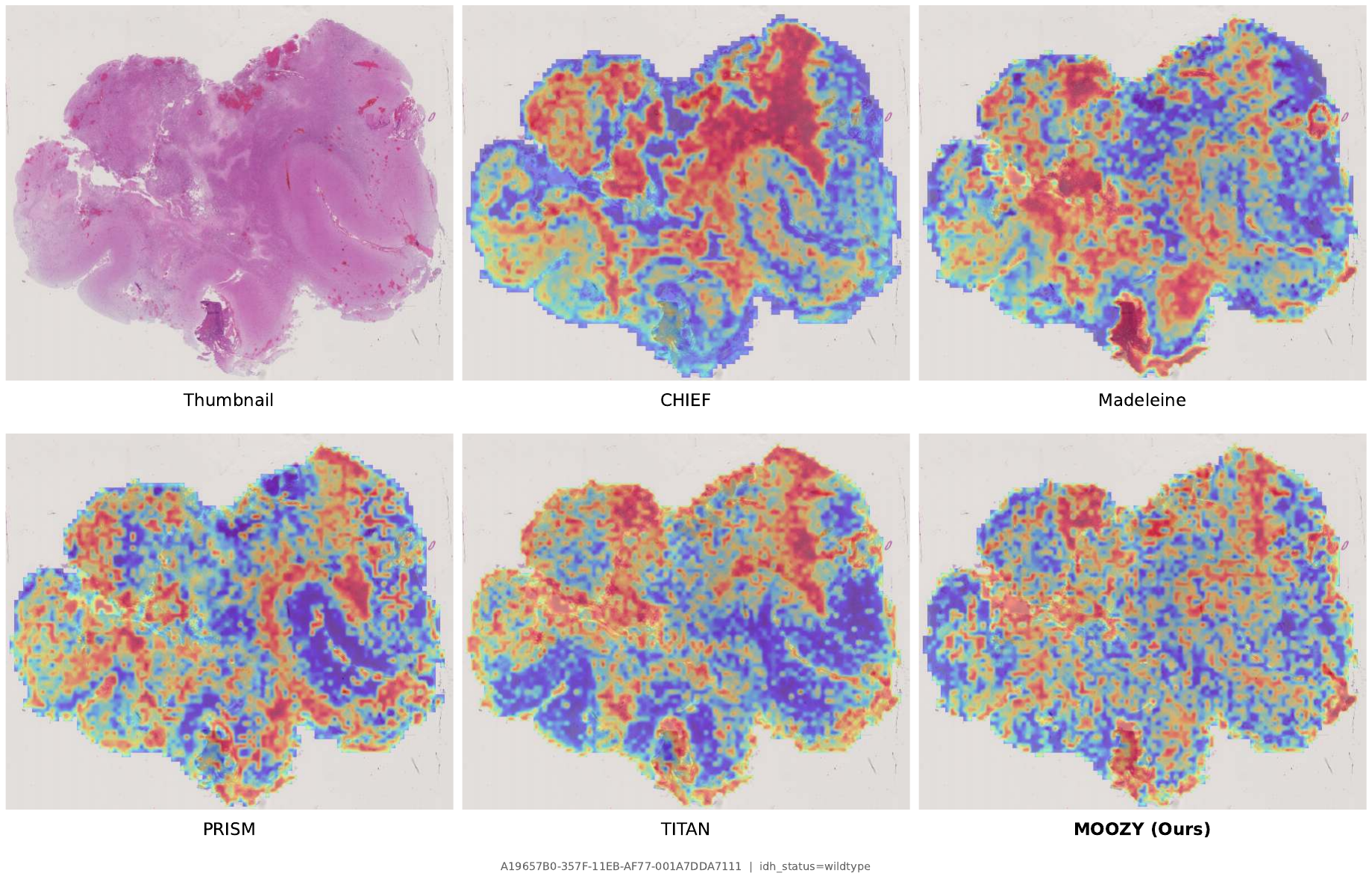}
  \captionsetup{skip=2pt}
  \caption{Additional attention map comparison across MOOZY and benchmarked models (example~2).}
  \label{fig:attention_maps_fig2}
\end{figure*}

\begin{figure*}[tb]
  \centering
  \includegraphics[width=\linewidth]{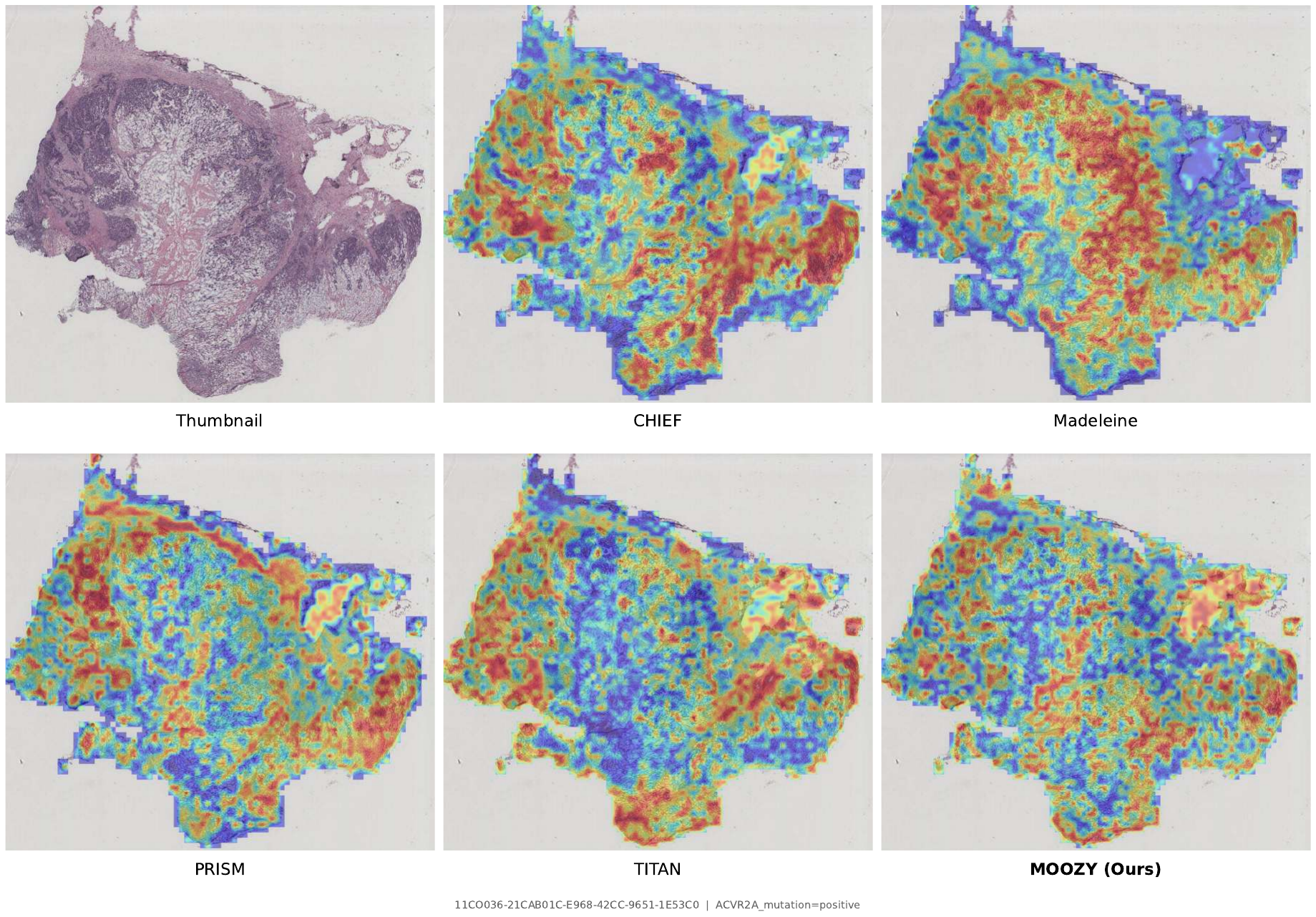}
  \captionsetup{skip=2pt}
  \caption{Additional attention map comparison across MOOZY and benchmarked models (example~3).}
  \label{fig:attention_maps_fig3}
\end{figure*}

\begin{figure*}[tb]
  \centering
  \includegraphics[width=\linewidth]{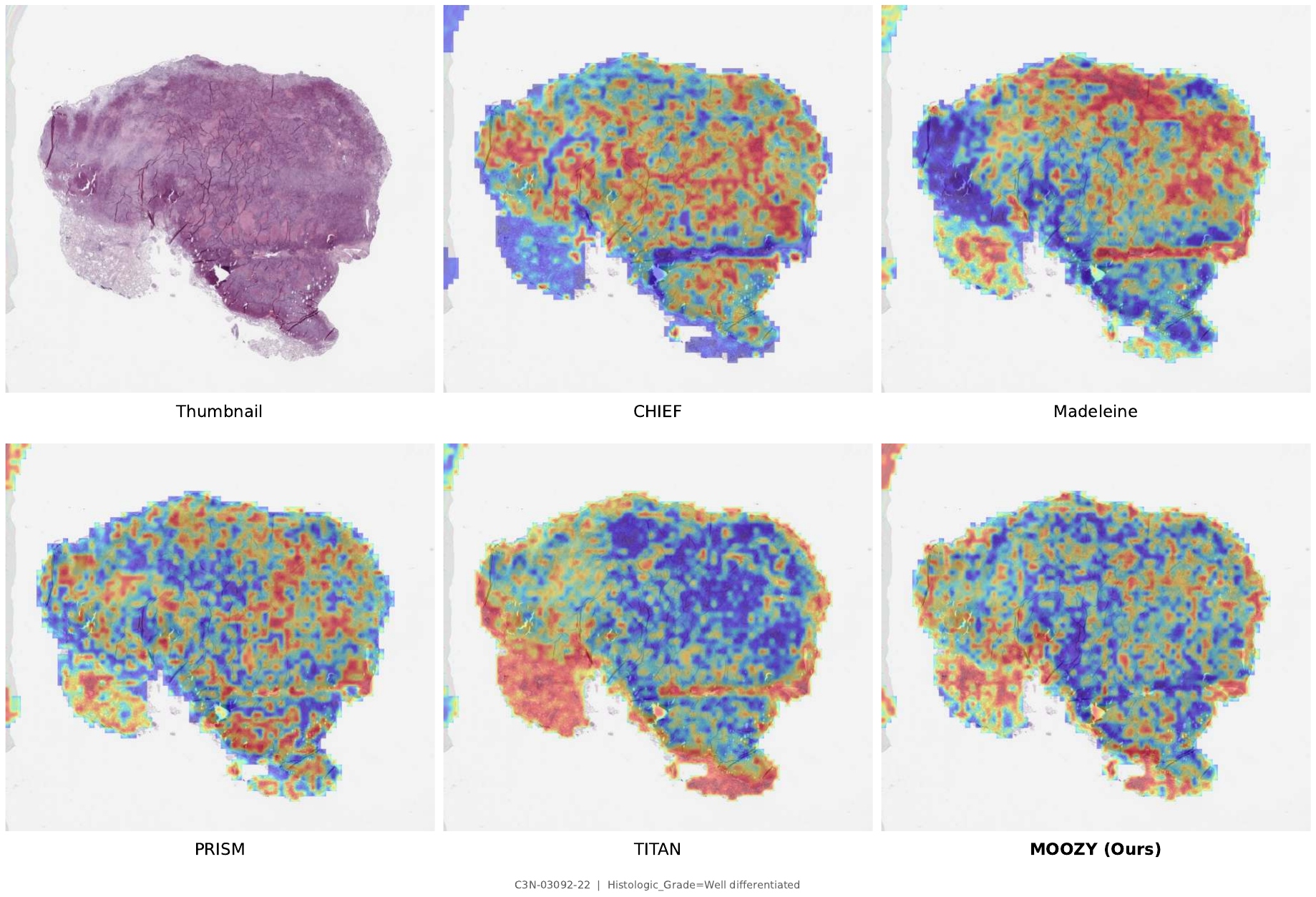}
  \captionsetup{skip=2pt}
  \caption{Additional attention map comparison across MOOZY and benchmarked models (example~4).}
  \label{fig:attention_maps_fig4}
\end{figure*}

\begin{figure*}[tb]
  \centering
  \includegraphics[width=\linewidth]{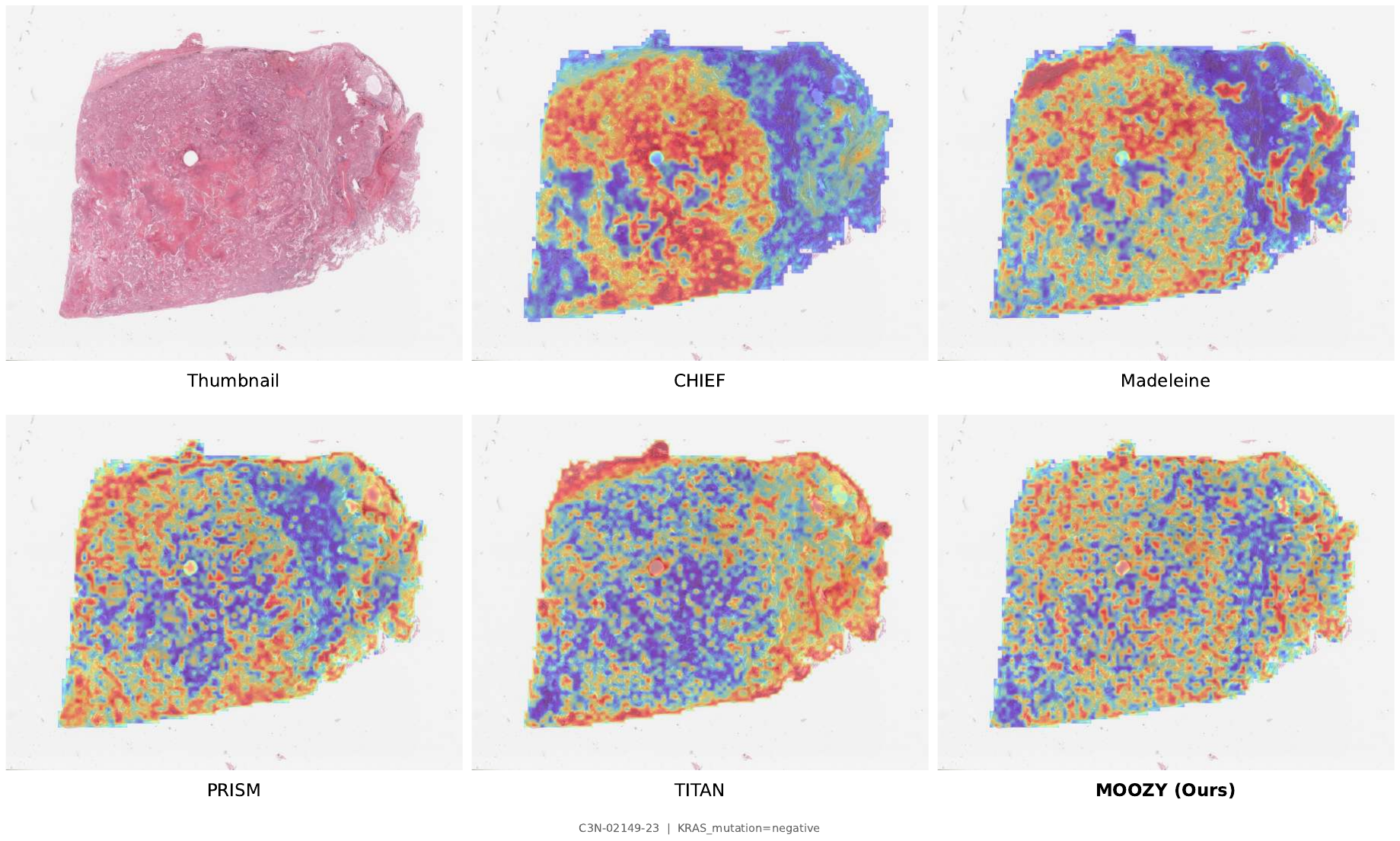}
  \captionsetup{skip=2pt}
  \caption{Additional attention map comparison across MOOZY and benchmarked models (example~5).}
  \label{fig:attention_maps_fig5}
\end{figure*}

\begin{figure*}[tb]
  \centering
  \includegraphics[width=\linewidth]{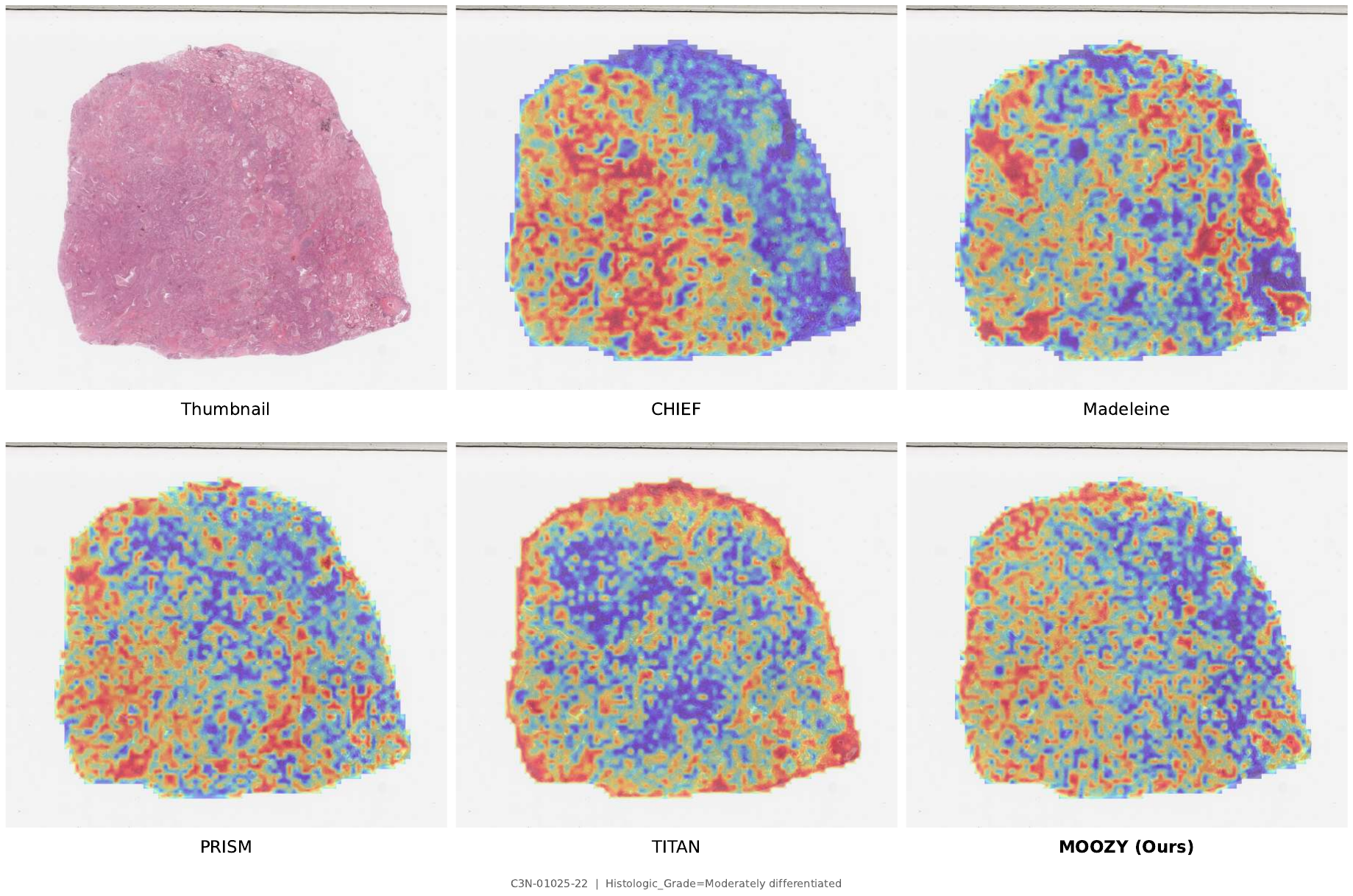}
  \captionsetup{skip=2pt}
  \caption{Additional attention map comparison across MOOZY and benchmarked models (example~6).}
  \label{fig:attention_maps_fig6}
\end{figure*}

Further cross-model attention-map comparisons are shown in \Cref{fig:attention_maps_fig2,fig:attention_maps_fig3,fig:attention_maps_fig4,fig:attention_maps_fig5,fig:attention_maps_fig6}. Together, these examples illustrate the same qualitative patterns discussed in \Cref{sec:results} across different slides and cohorts.

\section{Embedding Visualization Settings and t-SNE Results}
\label[appendix]{appendix_sec:embedding_visualization_config}

For all qualitative plots (\Cref{fig:embedding_qualitative_umap,fig:embedding_qualitative_tsne}), t-SNE uses perplexity 25 and 1000 optimization iterations, while UMAP uses neighborhood size 120, minimum distance 0.3, and cosine metric. Effective sample counts are $N{=}2152$ for CPTAC cancer type, $N{=}1172$ for anatomical site, and $N{=}1280$ for TCGA cancer type.

The t-SNE layouts below exhibit patterns consistent with the UMAP results in \Cref{fig:embedding_qualitative_umap}: MOOZY shows the clearest cluster separation on CPTAC and TCGA cancer type, TITAN is the strongest on anatomical site, and Madeleine and PRISM show comparatively weaker boundaries.

\providecommand{\embcell}[1]{%
  \includegraphics[width=\linewidth,trim=6 6 6 6,clip]{#1}%
}
\providecommand{\embcellfirst}[1]{%
  \includegraphics[width=\linewidth,trim=16 6 6 6,clip]{#1}%
}
\providecommand{\embcolhdr}[1]{{\fontsize{6.2}{6.8}\selectfont #1}}
\providecommand{\embrowhdr}[1]{%
  \parbox[c]{\linewidth}{\raggedright\fontsize{6.2}{6.8}\selectfont #1}%
}

\begin{figure*}[tb]
  \centering
  \setlength{\tabcolsep}{0pt}
  \renewcommand{\arraystretch}{0.88}
  \begin{tabular}{>{\raggedright\arraybackslash}m{0.105\textwidth}*{4}{>{\centering\arraybackslash}m{0.221\textwidth}}}
    & \embcolhdr{MOOZY} & \embcolhdr{TITAN} & \embcolhdr{Madeleine} & \embcolhdr{PRISM} \\
    \embrowhdr{CPTAC cancer type} &
      \embcellfirst{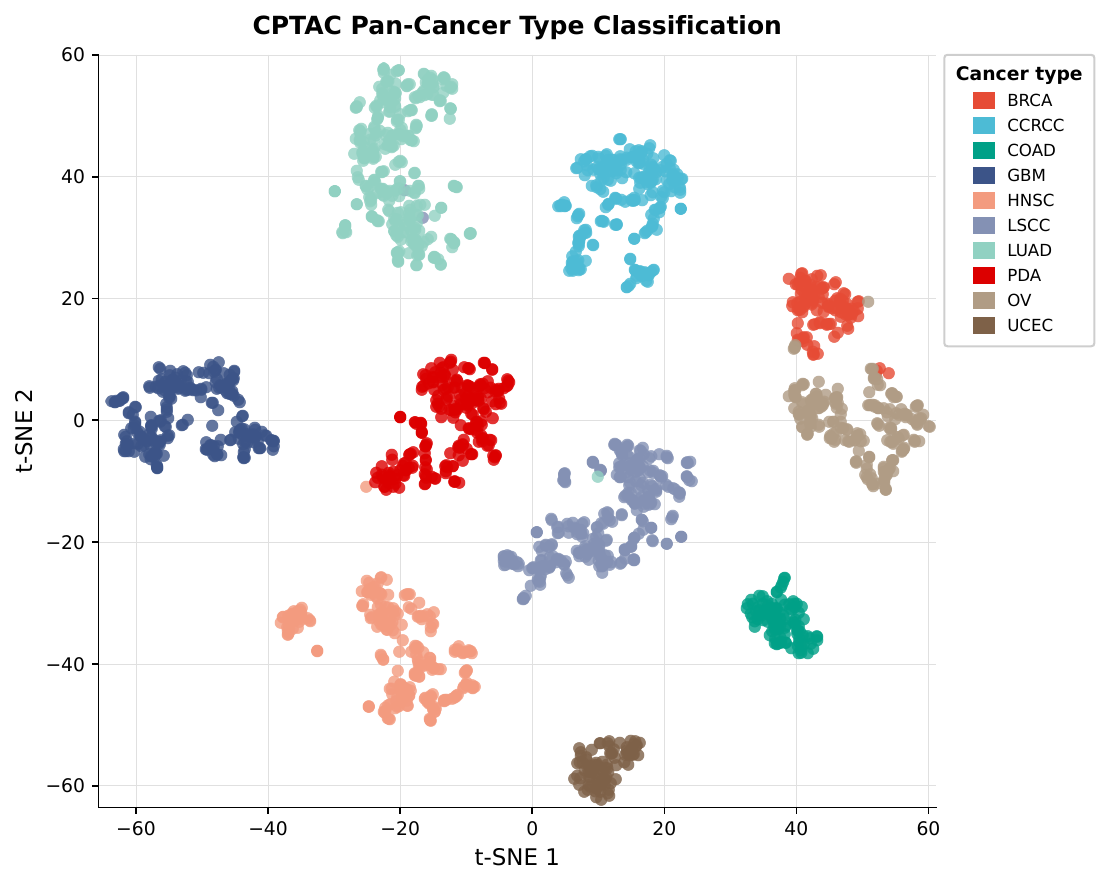} &
      \embcell{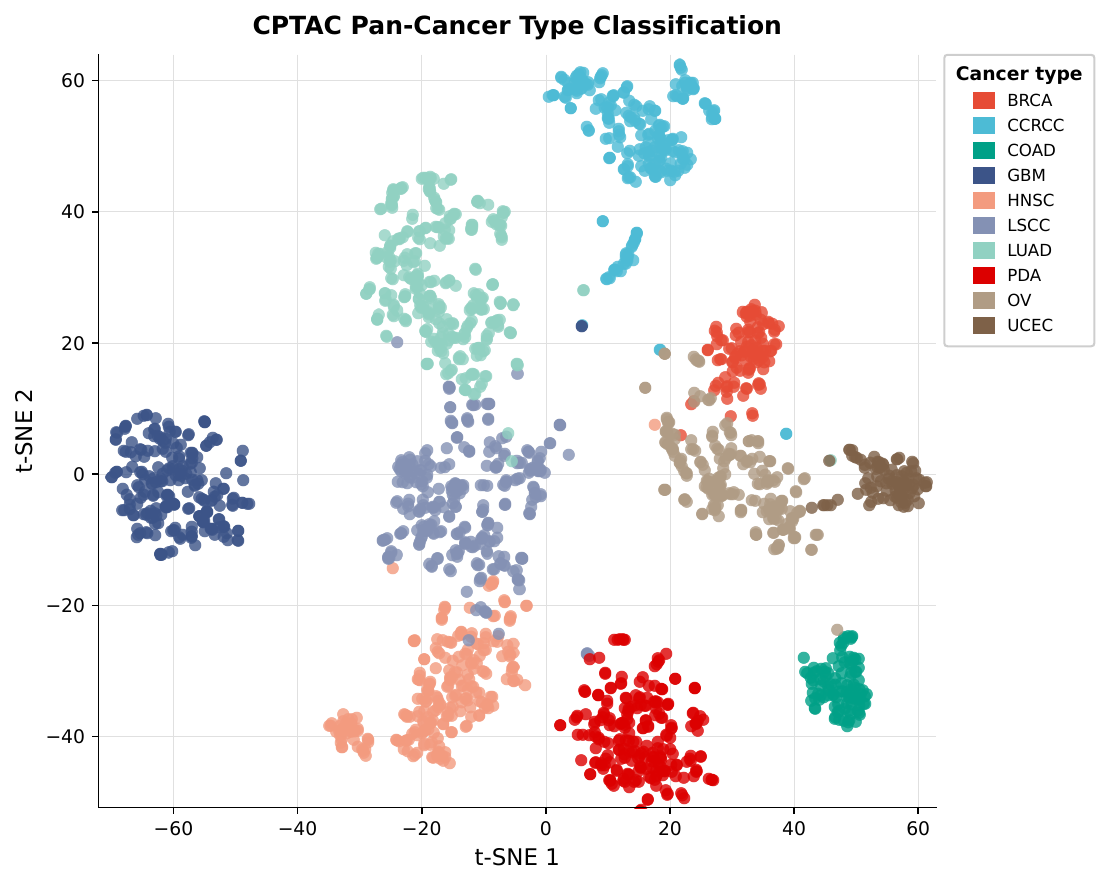} &
      \embcell{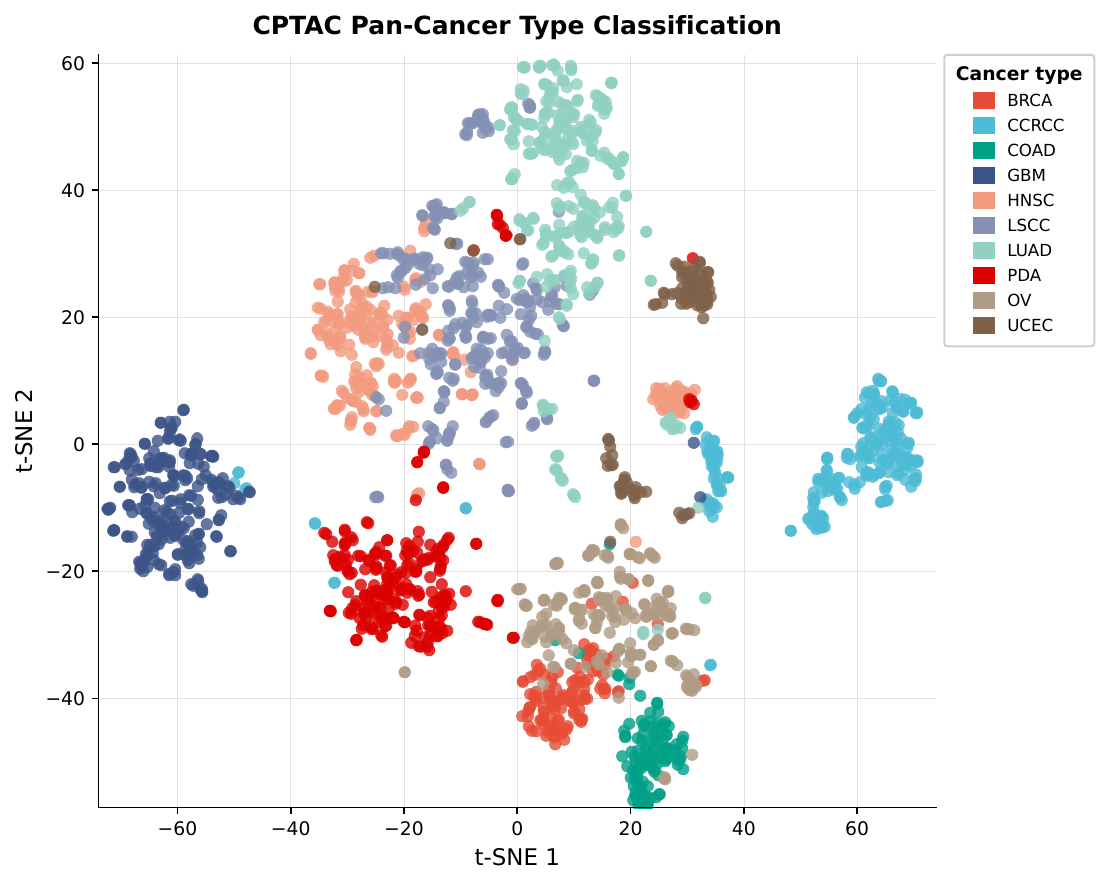} &
      \embcell{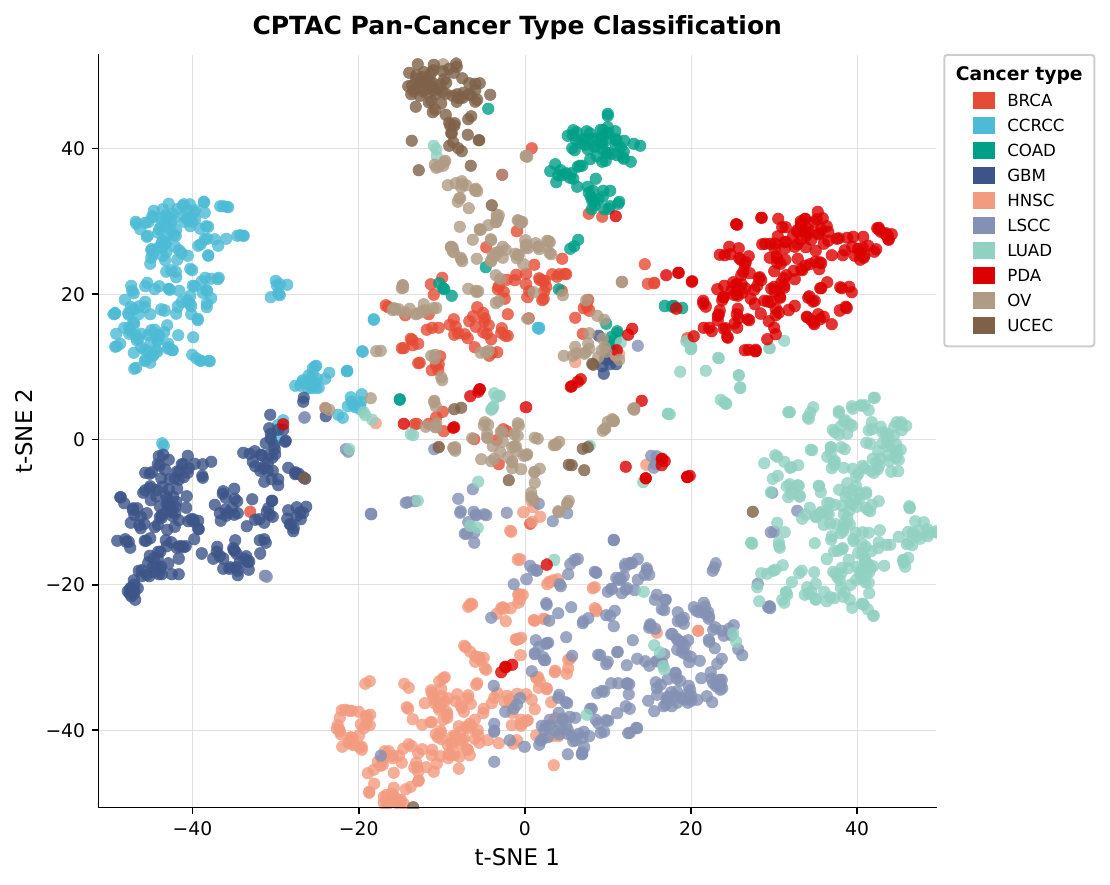} \\
    \embrowhdr{Anatomical site} &
      \embcellfirst{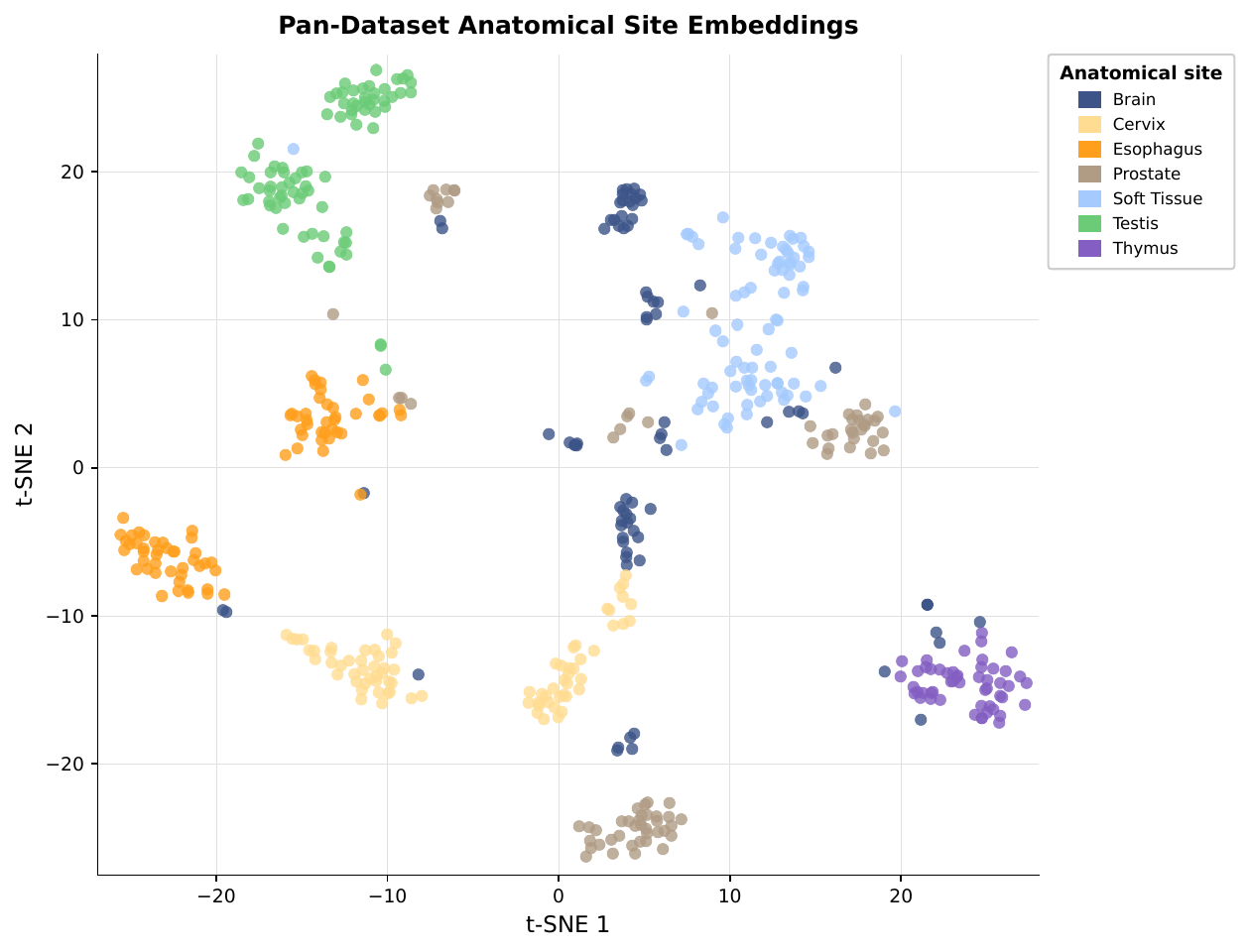} &
      \embcell{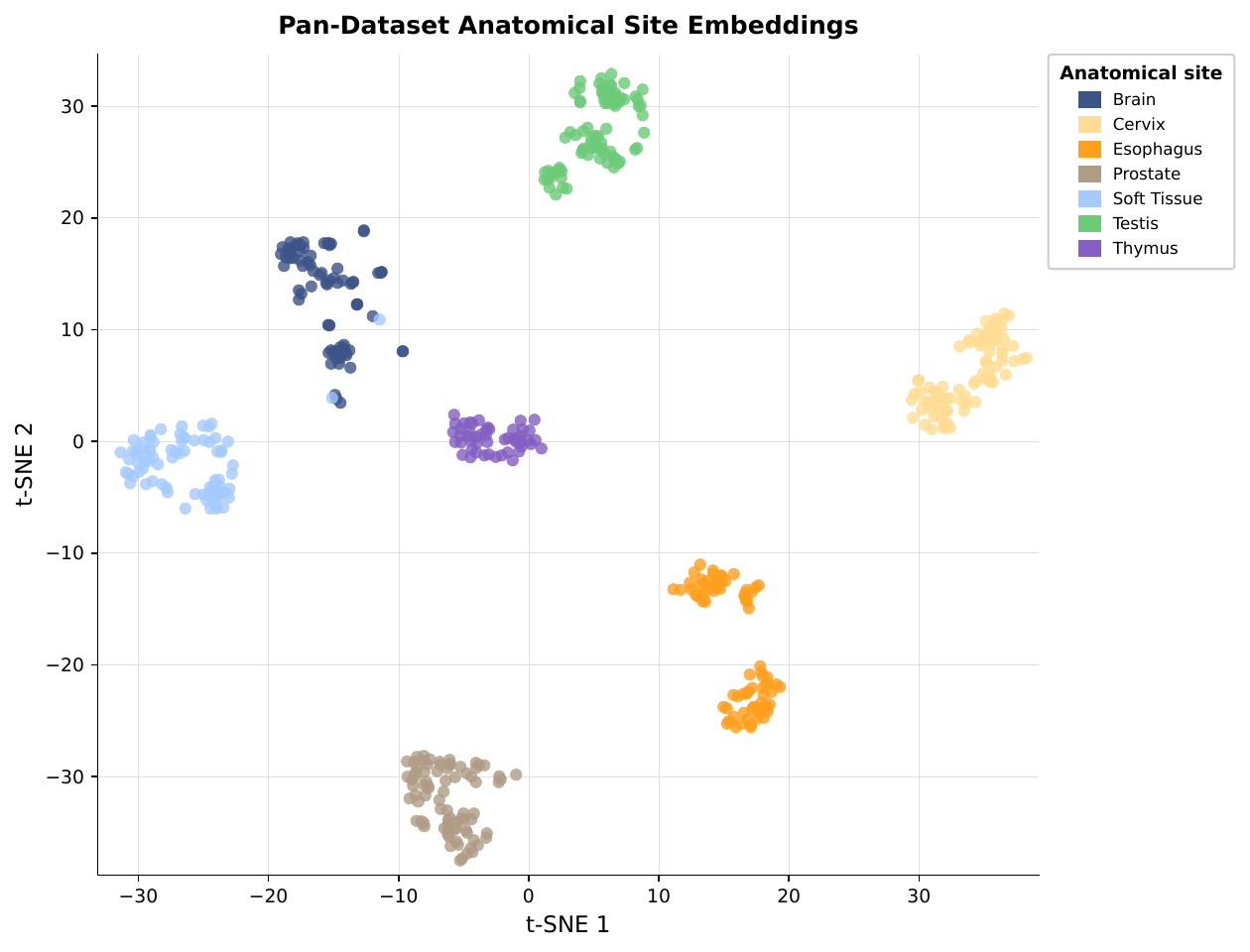} &
      \embcell{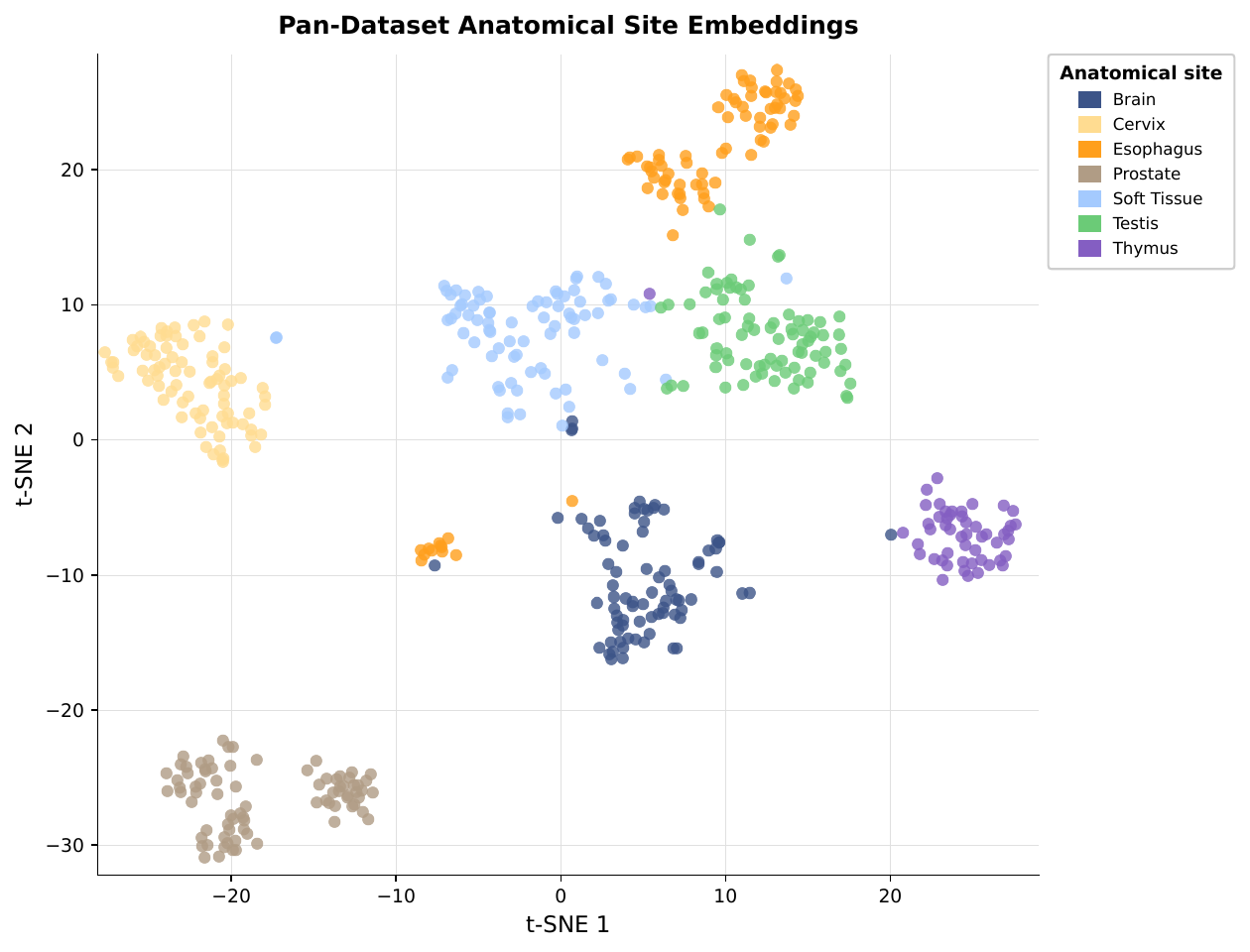} &
      \embcell{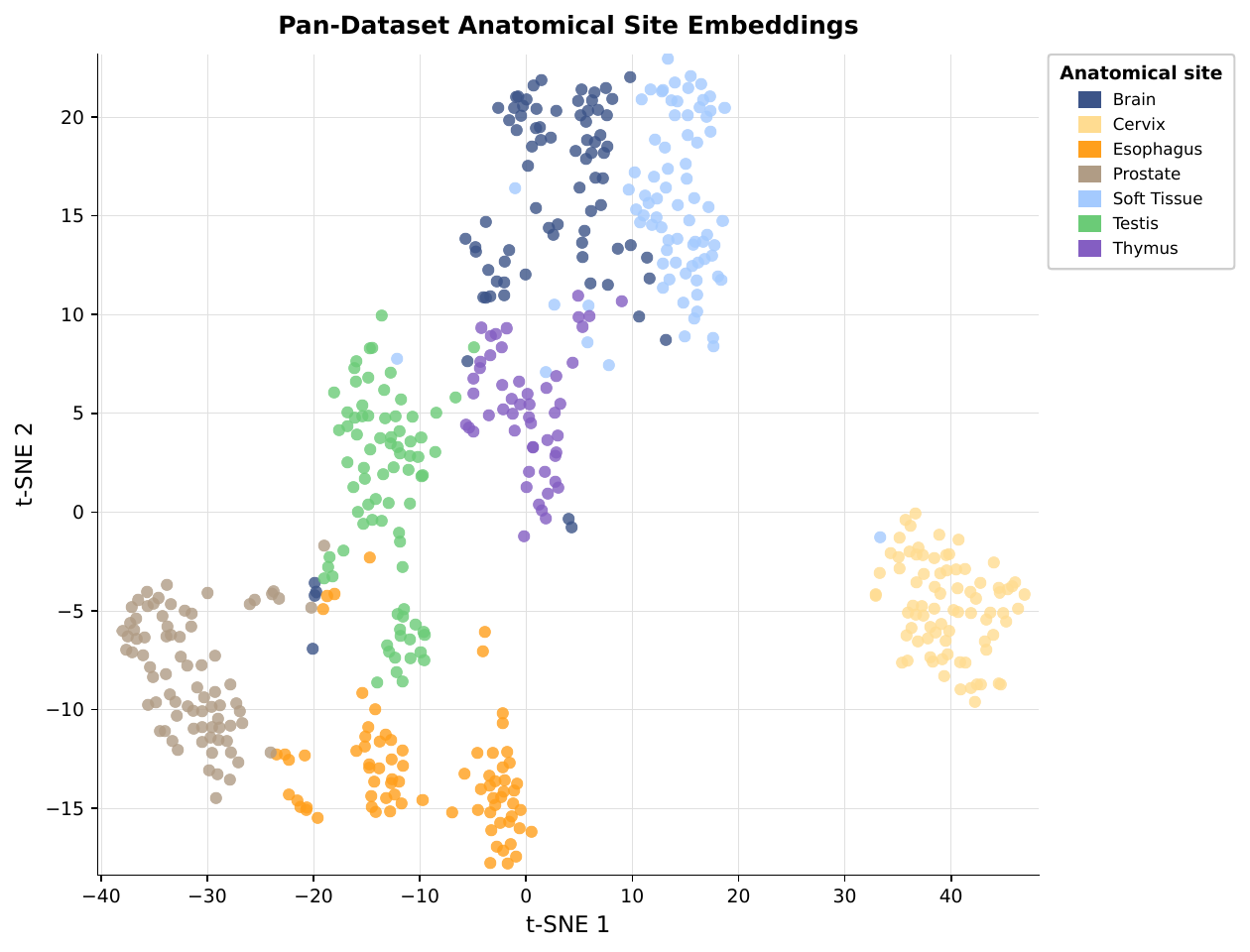} \\
    \embrowhdr{TCGA cancer type} &
      \embcellfirst{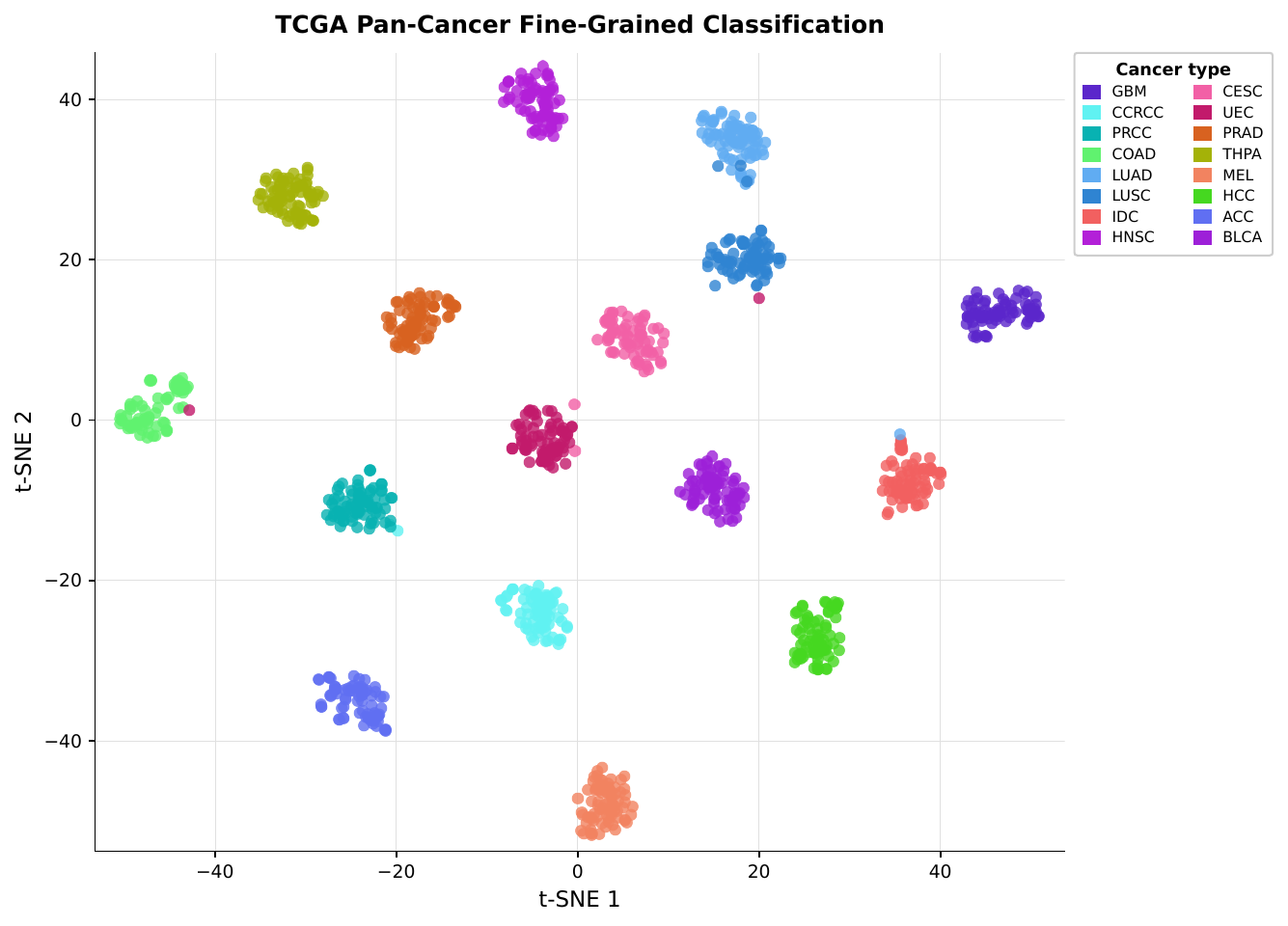} &
      \embcell{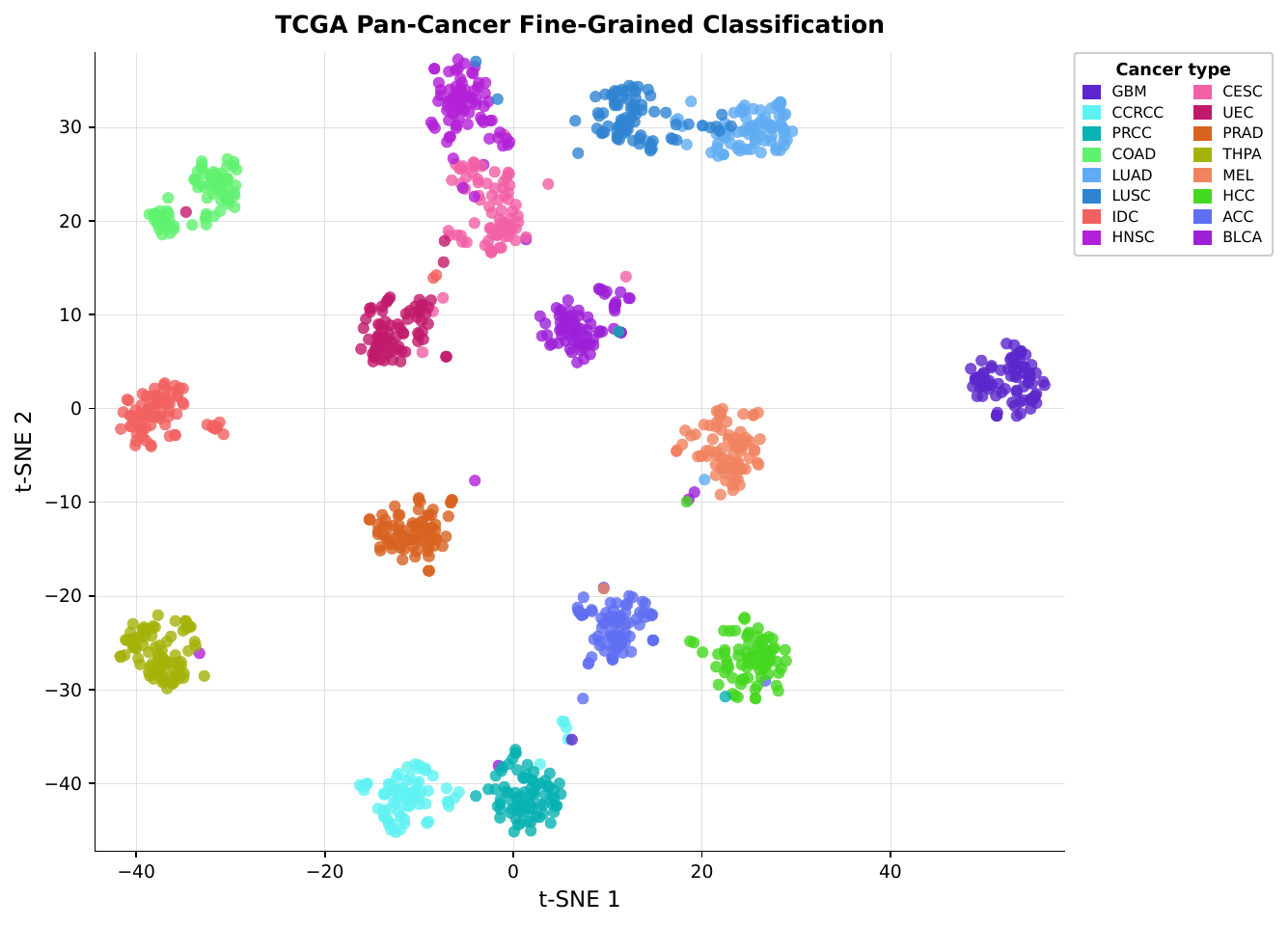} &
      \embcell{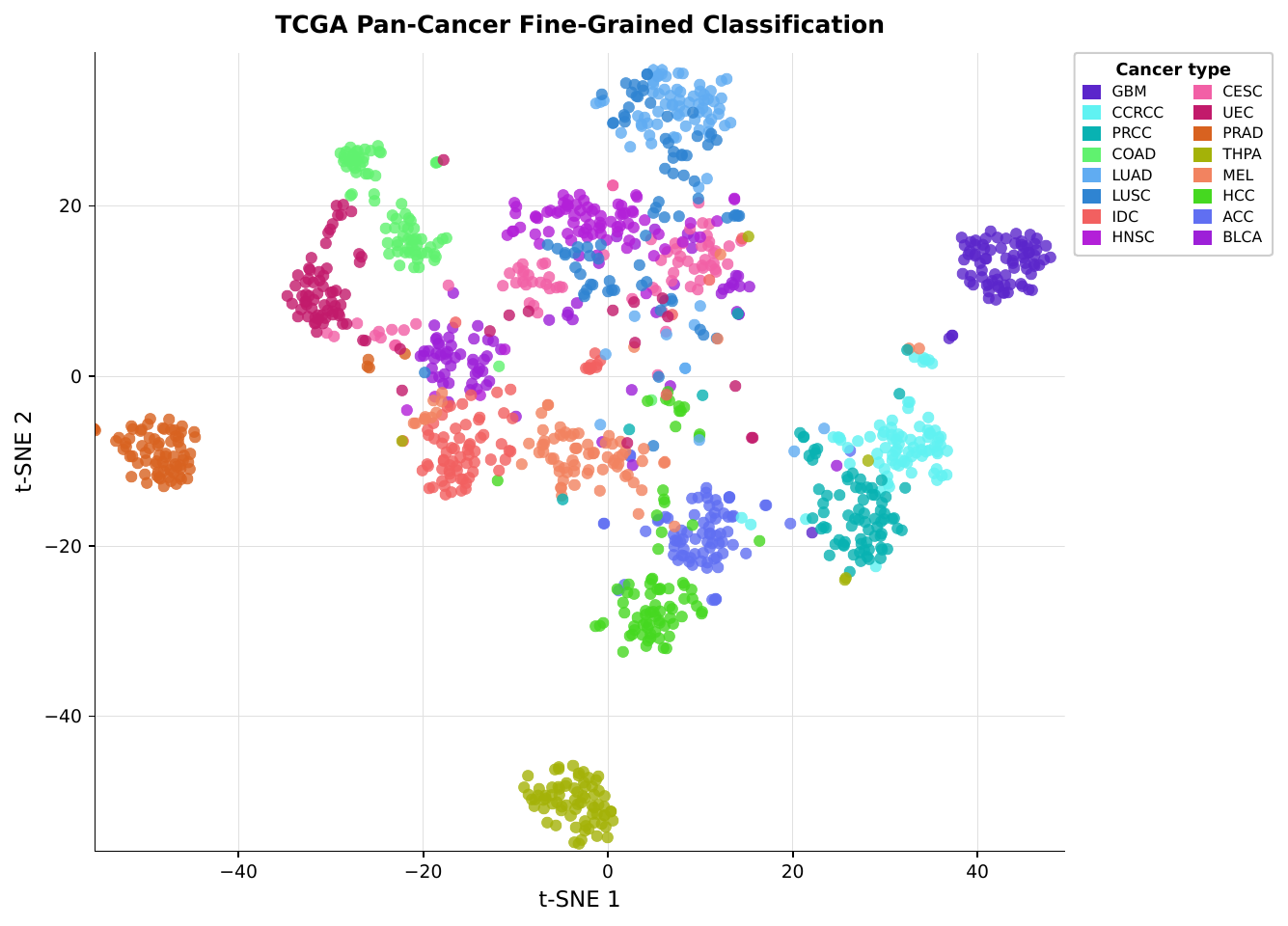} &
      \embcell{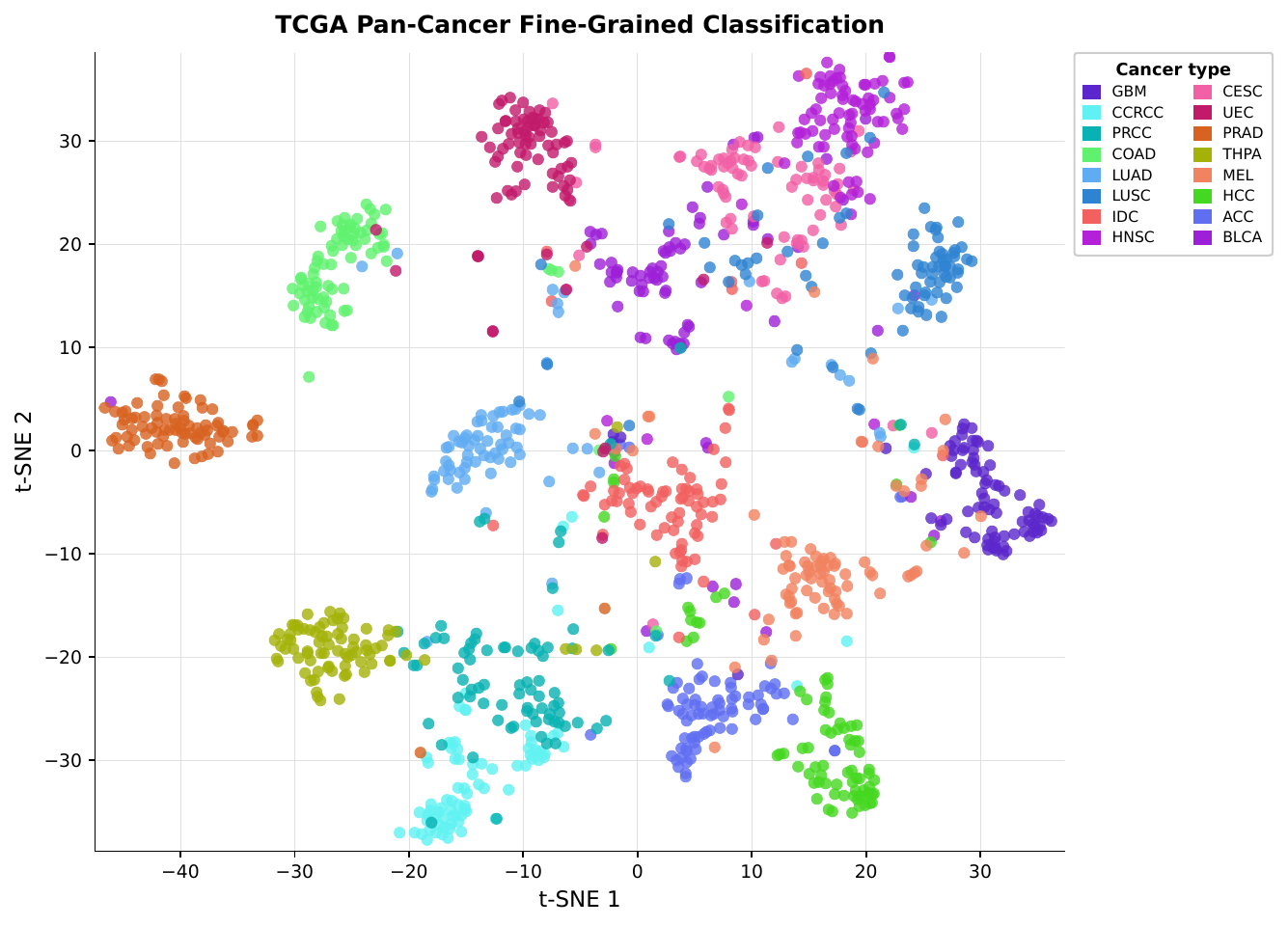} \\
  \end{tabular}
  \captionsetup{skip=3pt}
  \caption{t-SNE qualitative comparison across four slide encoders (columns) and three tasks (rows).}
  \label{fig:embedding_qualitative_tsne}
\end{figure*}

\section{Unsupervised Embedding Geometry Analysis}
\label[appendix]{appendix_sec:pca_stability_math}

\begin{figure*}[tb]
  \centering
  \includegraphics[width=\linewidth]{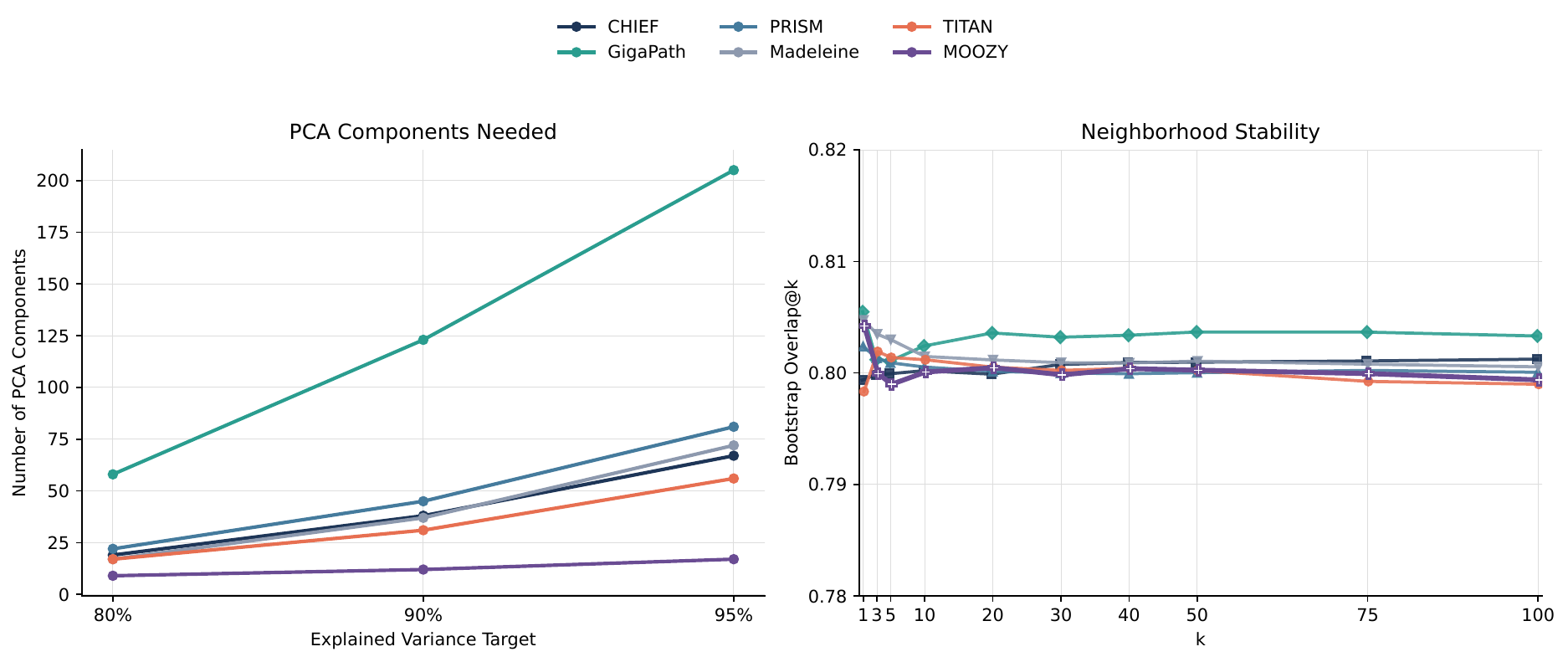}
  \caption{Encoder geometry comparison on 3{,}300 different slides. Left: PCA compactness, measured as the number of components needed to explain 80\%, 90\%, and 95\% variance (lower is more compact). Right: bootstrap neighborhood stability measured by overlap@k, where each repeat randomly subsamples 80\% of slides (higher is more stable).}
  \label{fig:encoder_pca_stability_qualitative}
\end{figure*}

We analyze embedding geometry on 3{,}300 unique slides using two unsupervised diagnostics shown in \Cref{fig:encoder_pca_stability_qualitative}: PCA compactness and bootstrap neighborhood stability. Let $X \in \mathbb{R}^{N \times D}$ denote the embedding matrix (one row per slide). PCA compactness measures how many orthogonal directions are needed to explain embedding variance.
We first center features:
\begin{equation}
\tilde{X} = X - \mathbf{1}\mu^\top, \quad
\mu = \frac{1}{N}\sum_{i=1}^{N} x_i,
\end{equation}
then compute covariance:
\begin{equation}
C = \frac{\tilde{X}^\top \tilde{X}}{N-1}.
\end{equation}
If $\lambda_1 \geq \lambda_2 \geq \dots \geq \lambda_D$ are eigenvalues of $C$, we clamp
\begin{equation}
\lambda_j \leftarrow \max(\lambda_j, 0)
\end{equation}
to avoid numerical negatives, and compute cumulative explained variance:
\begin{equation}
V(r) = \frac{\sum_{j=1}^{r} \lambda_j}{\sum_{j=1}^{D} \lambda_j}.
\end{equation}
For each threshold $\tau \in \{0.80, 0.90, 0.95\}$, compactness is the smallest rank $r_\tau$ such that $V(r_\tau)\geq\tau$.
Lower $r_\tau$ indicates less redundancy and more information-efficient embeddings. MOOZY is the most compact encoder, requiring 9/\allowbreak12/\allowbreak17 components at 80\%/\allowbreak90\%/\allowbreak95\% variance, versus TITAN (17/\allowbreak31/\allowbreak56), CHIEF (19/\allowbreak38/\allowbreak67), Madeleine (17/\allowbreak37/\allowbreak72), PRISM (22/\allowbreak45/\allowbreak81), and GigaPath (58/\allowbreak123/\allowbreak205).

To quantify robustness of local neighborhood structure under sampling perturbations, we first L2-normalize each embedding:
\begin{equation}
\hat{x}_i = \frac{x_i}{\max(\lVert x_i \rVert_2, 10^{-12})}.
\end{equation}
Using cosine distance, let $\mathcal{N}_{k}^{\mathrm{full}}(i)$ be the $k$ nearest neighbors of slide $i$ on the full set (excluding self).
For bootstrap repeat $b$, sample a subset $S_b \subset \{1,\dots,N\}$ uniformly without replacement:
\begin{equation}
|S_b| = m = \mathrm{round}(\rho N), \quad \rho = 0.8.
\end{equation}
Recompute neighbors on the subset to obtain $\mathcal{N}_{k,b}^{\mathrm{sub}}(i)$ for $i \in S_b$, and define per-slide overlap:
\begin{equation}
o_{i,k,b} = \frac{\left|\mathcal{N}_{k}^{\mathrm{full}}(i) \cap \mathcal{N}_{k,b}^{\mathrm{sub}}(i)\right|}{k},
\end{equation}
repeat-level overlap:
\begin{equation}
\bar{o}_{k,b} = \frac{1}{|S_b|}\sum_{i \in S_b} o_{i,k,b},
\end{equation}
and the final stability curve over $B$ repeats:
\begin{equation}
\mu_k = \frac{1}{B}\sum_{b=1}^{B} \bar{o}_{k,b}.
\end{equation}
Higher $\mu_k$ indicates more stable local structure. In practice, encoders are near-tied on this metric: at $k{=}30$, scores span 0.7998 to 0.8032, and the mean spread across $k$ is only 0.0029 (0.8002 to 0.8031). These unsupervised tests show that MOOZY has the strongest compactness while maintaining stability comparable to all baselines, indicating better representation efficiency without a meaningful robustness tradeoff.

\end{document}